\documentclass[11pt]{article}

\usepackage[final]{acl}

\usepackage{times}
\usepackage{latexsym}

\usepackage[T1]{fontenc}

\usepackage[utf8]{inputenc}

\usepackage{microtype}

\usepackage{inconsolata}

\usepackage{graphicx}

\usepackage{xspace}
\newcommand{\dataset}{\textsc{Mined}\xspace}

\usepackage[table]{xcolor}

\usepackage{dashrule}

\newcommand{\nbi}[1]{\noindent\textbf{\textit{#1}}}

\newcommand{\ie}{\textit{i.e., }}
\newcommand{\eg}{\textit{e.g., }}

\newcommand{\checkicon}{\raisebox{-.25em}{\includegraphics[width=1em]{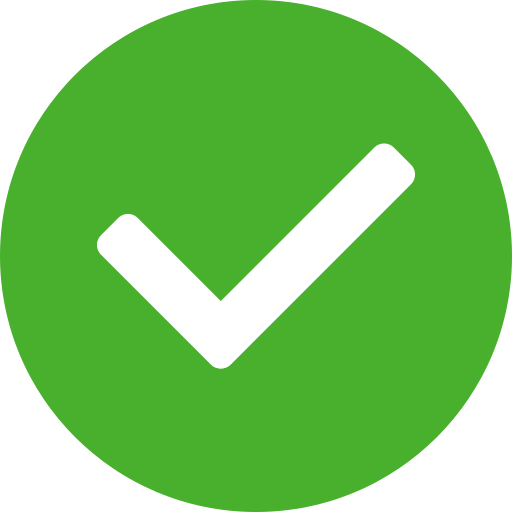}}}
\newcommand{\crossicon}{\raisebox{-.25em}{\includegraphics[width=1em]{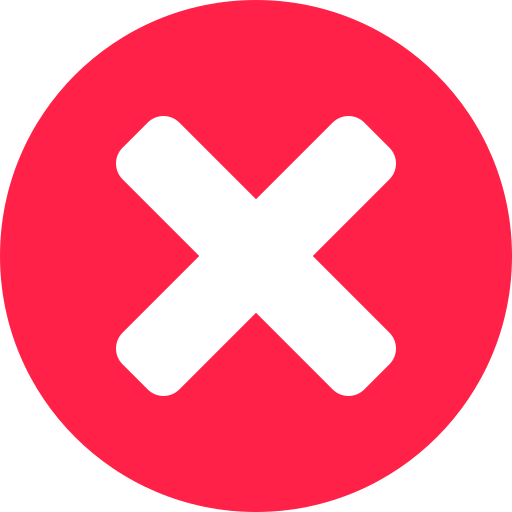}}}

\usepackage{booktabs}
\usepackage{enumitem}
\usepackage{pifont}
\usepackage{adjustbox}
\usepackage{caption}

\newcommand{\dalgshifted}{\raisebox{0.5\depth}{$\downarrow$}}
\newcommand{\daugshifted}{\raisebox{0.5\depth}{$\uparrow$}}

\newcommand{\en}[1]{``#1''}

\definecolor{backred}{RGB}{255, 190, 190}
\definecolor{backblue}{RGB}{210, 230, 250}
\definecolor{verylightgray}{gray}{0.95} 
\definecolor{backorange}{RGB}{255, 230, 204}
\definecolor{backyellow_soft}{RGB}{255, 252, 51}

\definecolor{myred}{RGB}{192,0,0}
\definecolor{mygreen}{RGB}{88,142,49}

\definecolor{verylightgray}{gray}{0.95} 
\definecolor{tong}{RGB}{152, 1, 0}

\definecolor{mygray}{gray}{.9}
\definecolor{mylinkcolor}{RGB}{115,194,251}

\usepackage{multirow}

\usepackage{amsmath} 
\usepackage[most]{tcolorbox}

\usepackage{amssymb}

\newcommand{\inbraces}[1]{\{#1\}}

\newcommand{\bfit}[1]{\textbf{\textit{#1}}}

\tcbset{promptstyle/.style={
  enhanced,
  boxrule=1pt,
  attach boxed title to top,
  coltitle=white,
  fonttitle=\large\bfseries,
  drop fuzzy shadow,
  halign=left,
  left=2mm
}}

\newtcolorbox[auto counter, number within=section]{graybox}[2][]{
  promptstyle,
  colback=gray!10,  
  colframe=gray!50!black, 
  colbacktitle=gray!80!black,  
  title={#2},
  #1
}

\newtcolorbox[auto counter, number within=section]{bluebox}[2][]{
  promptstyle,
  colback=blue!7,
  colframe=blue!45!black,
  colbacktitle=blue!55!black,
  title={#2},
  #1
}

\newtcolorbox[auto counter, number within=section]{greenbox}[2][]{
  promptstyle,
  colback=green!5,
  colframe=green!45!black,
  colbacktitle=green!55!black,
  title={#2},
  #1
}

\newtcolorbox[auto counter, number within=section]{redbox}[2][]{
  promptstyle,
  colback=red!5,
  colframe=myred!70!black,
  colbacktitle=myred!80!black,
  title={#2},
  #1
}

\newtcolorbox[auto counter, number within=section]{orangebox}[2][]{
  promptstyle,
  colback=orange!10,
  colframe=orange!70!black,
  colbacktitle=orange!80!black,
  title={#2},
  #1
}

\newtcolorbox[auto counter, number within=section]{cyanbox}[2][]{
  promptstyle,
  colback=cyan!5,
  colframe=cyan!70!black,
  colbacktitle=cyan!80!black,
  title={#2},
  #1
}

\newtcolorbox[auto counter, number within=section]{purplebox}[2][]{
  promptstyle,
  colback=purple!5,
  colframe=purple!70!black,
  colbacktitle=purple!80!black,
  title={#2},
  #1
}

\newtcolorbox[auto counter, number within=section]{pinkbox}[2][]{
  promptstyle,
  colback=pink!10,
  colframe=pink!70!black,
  colbacktitle=pink!80!black,
  title={#2},
  #1
}

\newtcolorbox[auto counter, number within=section]{brownbbox}[2][]{
  promptstyle,
  colback=brown!10,
  colframe=brown!70!black,
  colbacktitle=brown!80!black,
  title={#2},
  #1
}

\newtcolorbox[auto counter, number within=section]{creambox}[2][]{
  promptstyle,
  colback=yellow!10,
  colframe=yellow!70!black,
  colbacktitle=yellow!50!brown,
  title={#2},
  #1
}

\newtcolorbox[auto counter, number within=section]{olivebox}[2][]{
  promptstyle,
  colback=olive!10,
  colframe=olive!40!black,
  colbacktitle=olive!60!black,
  title={#2},
  #1
}

\newtcolorbox[auto counter, number within=section]{skybox}[2][]{
  promptstyle,
  colback=cyan!7,
  colframe=cyan!40!black,
  colbacktitle=cyan!60!black,
  title={#2},
  #1
}

\newtcolorbox[auto counter, number within=section]{lavenderbox}[2][]{
  promptstyle,
  colback=violet!10,
  colframe=violet!40!black,
  colbacktitle=violet!60!black,
  title={#2},
  #1
}

\newtcolorbox[auto counter, number within=section]{rosebox}[2][]{
  promptstyle,
  colback=pink!10,
  colframe=pink!40!black,
  colbacktitle=pink!60!black,
  title={#2},
  #1
}

\newtcolorbox[auto counter, number within=section]{peachbox}[2][]{
  promptstyle,
  colback=orange!15!pink!10,
  colframe=orange!40!black,
  colbacktitle=orange!60!black,
  title={#2},
  #1
}

\title{MINED: Probing and Updating with Multimodal Time-Sensitive Knowledge for Large Multimodal Models}

\author {
    Kailin Jiang* \textsuperscript{\rm1,2},
    Ning Jiang* \textsuperscript{\rm3},
    Yuntao Du \textsuperscript{\rm4,\rm5},
    Yuchen Ren \textsuperscript{\rm6}, 
    \textbf{Yuchen Li} \textsuperscript{\rm7}, 
    \textbf{Yifan Gao} \textsuperscript{\rm1}, \\
    \textbf{Jinhe Bi} \textsuperscript{\rm8}, 
    \textbf{Yunpu Ma} \textsuperscript{\rm8},
    \textbf{Bin Li} \textsuperscript{\rm1},
    \textbf{Lei Liu$\dagger$} \textsuperscript{\rm1},
    \textbf{Qing Li $\dagger$} \textsuperscript{\rm2} \\
    \textsuperscript{\rm1} University of Science and Technology of China\\
    \textsuperscript{\rm2} State Key Laboratory of General Artificial Intelligence, BIGAI\\
    \textsuperscript{\rm3} Northeast Forestry University,
    \textsuperscript{\rm4} C-FAIR\&school of software, Shandong University \\
    \textsuperscript{\rm5} State Key Lab. for Novel Software Technology, Nanjing University, P.R. China \\
    \textsuperscript{\rm6} The University of Sydney,
    \textsuperscript{\rm7} Anhui Polytechnic University,
    \textsuperscript{\rm8} Ludwig Maximilian University of Munich \\
    \texttt{kailinjiang@mail.ustc.edu.cn}
}

\begin{document}
\maketitle

\begin{abstract}
Large Multimodal Models (LMMs) encode rich factual knowledge via cross-modal pre-training, yet their static representations struggle to maintain an accurate understanding of time-sensitive knowledge. Existing benchmarks remain constrained by static designs, inadequately evaluating LMMs' ability to understand time-sensitive knowledge. To address this gap, we propose \dataset, a comprehensive benchmark containing 2,104 time-sensitive knowledge samples spanning six knowledge types, which evaluates temporal awareness along 6 key dimensions, including cognition, awareness, trustworthiness, understanding, reasoning, and robustness and 11 challenging tasks. Evaluating 15 widely used LMMs on \dataset shows that Gemini-2.5-Pro achieves the highest average CEM score of 63.07, while most open-source LMMs still lack time understanding ability. Meanwhile, LMMs perform best on organization knowledge, whereas their performance is weakest on sport. To address these challenges, we investigate the feasibility of updating time-sensitive knowledge in LMMs through knowledge editing methods and observe that LMMs can effectively update knowledge via knowledge editing methods in single editing scenarios.
\end{abstract}

\section{Introduction}
\label{sec:intro}

Large Multimodal Models demonstrate remarkable capabilities in general understanding and complex reasoning through large-scale pre-training, yet they face significant limitations due to their inherently static parameterized representations. While these models encode rich factual knowledge, their internal parameters often lag behind the continuous evolution of real-world facts. Consequently, they are prone to hallucinations or providing outdated outputs when handling queries that demand time-sensitive or up-to-date knowledge.

\begin{figure}[t]
  \centering

\includegraphics[width=0.6\linewidth]{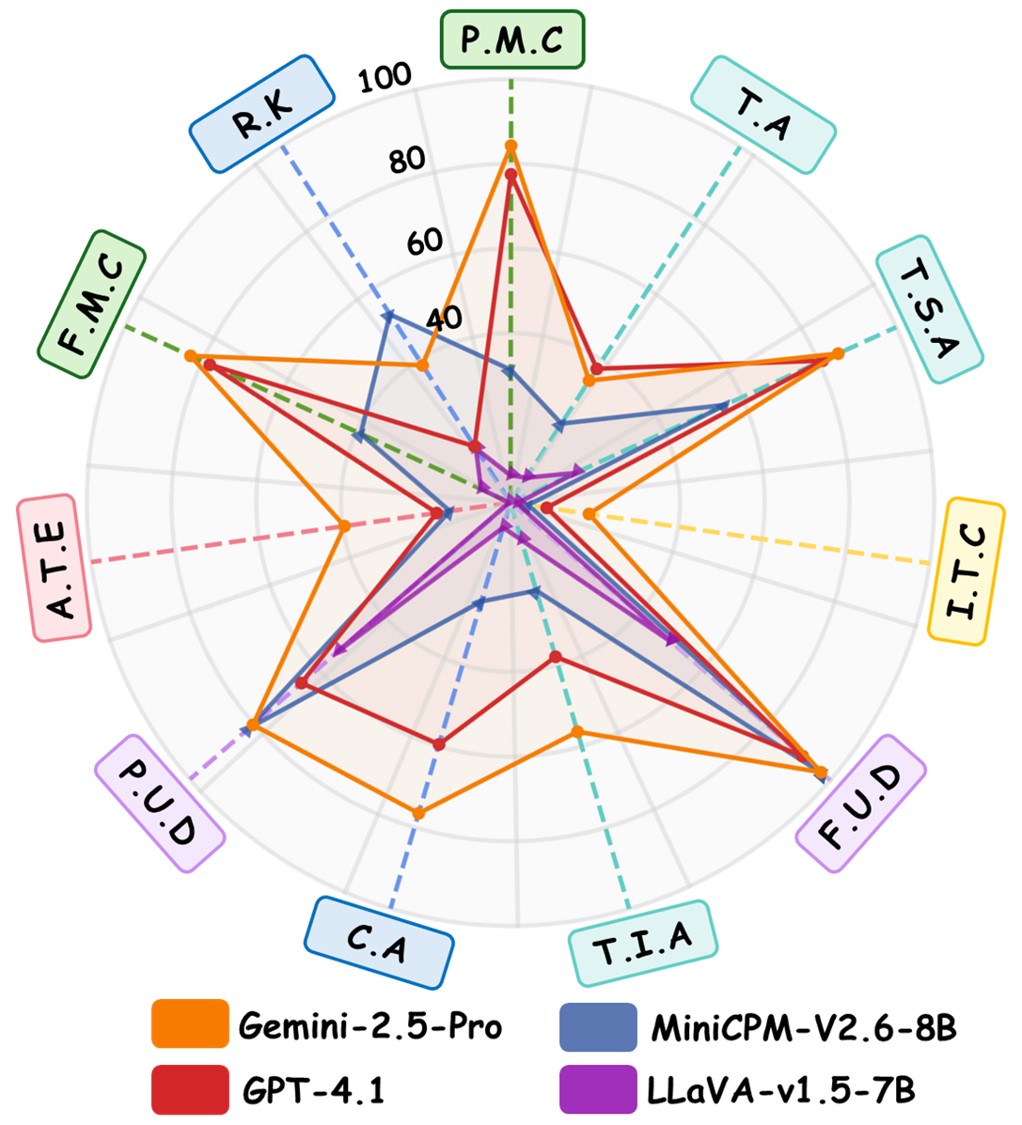}
  \caption{We evaluate temporal awareness of time-sensitive knowledge of SOTA LMMs across eleven challenging tasks.}
  \label{fig:overview_performance}
\end{figure}

To evaluate how models perceive and adapt to temporal awareness, researchers have primarily focused on benchmarks in the textual domain. Traditional datasets such as TimeQA \citep{chen2021dataset} and TempReason \citep{tan2023towards} assess basic time perception, yet a more profound challenge lies in whether models can effectively apply time-sensitive knowledge in continuously evolving scenarios. To capture this dynamic nature, recent studies have utilized dynamically updated knowledge bases \citep{kasai2023realtime}, rapid news streams \citep{zhang2024analyzing}, or real-time Wikipedia updates \citep{tang2025evowiki}. Notably, EvolveBench \citep{zhu2025evolvebench} further addresses real-world complexities, including temporal misalignment and outdated knowledge, assessing the temporal capabilities of LLMs through both cognitive and conscious dimensions.

While textual temporal reasoning has advanced, extending it to multimodal scenarios remains challenging due to cross-modal alignment complexities. Recent efforts like LiveVQA \citep{fu2025seeking} explore real-time visual knowledge updates but overlook critical practical issues such as temporal misalignment, and conflicting information. Consequently, current evaluations fail to fully capture the complexity of temporal reasoning in LMMs.


\begin{figure*}[t]
  \centering
\includegraphics[width=0.9\linewidth]{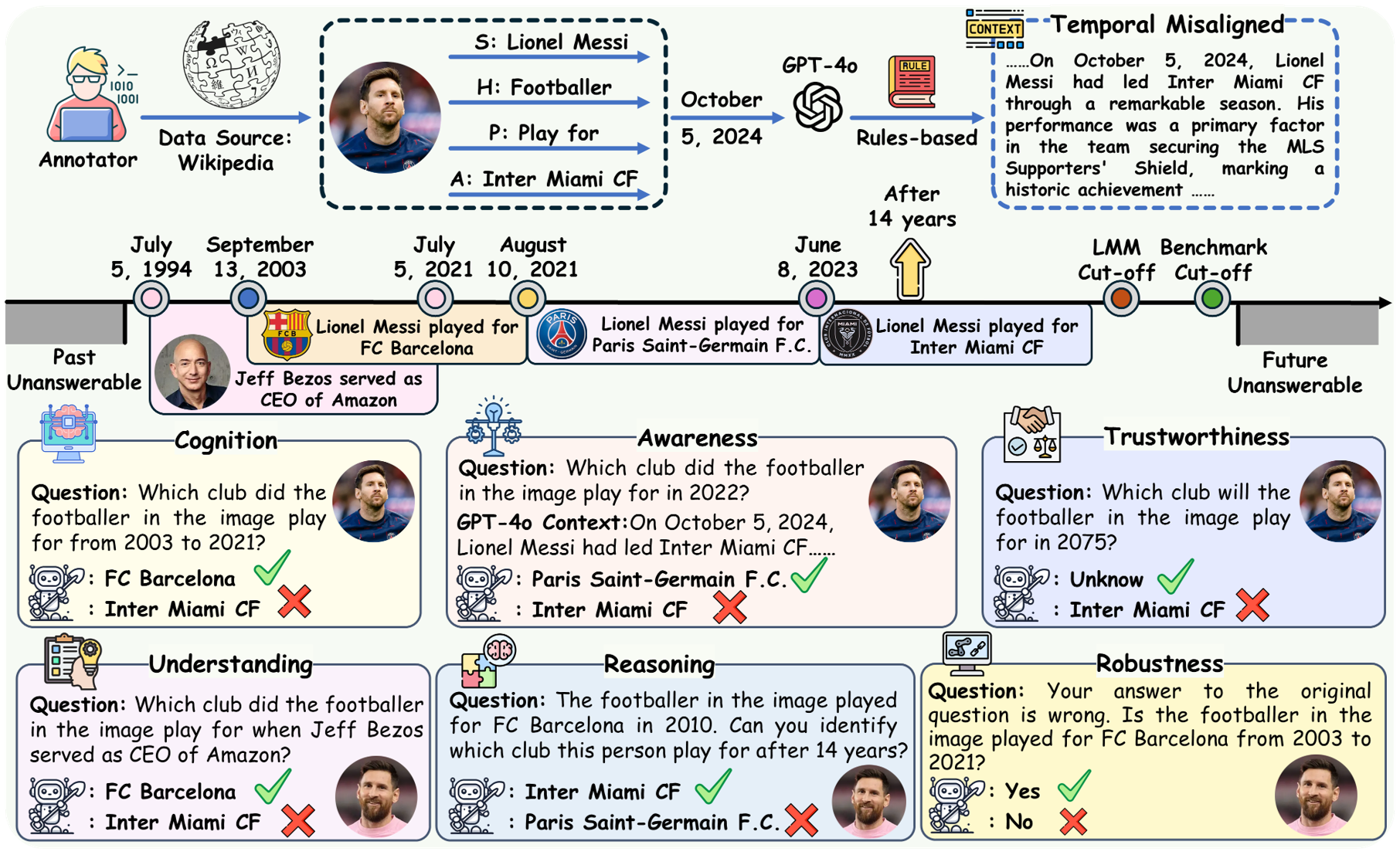}
 \vspace{-6pt}
  \caption{\textbf{Overview of the construction of \dataset.} To rigorously assess LMMs' temporal capabilities, we propose a six-dimensional evaluation framework tailored for time-sensitive knowledge.}
  \label{fig:overview}

\end{figure*}

To address this gap, we introduce \textbf{\dataset}, a novel benchmark designed to evaluate LMMs’ temporal awareness of time-sensitive knowledge across six key dimensions: \ding{182} \textbf{Cognition}, which measures a LMMs’ ability to recall and extract internal knowledge and apply it effectively; \ding{183} \textbf{Awareness}, which tests LMMs’ ability to detect temporal misalignment between an external context and user query; \ding{184} \textbf{Trustworthiness}, which evaluates the LMMs’ ability to identify and refuse to answer queries that contain invalid temporal information; \ding{185} \textbf{Understanding}, which examines the performance of LMMs when confronted with queries containing implicit temporal concepts; \ding{186} \textbf{Reasoning}, which evaluates the analytical ability of LMMs for temporal reasoning tasks; and \ding{187} \textbf{Robustness}, measuring the ability of LMMs to correct time comprehension errors. These dimensions collectively provide a holistic framework for assessing the temporal competence of LMMs.


We conduct extensive evaluations of 15 widely used LMMs on \dataset to assess their temporal understanding capabilities. Experimental results indicate that Gemini-2.5-Pro achieve the highest CEM score of 63.07. However, most open-source LMMs, such as LLaVA-v1.5 {\small(7B)} and Qwen-VL {\small(7B)}, still exhibit notable deficiencies in comprehending time-sensitive knowledge. These findings underscore the need for further improvements in time-sensitive knowledge understanding among existing LMMs. To address this challenge, we employ knowledge editing methods to update time-sensitive knowledge that LLaVA-v1.5 {\small(7B)} and Qwen-VL {\small(7B)} initially failed to answer. Results indicate that knowledge editing methods can effectively update time-sensitive knowledge in single editing scenarios.


\begin{itemize}[leftmargin=2.2em, itemsep=0pt] 
    \item We propose \dataset, a comprehensive benchmark designed to evaluate LMMs’ temporal awareness of time-sensitive knowledge. 
    \item We perform extensive experiments on 15 widely-used LMMs, the results reveal several limitations for current LMMs in handling temporal multimodal knowledge, establishing a foundation for further research on temporal understanding in multimodal systems.
    \item We explore the feasibility of knowledge editing methods for updating missing time-sensitive knowledge in LMMs, providing insights for enhancing temporal capabilities.
\end{itemize}

\section{Related Work}

\subsection{Large Multimodal Model}

LMMs have evolved from early contrastive models like CLIP \citep{radford2021learning} to systems supporting joint vision-language reasoning. Contemporary models such as LLaVA-v1.5 \citep{llava} and Qwen2.5-VL \citep{bai2025qwen25vl} integrate visual encoders with LLMs through unified alignment architectures. Furthermore, recent advancements in Gemini-2.5-Pro \citep{comanici2025gemini25} and Kimi-Latest \citep{kimiteam2025kimik15} have significantly enhanced reasoning and long-context capabilities through optimized decoding strategies.

\subsection{Temporal Reasoning Benchmarks}

Temporal reasoning involves inferring temporal expressions and logical relationships. While benchmarks like TimeQA \citep{chen2021dataset}, MenatQA \citep{wei2023menatqa}, TempReason \citep{tan2023towards}, and UnSeenTimeQA \citep{uddin2025unseentimeqa} evaluate contextual understanding in LLMs, they largely ignore time-sensitive knowledge. EvolveBench \citep{zhu2025evolvebench} addresses this gap by evaluating dynamic knowledge integration. In the multimodal domain, research remains scarce; LiveVQA \citep{fu2025seeking} and MMKU-Bench \citep{Fu2026MMKUBenchAM} evaluate real-time knowledge acquisition but overlook critical influence of time-sensitive knowledge.

Recognizing the limitations of existing benchmarks that primarily focus on textual reasoning and lack systematic multimodal evaluation, we introduce \dataset. This novel multi-dimensional benchmark addresses the gap by providing a comprehensive, fine-grained evaluations of LMMs' time-sensitive knowledge understanding. Table~\ref{tab:related_work_comparsion} presents the comparison with related works.

\begin{table}[t] 
    \centering
    \setlength{\tabcolsep}{2pt} 
    \renewcommand{\arraystretch}{1.2} 
    
    \resizebox{\linewidth}{!}{%
        \begin{tabular}{l c c c c c c c c}
            \toprule
            \textbf{Benchmark} & \textbf{Multi.} & \textbf{Cog.} & \textbf{Awa.} & \textbf{Tru.} & \textbf{Und.} & \textbf{Rea.} & \textbf{Rob.} & \textbf{P-Agr.} \\
            \midrule
            \textbf{TimeQA}       & \crossicon & \checkicon & \crossicon & \checkicon & \checkicon & \crossicon & \crossicon & \checkicon \\
            \textbf{MenatQA}      & \crossicon & \checkicon & \checkicon & \checkicon & \checkicon & \crossicon & \crossicon & \crossicon \\
            \textbf{TempReason}   & \crossicon & \checkicon & \crossicon & \crossicon & \checkicon & \crossicon & \crossicon & \crossicon \\
            \textbf{DyKnow}       & \crossicon & \checkicon & \crossicon & \crossicon & \crossicon & \crossicon & \crossicon & \checkicon \\
            \textbf{UnSeenTimeQA} & \crossicon & \crossicon & \crossicon & \crossicon & \crossicon & \checkicon & \crossicon & \crossicon \\
            \textbf{EvoWiki}      & \crossicon & \checkicon & \crossicon & \crossicon & \crossicon & \crossicon & \crossicon & \crossicon \\
            \textbf{EvolveBench}  & \crossicon & \checkicon & \checkicon & \checkicon & \checkicon & \checkicon & \crossicon & \checkicon \\
            \textbf{LiveVQA}      & \checkicon & \checkicon & \crossicon & \crossicon & \crossicon & \crossicon & \crossicon & \crossicon \\
            \midrule
            \textbf{MINED} (Ours) & \checkicon & \checkicon & \checkicon & \checkicon & \checkicon & \checkicon & \checkicon & \checkicon \\
            \bottomrule
        \end{tabular}%
    }
    \vspace{-5pt}
    \caption{Overall comparison with existing relevant benchmarks. P-Agr is Prompt Agreement (Section~\ref{sec:Setup}).}
    \label{tab:related_work_comparsion}
    
\end{table}

\section{Benchmark Construction}\label{sec:Benchmark_Construction}

In this section, we introduce the construction process of the \dataset benchmark. The dataset is categorized into original and task data, with the original data primarily collected by two professional annotators from Wikipedia across six domains: country, sport, company, university, organization, and competition. Further details in Appendix~\ref{appendix:benchmark_details}.

Based on the original data, we obtain a quadruple $(S, H, P, A)$ in Figure~\ref{fig:overview} to represent each time-sensitive knowledge, where $S$ is the subject (e.g., a person name like Lionel Messi), $H$ is the hypernym corresponding to the subject (e.g., Lionel Messi's hypernym is footballer), $P$ is the property (e.g., the property between Lionel Messi and club is ``play for''), and $A = [a_1, a_2, \cdots, a_n]$ is a list of attribute values for that property, which change over time. \textbf{Subsequent sections detail the conversion of quadruples into task data.}

\begin{figure*}[t!]
\centering

\begin{minipage}[t]{0.60\textwidth} 
    \vspace{0pt} 
    \centering
    
    \includegraphics[width=0.45\linewidth]{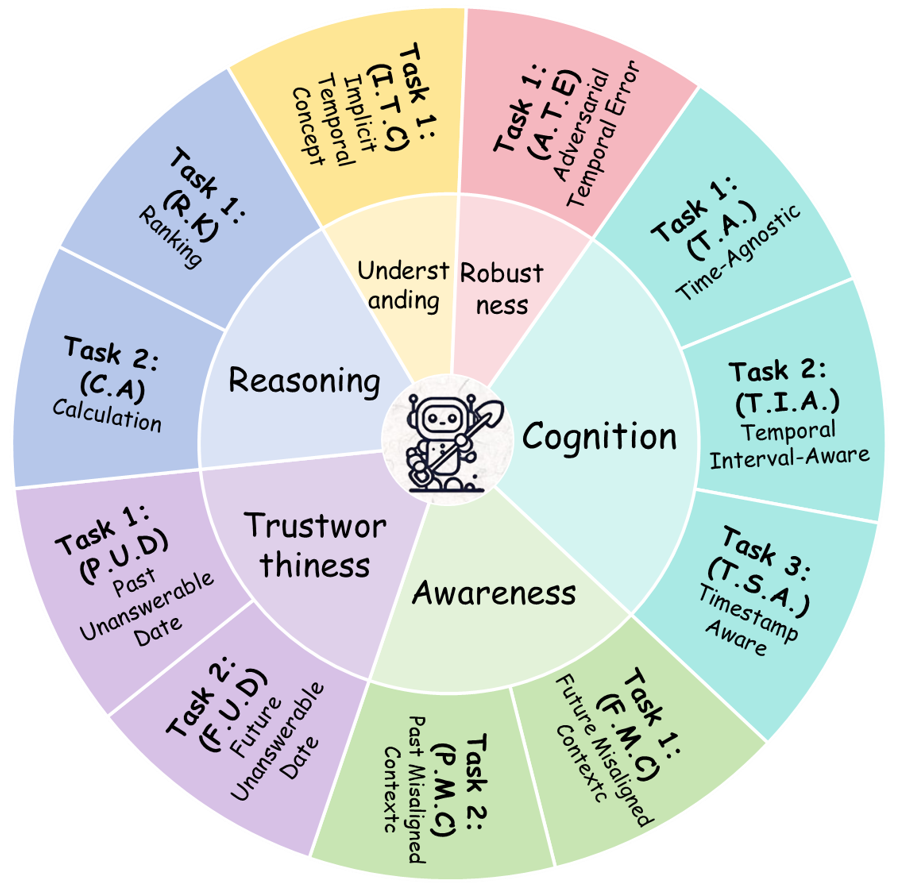} 
    \hfill 
    \includegraphics[width=0.45\linewidth]{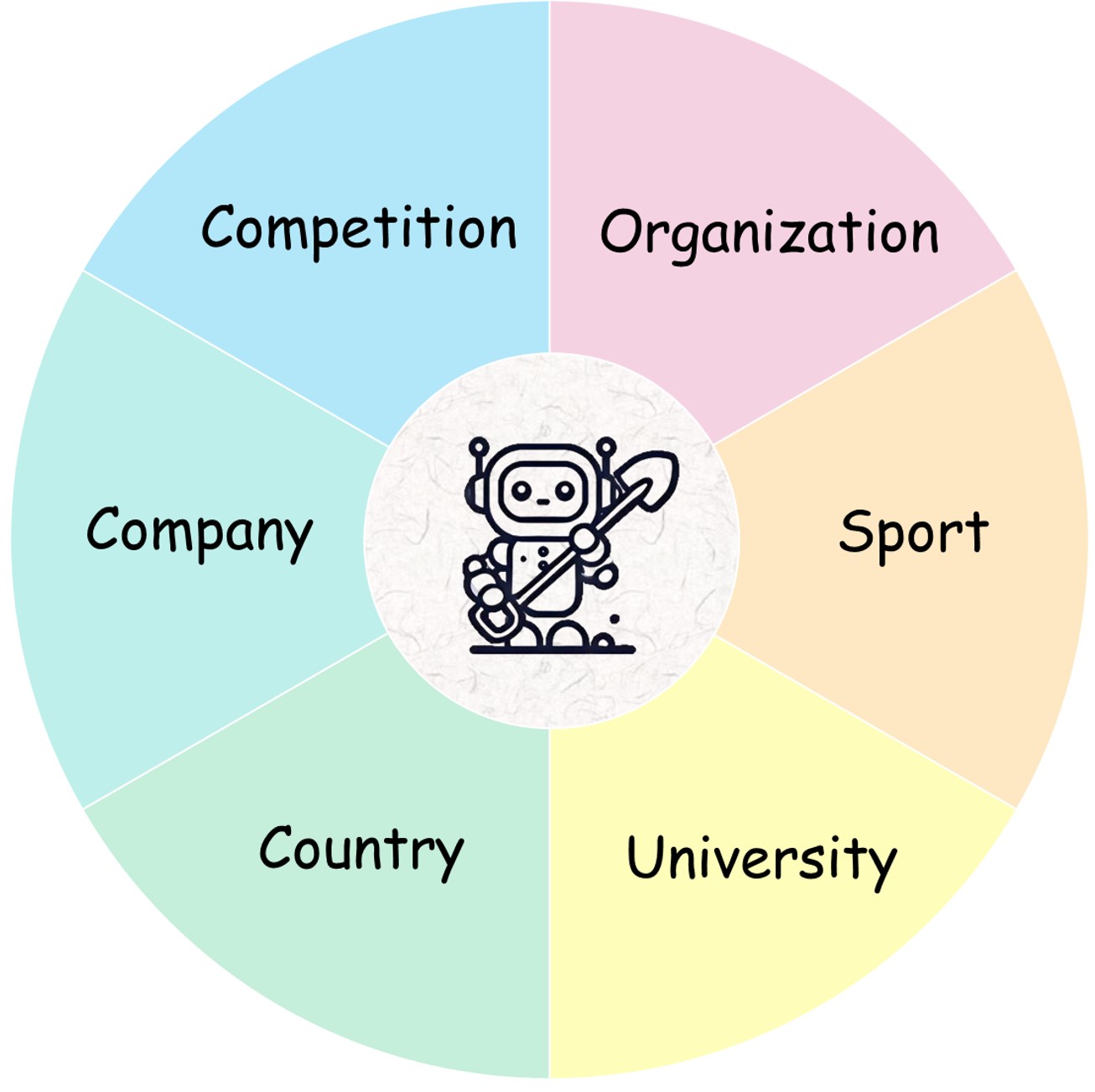} 
    
    \vspace{-5pt} 
    \caption{
    Fine-grained tasks (left) and knowledge (right) types.}
    \label{fig:combined_type}
\end{minipage}
\hfill 
\begin{minipage}[t]{0.32\textwidth} 
    \vspace{0pt} 
    \small
    \centering
    
    \begin{adjustbox}{width=\linewidth, valign=t}
        \begin{tabular}{lc}
        \toprule
        \textbf{Statistic} & \textbf{Number} \\
        \midrule
        Total questions & 4,208 \\
        \quad - Cognition questions & 1,328 (31.6\%) \\
        \quad - Awareness questions & 834 (19.8\%) \\
        \quad - Trustworthiness questions & 828 (19.7\%) \\
        \quad - Understanding questions & 510 (12.1\%) \\
        \quad - Reasoning questions & 324 (7.7\%) \\
        \quad - Robustness questions & 384 (8.1\%) \\
        Total dimension/subtasks & 6/11 \\
        Total fine-grained knowledge types & 6 \\
        Number of unique images & 450 \\
        \midrule
        Maximum question length & 54 \\
        Maximum answer length & 13 \\
        Average question length & 11.4 \\
        Average answer length & 2 \\
        \bottomrule
        \end{tabular}
    \end{adjustbox}

    \vspace{-5pt}
    \captionsetup{type=table} 
    \caption{Key Statistics of \dataset.}
    \label{table:statistics}
    
\end{minipage}
\vspace{-8pt}
\end{figure*}

\subsection{Cognition of Time-Sensitive Knowledge}\label{sec:Cognition}

We propose three cognitive tasks to evaluate the ability of LMMs to probe for time-sensitive knowledge. Given an image of entity $S$ and property $P$, the model must leverage its parameters to output the correct fact for queries.

\textbf{Time-Agnostic {\small(T.A)}} refers to using ``current'' or ``currently'' to prompt the model to provide the latest answer in $A$ without giving a clear time node. \textbf{Temporal Interval-Aware {\small(T.I.A)}} refers to randomly selecting a time period (from $T_{start}$ to $T_{end}$) from $A$ to prompt the model to provide the corresponding answer. \textbf{Timestamp-Aware {\small(T.S.A)}} refers to using random dates between $T_{start}$ and $T_{end}$ to prompt the model to provide corresponding answers.

\subsection{Awareness of Temporal Misalignment}\label{sec:Awareness}

Next, we evaluate how LMMs handle internal parametric knowledge when external context is temporal misaligned with timestamps in user queries. 

\textbf{Future Misaligned Context {\small(F.M.C)}:} We query a random past timestamp $T_{past}$, but provide a context $C_{current}$ generated by GPT-4o that describes the latest attribute $a_{current}$ for $(S, P)$. This creates a temporal conflict where the context information is accurate but futuristic relative to the query time $T_{past}$. \textbf{Past Misaligned Context {\small(P.M.C)}:} The query targets the current timestamp $T_{current}$. Conversely, we prompt GPT-4o with a past attribute $a_{past}$ to generate an outdated context $C_{past}$ describing $(S, P, a_{past})$. This evaluates the model's robustness against obsolete information in the context.

\subsection{Trustworthiness of Unanswerable Date}\label{sec:Trustworthiness}

We introduce credibility to assess LMM's hallucinations on unanswerable queries. A query is deemed unanswerable if the timestamp $T$ falls outside the valid time range (\ie, before the earliest or after the latest record) defined in $A$ for the target $(S, P)$.

\textbf{Past Unanswerable Date {\small(P.U.D)}:} We extract the earliest record from attribute list $A$ and subtract a certain year from it to construct an unanswerable date in the past. In Figure~\ref{fig:overview}, Lionel Messi had not started his professional career before 2003, so we select a time point prior to that year as the past unanswerable date. \textbf{Future Unanswerable Date {\small(F.U.D)}:} We take the latest record from $A$ and add a certain year to construct an unanswerable future date. For example, ``Which club will the footlocker in the image play for in 2075?" in Figure~\ref{fig:overview}.

\subsection{Understanding of Temporal Concept}\label{sec:Understanding}

This dimension evaluates how effectively LMMs understand temporal concepts expressed in different formats. In previous evaluations, explicit time formats (e.g., \en{DD Month YYYY}) were used to denote temporal information. For implicit temporal expressions, temporal intervals $[T_{start}, T_{end}]$ are defined based on historical events.

\textbf{Implicit Temporal Concept {\small(I.T.C)}:} In Figure~\ref{fig:overview}, the phrase \en{when Jeff Bezos served as CEO of Amazon} corresponds to the period from July 5, 1994, to July 5, 2021. Such implicit temporal representations are denoted as $T_{implicit}$.

\subsection{Temporal Reasoning}\label{sec:Reasoning}

We propose two tasks to evaluate temporal reasoning in LMMs: a ranking task for chronological ordering to assess temporal logic, and a calculation task involving time intervals and durations to measure numerical precision.

\textbf{Ranking {\small(R.K)}:} Two past events $a_{1}$ and $a_{2}$ are randomly selected from attribute list $A$. The model first recalls the respective time periods for $a_{1}$ and $a_{2}$ based on the input and compares them to determine their correct chronological order. \textbf{Calculation {\small(C.A)}:} For events $a_{1}$ and $a_{2}$, dates $t_{1}$ and $t_{2}$ is randomly selected from their respective time intervals $[T_{start}, T_{end}]$, and the number of days between them, denoted as $T_{\Delta}$, is calculated. Given $t_{1}$ and $T_{\Delta}$, the task requires the model to perform the necessary computation and infer the correct date corresponding to the target event $a_{2}$.

\subsection{Robustness of Time-Sensitive Knowledge}\label{sec:Robustness}

Robustness evaluates the model's capacity to identify and self-correct errors when provided with appropriate prompts.

\textbf{Adversarial Temporal Error {\small(A.T.E)}:} We extract knowledge samples for which all LMMs provided incorrect answers across three cognitive subtasks. Using the prompt: \en{Your answer to the original question is wrong} followed by a rephrased interrogative form, we examine whether the models can correct their previous errors.

\subsection{Benchmark Analysis}

\textbf{Category Distribution and Key Statistics:} \dataset comprises 4,208 questions across 6 key dimensions and 6 fine-grained categories, demonstrating its diversity (Table~\ref{table:statistics}, Figure~\ref{fig:combined_type}). Regarding \dataset's details, construction pipeline, experiment resources, chat templates and case studies, please refer to Appendices~\ref{appendix:benchmark_details},~\ref{appendix:Experiment_Resources},~\ref{appendix:Case} and~\ref{appendix:1_chat_templates_quantitative_examples}.


\section{\label{sec:probing}Experiment}

\subsection{\label{sec:Setup} Experimental Setup}

\noindent \textbf{Large Multimodal Models.} In this paper, we evaluate 15 widely used LMMs on \dataset, including: LLaVA-v1.5 \citep{llava}, Qwen-VL \citep{qwen_vl}, mPLUG-Owl2 \citep{ye2023mplugowl2}, LLaVA-Next \citep{liu2024llavanext}, LLaVA-OneVision \citep{li2024llavaonevision}, mPlug-Owl3 \citep{ye2024mplugowl3}, MiniCPM-V2.6 \citep{yao2024minicpm}, Qwen2-VL \citep{wang2024qwen2vl}, InternVL2.5 \citep{internvl2_5}, Qwen2.5-VL \citep{bai2025qwen25vl}, GPT-4.1 \citep{openai2023gpt4}, Kimi-Latest \citep{kimiteam2025kimik15}, Doubao-1.5-Vision-Pro, Gemini-2.5-Pro \citep{comanici2025gemini25}, Seed-1.6-Vision.

\begin{table*}[t!]
  \centering

  \vspace{-5pt}
  \renewcommand{\arraystretch}{1.1}  
  \resizebox{\textwidth}{!}{%
    \begin{tabular}{l|c c c|c c|c c|c|c c|c|c}
      \toprule
      \multirow{2.5}{*}{\textbf{{(Release Time)} Models}} 
        & \multicolumn{3}{c|}{\textbf{Cog.}} 
        & \multicolumn{2}{c|}{\textbf{Awa.}} 
        & \multicolumn{2}{c|}{\textbf{Tru.}} 
        & \multicolumn{1}{c|}{\textbf{Und.}} 
        & \multicolumn{2}{c|}{\textbf{Rea.}} 
        & \multicolumn{1}{c|}{\textbf{Rob.}} 
        & \multirow{2.5}{*}{\textbf{Avg.}} \\
      \cmidrule{2-12}
        & \textbf{T.A} \daugshifted & \textbf{T.I.A}\daugshifted  & \textbf{T.S.A} \daugshifted
        & \textbf{F.M.C}  \daugshifted & \textbf{P.M.C} \daugshifted
        & \textbf{P.U.D} \daugshifted & \textbf{F.U.D}  \daugshifted
        & \textbf{I.T.C}  \daugshifted  
        & \textbf{R.K} \daugshifted & \textbf{C.A}  \daugshifted
        & \textbf{A.T.E} \daugshifted \\

      \midrule
      \rowcolor{gray!10}
      \multicolumn{13}{c}{\fontsize{10}{12}\selectfont \textit{\textbf{Open-source LMMs}}} \\	


      {\small(2023.04)} LLaVA-v1.5 {\small(7B)}  
      & \colorbox{backblue!75}{6.96}  & \colorbox{backblue!75}{9.25}  & \colorbox{backblue!75}{16.88} & 7.66  & \colorbox{backblue!75}{6.40}  & 53.99 & 50.00  & \colorbox{backblue!75}{1.57} & \colorbox{backblue!75}{15.12} & \colorbox{backblue!75}{6.17}  & 0.39  & \colorbox{backblue!75}{15.85} \\
      {\small(2023.08)} Qwen-VL {\small(7B)}     
      & 12.45 & 17.30 & 42.09 & \colorbox{backblue!75}{6.04}  & 6.91  & 81.28 & 70.17  & 3.53 & 25.00 & 17.59 & \colorbox{backblue!75}{0.00}  & 25.67 \\
      {\small(2023.11)} mPLUG-Owl2 {\small(7B)} 
      & 10.59 & 14.53 & 44.62 & 42.69 & 38.67 & \colorbox{backblue!75}{11.47} & \colorbox{backblue!75}{44.20}  & 2.16 & 42.90 & 14.20 & 6.12  & 24.74 \\
      {\small(2024.01)}  LLaVA-Next$_{M.}$ {\small(7B)} 
      & 10.69 & 14.53 & 41.14 & 33.69 & 28.87 & 96.74 & 90.22  & 3.73 & 38.58 & 20.99 & \colorbox{backblue!75}{0.00}  & 34.47 \\
      {\small(2024.08)} LLaVA-OV {\small(7B)}   
      & 11.86 & 11.34 & 26.79 & 30.93 & 31.35 & 39.61 & 76.21  & 3.63 & 51.54 & 8.95  & 2.21  & 26.77 \\
      {\small(2024.08)} mPlug-Owl3 {\small(8B)}  
      & 9.80  & 10.03 & 29.01 & 29.77 & 28.31 & 97.95 & 99.76  & 3.14 & 41.98 & 7.10  & 3.65  & 32.77 \\
      {\small(2024.08)} MiniCPM-V2.6 {\small(8B)} 
      & 22.16 & 21.66 & 55.70 & 38.88 & 31.35 & 81.52 & 97.83  & 4.22 & \colorbox{backyellow_soft!40}{52.78} & 24.38 & 14.45 & 40.45 \\
      {\small(2024.09)} Qwen2-VL$_{I.}$ {\small(7B)}  
      & 15.98 & 16.72 & 31.96 & 17.90 & 11.46 & \colorbox{backyellow_soft!40}{99.52} & 99.76  & 4.61 & 49.38 & 14.20 & 9.90  & 33.76 \\
      {\small(2024.12)} InternVL2.5 {\small(8B)} 
      & 20.49 & 18.46 & 44.83 & 42.37 & 38.26 & 98.31 & \colorbox{backyellow_soft!40}{99.88}  & 4.22 & \colorbox{backred!50}{61.73} & 19.14 & \colorbox{backblue!75}{0.00}  & 40.70 \\
      {\small(2025.02)} Qwen2.5-VL$_{I.}$ {\small(7B)} 
      & 18.33 & 16.86 & 41.67 & 40.04 & 33.98 & \colorbox{backred!50}{99.64} & 99.76  & 4.02 & 38.89 & 25.00 & 16.86 & 39.55 \\

    \midrule
    \rowcolor{gray!10}
      \multicolumn{13}{c}{\fontsize{10}{12}\selectfont \textit{\textbf{Closed-source LMMs}}} \\	


      {\small(2025.02)} Kimi-Latest 
      & 26.41 & 26.60 & 72.43 & 68.64 & 67.27 & 72.10 & 85.39  & 7.06 & 45.99 & 42.59 & 6.38  & 47.35 \\
      {\small(2025.02)} Doubao-1.5-Vision-Pro 
      & 35.78 & 27.91 & 69.83 & 74.36 & 70.76 &  93.12 & \colorbox{backred!50}{100.00} & 5.29 & 18.52 & 34.57 & 12.24 & 49.31  \\
      {\small(2025.03)} Gemini-2.5-Pro & 34.25 & \colorbox{backred!50}{56.40} & \colorbox{backred!50}{84.96} & \colorbox{backred!50}{83.09} & \colorbox{backred!50}{84.30} & 80.31 & 97.10 & \colorbox{backred!50}{18.73} & 38.48 & \colorbox{backred!50}{76.54} & \colorbox{backred!50}{39.58} & \colorbox{backred!50}{63.07} \\
      {\small(2025.04)} GPT-4.1  & \colorbox{backred!50}{37.58} & 37.94 & \colorbox{backyellow_soft!40}{80.91} & \colorbox{backyellow_soft!40}{78.07} & 77.49 & 65.22 & 91.30 & \colorbox{backyellow_soft!40}{8.63} & 15.74 & \colorbox{backyellow_soft!40}{59.57} & 17.58 &  51.82 \\
      {\small(2025.08)} Seed-1.6-Vision 
      & \colorbox{backyellow_soft!40}{37.19} & \colorbox{backyellow_soft!40}{41.76} & 78.69 & 75.95 & \colorbox{backyellow_soft!40}{80.71} & 74.15 & 96.86  & 7.55 & 21.60 & \colorbox{backyellow_soft!40}{59.57} & \colorbox{backyellow_soft!40}{32.68} &  \colorbox{backyellow_soft!40}{55.16} \\

      \bottomrule
    \end{tabular}%
  }

  \vspace{-5pt}
  \caption{\textbf{Overall Performance Comparison ($\%$) on \dataset.} 
The top two and worst performing results are highlighted in 
\colorbox{backred!50}{red} (1\textsuperscript{st}), 
 \colorbox{backyellow_soft!40}{yellow} (2\textsuperscript{nd}) and \colorbox{backblue!75}{blue} (bottom) backgrounds, respectively. Subscripts ${M.}$ and ${I.}$ stand for Mistral-7B and Instruct, respectively.}
 
  \label{table:main_results_cem}
\vspace{-15pt}
\end{table*}

\noindent \textbf{Evaluation Protocol:} In the evaluation of all subtasks, the model is considered to have correctly responded to the time-sensitive knowledge only when its output exactly matches the corresponding ground truth. Therefore, we evaluate the model’s outputs using Cover Exact Match (CEM) \citep{xu2023lvlm} score for each subtask. The model’s capacity in this dimension is defined as the average CEM score across all subtasks.

\vspace{-16pt}

\begin{equation}
C_d = \frac{1}{N} \sum_{i=1}^{N} CEM_i, \hspace{0.2em} CEM = \mathbb{I}(\hat{y} \subseteq {Y})
\end{equation}

Where $N$ is the subtask count in dimension $d$, $CEM_i$ is the score of the $i$-th subtask, and ${Y}$ and $\hat{y}$ denote the model prediction and ground truth.

\noindent \textbf{Prompt Agreement:} To mitigate uncertainty from prompt variations, we design four semantically equivalent prompts (``Question'', ``Generalization Question'',``Image'', and ``Generalization Image'') for each knowledge instance. Final score is computed by averaging the scores across these configurations, a strategy termed \en{Prompt Agreement}.

\subsection{Analysis of Main Results\label{sec:main_result}}

Table~\ref{table:main_results_cem} summarizes the performance of 15 LMMs on \dataset, with additional results provided in Appendix~\ref{appendix:more_results}. From these results, we observe:

\nbi{Obs 1: LMMs exhibit improved cognitive performance when queries are framed as timestamp-aware task.} When evaluating the cognitive capacities of LMMs, we present queries conveying identical knowledge in three distinct temporal formats: Time-Agnostic, Temporal Interval-Aware, and Timestamp-Aware. For the knowledge ``Lionel Messi played for Inter Miami CF'', Time-Agnostic, Temporal Interval-Aware, and Timestamp-Aware queries are formulated as follows: ``Which club does the person in the image currently play for?'', ``Which club did the footballer play for between 2023 and 2024?'', and ``Which club did the footballer play for on 1 January 2024?'', respectively. 

Table~\ref{table:main_results_cem} indicates that LMMs perform best on Timestamp-Aware tasks. This is likely attributed to the narrower scope of retrieving specific point-in-time knowledge, compared to the more challenging broad temporal contexts required by Time-Agnostic and Interval-Aware queries. However, the top-performing Gemini-2.5-Pro still fails to recall approximately 15\% of the knowledge, underscoring the persistent challenge of temporal sensitivity.

\begin{figure}[t!] 
  \centering
  \includegraphics[width=1\linewidth]{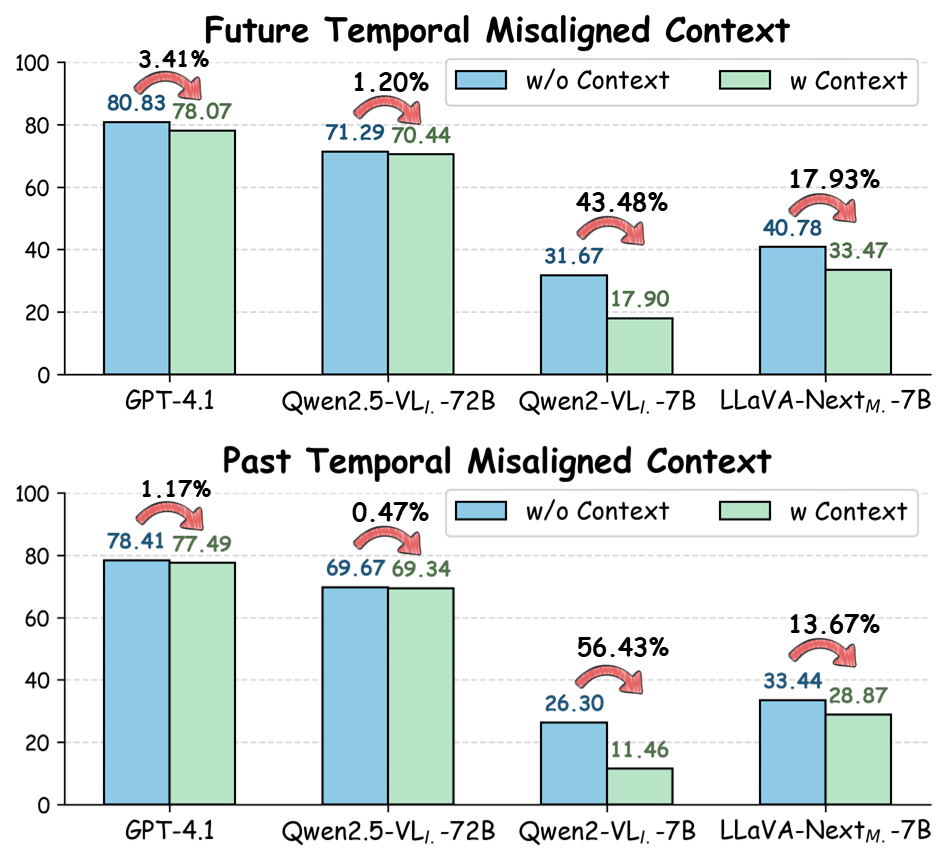}
  \vspace{-22pt}
        \caption{Comparison of performance with and without misaligned context.}
        \label{fig:awareness}
\end{figure}

\nbi{Obs 2: LMMs are vulnerable to temporal misaligned context, especially from past temporal misaligned contexts.} Compared to T.S.A. results, LMMs' performance degrades when queries are accompanied by temporal misaligned context, which impedes correct knowledge recall. In Figure~\ref{fig:awareness}, we use the same timestamp in the queries, with only difference being whether the input query included the relevant but temporal misaligned text. We observe that closed-source models and larger open-source models exhibit greater robustness to temporally misaligned context, whereas smaller open-source models suffer significant performance degradation. For instance, Qwen2-VL$_{I.}$ {\small(7B)} shows declines of 43.84\% on F.M.C and 56.43\% on P.M.C. These results suggest that smaller models are more susceptible to misleading temporal contexts, especially those involving past misalignment.


\begin{figure}[t!] 
  \centering
  \includegraphics[width=1.0\linewidth]{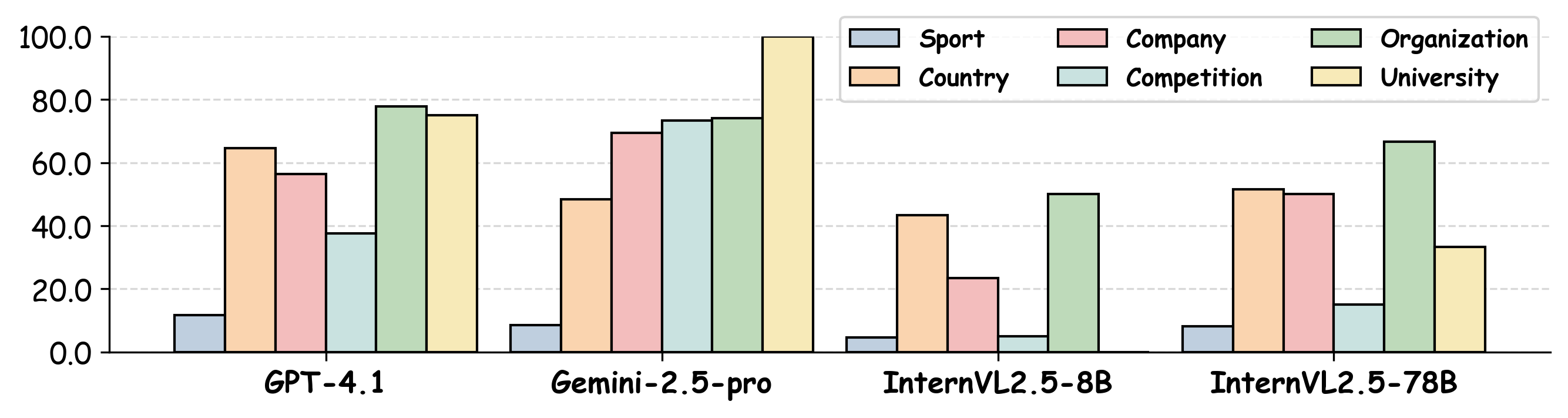}
  \vspace{-20pt}
  \caption{The cognitive capacity of various LMMs across six specific knowledge types.}
  \label{fig:knowledge_type_score}
\end{figure}

\nbi{Obs 3: LMMs are better at rejecting questions with unanswerable future dates than those with past dates.} As indicated by P.U.D and F.U.D results in Table~\ref{table:main_results_cem}, most LMMs are capable of effectively rejecting questions that contain unanswerable dates from either the past or the future. This is likely because such dates are absent from the training data, allowing the models to reject them with greater confidence. Furthermore, LMMs show a slightly stronger propensity to reject questions with unanswerable future dates, likely because these represent entirely unseen temporal concepts, resulting in even greater refusal certainty. Surprisingly, both Qwen2-VL$_{I.}$ {\small(7B)} (average CEM score of 99.64) and Qwen2.5-VL$_{I.}$ {\small(7B)} (average CEM score of 99.70) demonstrate exceptional performance in question refusal, a capability potentially attributable to enhanced defensive mechanisms from their instruction tuning process.

\nbi{Obs 4: All LLMs perform terribly on tasks involving implicit temporal concepts.} In the I.T.C column of Table~\ref{table:main_results_cem}, all LLMs perform terribly, with even the top-performing model, Gemini-2.5-Pro, recalling less than 20\% of relevant knowledge. This indicates a fundamental deficiency in understanding and utilizing implicit temporal concepts.

\begin{figure}[t!] 
  \centering
  \includegraphics[width=1\linewidth]{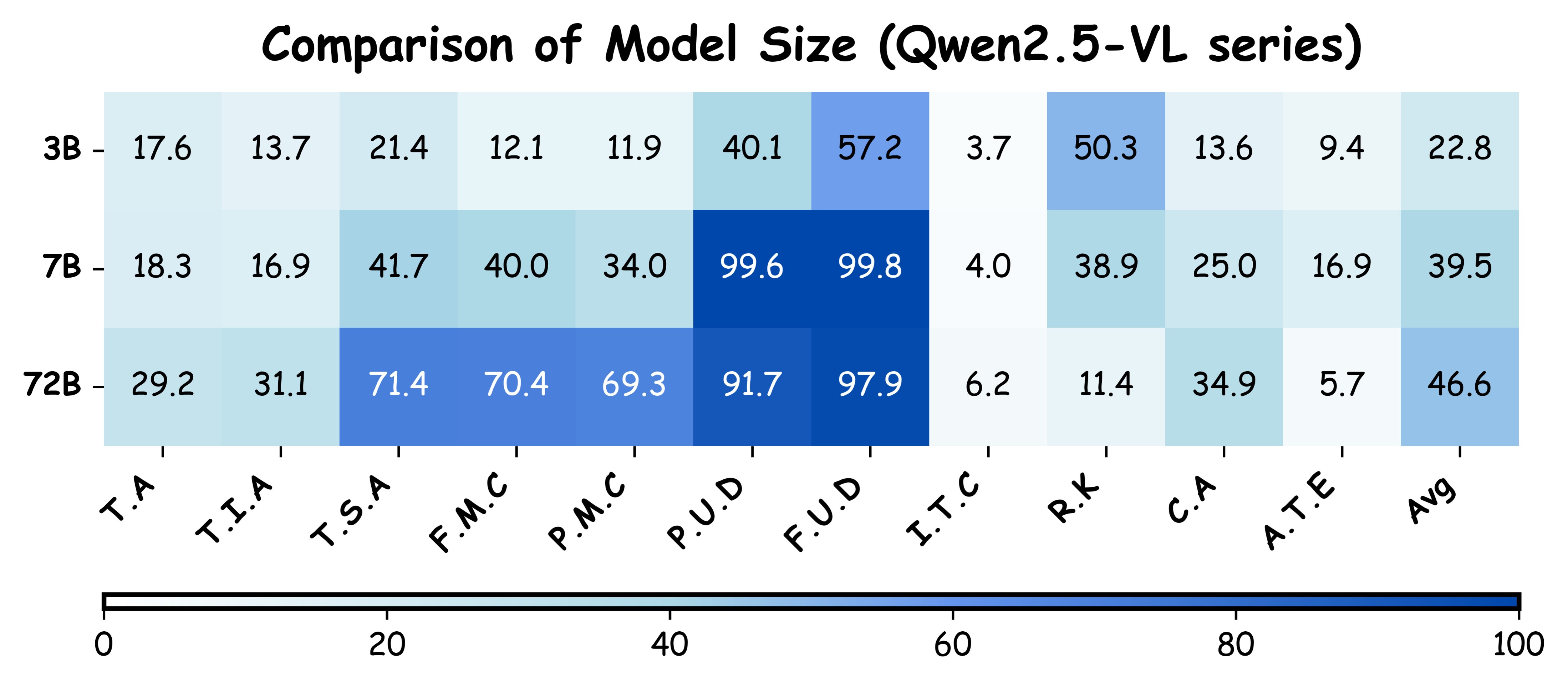}
  \vspace{-20pt}
  \caption{Analysis of impact of different model sizes.}
  \label{fig:model_size_base_llm}
\end{figure}

\nbi{Obs 5: Open-source LMMs demonstrate stronger performance on simpler ranking task, whereas closed-source LMMs excel in more complex calculation task.} Unexpectedly, MiniCPM-V2.6 {\small(8B)} and InternVL2.5 {\small(8B)} achieved the highest performance on ranking task, while models such as GPT-4.1 and Doubao-1.5-Vision-Pro scored below 20\% in CEM. Figure~\ref{fig:model_size_base_llm} further illustrates this phenomenon, showing a decline in ranking performance within the Qwen2.5-VL$_{I.}$ series as model size increases $50.3_{(3B)} \rightarrow 38.9_{(7B)} \rightarrow 11.4_{(72B)}$, potentially due to overthinking. Larger models, despite their enhanced reasoning capabilities, may overcomplicate simple tasks like ranking, leading to reduced effectiveness. In contrast, on more challenging calculation task, closed-source LMMs including Gemini-2.5-Pro and GPT-4.1 demonstrated superior performance.


\nbi{Obs 6: Current LMMs demonstrate limited adversarial robustness against temporal errors.} According to the A.T.E results in Table~\ref{table:main_results_cem}, models such as Qwen-VL {\small(7B)}, LLaVA-Next$_{M.}$ {\small(7B)}, and InternVL2.5 {\small(8B)} fail to correct any prior errors, demonstrating severely limited robustness. Even the top-performing model, Gemini-2.5-Pro, corrects fewer than 40\% of errors. These results indicate a significant need for improvement in temporal reasoning robustness across current models.

\nbi{Obs 7: More recent LMMs exhibit better temporal awareness performance.} Avg. results in Table~\ref{table:main_results_cem} reveal an approximate trend: more recent LMMs generally achieve superior overall performance, indicating a link between temporal awareness and recency of development.

\subsection{\label{sec:Exploratory}Analysis of Exploratory Results}

In this section, we present further explorations into evaluation of time-sensitive knowledge, yielding the following observations.

\nbi{Exploration 1: Fine-grained Knowledge Types.} All LMMs show consistent trends in recalling time-sensitive knowledge across domains. As shown in Figure~\ref{fig:knowledge_type_score}, LMMs perform better on queries related to organization, company, and country leaders, but worse on athletes and competition champions,likely due to the broader coverage of the former in public knowledge sources. Furthermore, closed-source models outperform open-source variants on university president queries, indicating potential discrepancies in their pretraining corpora.

\begin{figure*}[t!]
    \centering

    \begin{minipage}[t]{0.34\textwidth}
        \vspace{0pt} 
        \centering
        \renewcommand{\arraystretch}{1.13}
        
        \resizebox{\linewidth}{!}{%
            \begin{tabular}{l|c c c}
                \toprule
                \multirow{2.5}{*}{\textbf{Model}}
                & \multicolumn{3}{c}{\textbf{Time-Agnostic}} \\
                \cmidrule{2-4}
                & \textbf{Lat.} \daugshifted & \textbf{Out.} \dalgshifted & \textbf{Irr.}\dalgshifted \\
                \midrule
                \rowcolor{gray!10}
                \multicolumn{4}{c}{\fontsize{10}{12}\selectfont \textit{\textbf{Open-source LMMs}}} \\    
                LLaVA-v1.5 {\small(7B)} & 14.90 & 27.45 & 57.65 \\
                LLaVA-Next$_{M.}$ {\small(7B)} & 19.22 & 36.47 & 44.31 \\
                InternVL2.5 {\small(1B)} & 14.12 & 33.73 & 44.31 \\
                InternVL2.5 {\small(8B)} & 16.08 & 43.92 & 40.00 \\
                Qwen2.5-VL$_{I.}$ {\small(7B)} & 20.00 & 56.86 & 23.14\\
                \midrule
                \rowcolor{gray!10}
                \multicolumn{4}{c}{\fontsize{10}{12}\selectfont \textit{\textbf{Closed-source LMMs}}} \\    
                Kimi-Latest & 24.71 & 58.82 & 16.47 \\
                GPT-4.1 & 28.04 & 53.53 & 18.43 \\
                Seed-1.6-Vision & 21.57 & 64.31 & 14.12 \\
                \bottomrule
            \end{tabular}%
        }
        \vspace{-3pt}
        \captionsetup{type=table} 
        \caption{Fine-grained analysis of predicted output in Time-Agnostic task.}
        \label{tab:time_agnostic_analysis}
        
    \end{minipage}
    \hfill 
    \begin{minipage}[t]{0.61\textwidth}
        \vspace{0pt} 
        \centering
        
        \includegraphics[width=1\linewidth, keepaspectratio]{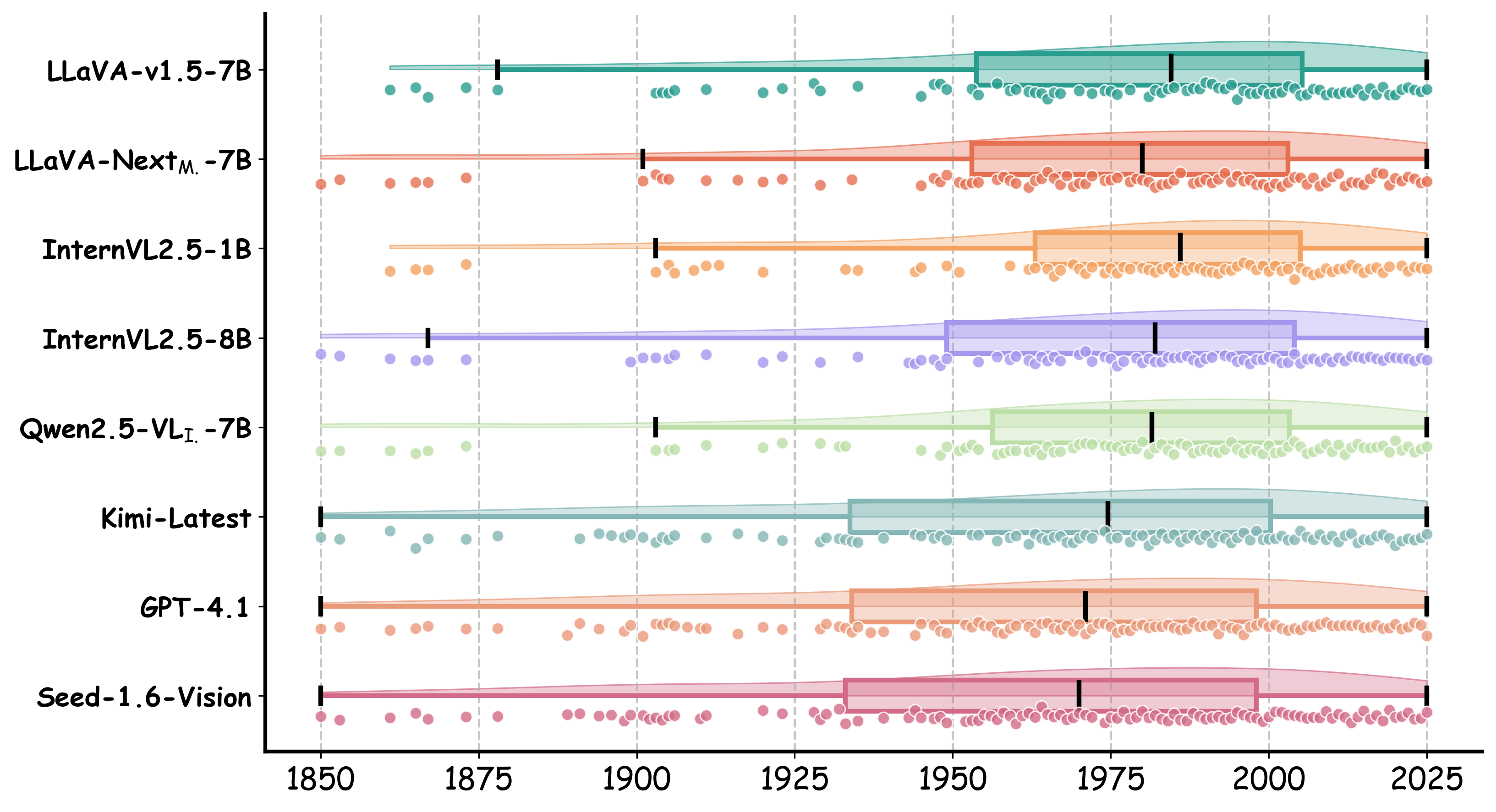}
        
        \captionsetup{skip=5pt} 
        \caption{Approximating temporal distribution of internal knowledge of LMMs.}
        \label{fig:time_distribution}
    \end{minipage}
    
\vspace{-8pt}

\end{figure*}

\nbi{Exploration 2: Model Size.} Figure~\ref{fig:model_size_base_llm} illustrates that increasing model size generally improves performance across most tasks, with notable exceptions in R.K, P.U.D, F.U.D, and A.T.E.

\newtcbox{\rebox}{on line, box align=base, colback=red!10, colframe=white, size=fbox, arc=3pt,
before upper=\strut, top=-2pt, bottom=-4pt, left=-2pt, right=-2pt, boxrule=0pt} 

\newtcbox{\grbox}{on line, box align=base, colback=green!10, colframe=white, size=fbox, arc=3pt, 
before upper=\strut, top=-2pt, bottom=-4pt, left=-2pt, right=-2pt, boxrule=0pt}

\nbi{Exploration 3: Fine-grained Analysis of Time-Agnostic and Temporal Distribution.} In the Time-Agnostic task, we further categorize the model’s outputs into fine-grained labels. Since Prompt Agreement is adopted, each knowledge yields four outputs. If any output contains the most up-to-date value from the attribute list $A$, it is labeled as \textbf{Latest}. If none includes the latest value but at least one contains an outdated answer, it is marked as \textbf{Outdated}. All other cases are categorized as \textbf{Irrelevant}. In Table~\ref{tab:time_agnostic_analysis}, open-source models not only produce a limited number of latest responses but also generate a substantial portion of irrelevant responses. In contrast, closed-source models reduce the frequency of irrelevant responses but still exhibit a high proportion of outdated responses. These statistical results indicate that a significant portion of model-generated responses are either outdated or irrelevant, highlighting a pronounced issue of inaccurate time-sensitive knowledge. Figure~\ref{fig:time_distribution} provides an approximate visualization of the temporal distribution of knowledge within LMMs. Closed-source models demonstrate a broader temporal coverage. In contrast, the internal knowledge of open-source models is concentrated in more recent time periods, indicating a comparative difficulty in recalling information from distant historical contexts.

\begin{table}[t!] 
    \centering
    \renewcommand{\arraystretch}{1.25}
    
    \resizebox{\linewidth}{!}{%
        \begin{tabular}{l|l l l|l l l}
            \toprule
            \multirow{2.5}{*}{\textbf{Model}}
            & \multicolumn{3}{c|}{\textbf{Future Misaligned Context}} & \multicolumn{3}{c}{\textbf{Past Misaligned Context}} \\
            \cmidrule{2-4} \cmidrule{5-7}
            & \textbf{Con.} \dalgshifted & \textbf{Oth.} \dalgshifted & \textbf{Irr.}\dalgshifted & \textbf{Con.} \dalgshifted & \textbf{Oth.} \dalgshifted & \textbf{Irr.}\dalgshifted \\
            \midrule

            \rowcolor{gray!10}
            \multicolumn{7}{c}{\fontsize{10}{12}\selectfont \textit{\textbf{w/ Misaligned Context}}} \\
            GPT-4.1 & 7.9 & 5.6 & 8.4 & 10.6 & 4.8 & 7.0 \\
            Qwen2-VL$_{I.}$ {\small(7B)} & 64.7 & 5.9 & 11.4 & 77.2 & 4.4 & 6.9 \\
            LLaVA-Next$_{M.}$ {\small(7B)} & 52.4 & 5.0 & 9.1 & 57.5 & 5.4 & 8.3 \\
            Qwen2.5-VL$_{I.}$ {\small(72B)} & 8.8 & 8.2 & 12.6 & 12.2 & 8.0 & 10.5 \\
            \midrule
            \rowcolor{gray!10}
            \multicolumn{7}{c}{\fontsize{10}{12}\selectfont \textit{\textbf{w/o Misaligned Context}}} \\
            \multirow{2}{*}{GPT-4.1}
            & 3.9 & 6.8 & 8.5 & 6.0 & 7.5 & 8.1 \\
            & \grbox{(-4.0)} & \rebox{(+1.2)} & \rebox{(+0.1)} & \grbox{(-4.6)} & \rebox{(+2.6)} & \rebox{(+1.1)} \\
            \multirow{2}{*}{Qwen2-VL$_{I.}$ {\small(7B)}}
            & 5.5 & 23.4 & 39.4 & 12.2 & 20.6 & 40.9 \\
            & \grbox{(-59.2)} & \rebox{(+17.5)} & \rebox{(+28.0)} & \grbox{(-65.0)} & \rebox{(+16.2)} & \rebox{(+34.0)} \\
            \multirow{2}{*}{LLaVA-Next$_{M.}$ {\small(7B)}}
            & 7.8 & 15.2 & 36.2 & 12.5 & 14.8 & 39.3 \\
            & \grbox{(-44.6)} & \rebox{(+10.2)} & \rebox{(+27.1)} & \grbox{(-45.0)} & \rebox{(+9.4)} & \rebox{(+31.0)} \\
            \multirow{2}{*}{Qwen2.5-VL$_{I.}$ {\small(72B)}}
            & 5.7 & 10.1 & 12.9 & 8.0 & 9.6 & 13.8 \\
            & \grbox{(-3.1)} & \rebox{(+1.9)} & \rebox{(+0.3)} & \grbox{(-4.2)} & \rebox{(+1.6)} & \rebox{(+3.3)} \\
            \bottomrule
        \end{tabular}%
    }

    \caption{Error analysis when provide misaligned context.}
    \label{tab:error_analysis}
\end{table}

\nbi{Exploration 4: Error analysis of Awareness of Temporal Misalignment.} Table~\ref{tab:error_analysis} provides a detailed error analysis of awareness experiment. The \rebox{red} values in the bracket mean a negative effect, while \grbox{green} means a positive.  \textbf{Con.} to context-based answers, \textbf{Oth.} to other answers, and \textbf{Irr.} to irrelevant ones. Surprisingly, even when provided with relevant context, models still generate responses that are irrelevant to the query or contain incorrect values from attribute list $A$, rather than leveraging the given context. This finding underscores the need to further investigate how models integrate external information with their internal knowledge.

\section{\label{sec:challenges}Can we update LMMs with time-sensitive knowledge?
}

Section~\ref{sec:probing} reveals that existing LMMs struggle to effectively process time-sensitive knowledge, while also being hampered by substantial amounts of outdated and irrelevant information. Knowledge editing updates factual knowledge in LLMs and LMMs, enabling efficient correction of outdated or inaccurate information without full retraining. Building on prior work \citep{cheng2023edit,huang2024vlkeb,mike2024,zhang2025mcmke}, we ask: \textit{Can LMMs be effectively updated with time-sensitive knowledge?} We explore multimodal time-sensitive knowledge editing and updating in real-world scenarios. We observe that LLaVA-v1.5 {\small(7B)} and Qwen-VL {\small(7B)} perform poorly and are therefore used as outdated models for knowledge editing. Regarding the selection of editing data, we extracted samples from these two models where CEM score is not 100 across five dimensions: cognition, trustworthiness, understanding, reasoning and robustness. Evaluation metric follows the protocol in Section~\ref{sec:Setup}. For more details, please refer to Appendix~\ref{appendix:ke}.

\noindent \textbf{Methods and Editing Setting:}  
We adopt two categories of multimodal knowledge editing approaches: parameter-modifying, like FT-LLM, FT-VIS, MEND \citep{mend} and parameter-preserving, like SERAC \citep{serac}, IKE \citep{ike}. We adopt the following two types of editing settings: \ding{182} Single editing restores weights after each edit, whereas \ding{183} lifelong editing examines the cumulative effects of editing entire dataset before evaluating all instances.

\begin{table}[t] 
  \centering
  \vspace{-5pt}
  \renewcommand{\arraystretch}{1.2} 
  \setlength{\tabcolsep}{1.5pt}     
  
  \resizebox{\linewidth}{!}{
    \begin{tabular}{l|c c c|c c|c|c c|c|c}
      \toprule
      
      \multirow{2.5}{*}{\textbf{Method}} 
        & \multicolumn{3}{c|}{\textbf{Cog.}} 
        & \multicolumn{2}{c|}{\textbf{Tru.}} 
        & \multicolumn{1}{c|}{\textbf{Und.}} 
        & \multicolumn{2}{c|}{\textbf{Rea.}} 
        & \multicolumn{1}{c|}{\textbf{Rob.}} 
        & \multirow{2.5}{*}{\rotatebox{50}{\textbf{Avg}}} \\ 
      
      \cmidrule(lr){2-10} 
        & \rotatebox{50}{\textbf{T.A}} & \rotatebox{50}{\textbf{T.I.A}} & \rotatebox{50}{\textbf{T.S.A}}
        & \rotatebox{50}{\textbf{P.U.D}} & \rotatebox{50}{\textbf{F.U.D}} 
        & \rotatebox{50}{\textbf{I.T.C}} 
        & \rotatebox{50}{\textbf{R.K}} & \rotatebox{50}{\textbf{C.A}} 
        & \rotatebox{50}{\textbf{A.T.E}} & \\
      \midrule
      
      \rowcolor{gray!10}
      \multicolumn{11}{c}{\fontsize{10}{12}\selectfont \textit{\textbf{LLaVA-v1.5 {\small(7B)}}}} \\
      
      FT-LLM  & \colorbox{backred!50}{98.0} & \colorbox{backred!50}{93.5} & 92.9 & \colorbox{backred!50}{100.0} & \colorbox{backred!50}{100.0} & \colorbox{backred!50}{96.2} & \colorbox{backred!50}{96.0} & \colorbox{backred!50}{97.8} & \colorbox{backred!50}{100.0} & \colorbox{backred!50}{97.2} \\
      FT-VIS  & 85.8 & 82.9 & 94.9 & 79.2  & 76.5  & 78.3 & 93.3 & 88.6 & 99.6  & 86.6 \\
      MEND    & 66.8 & 69.8 & 74.0 & \colorbox{backblue!75}{26.6}  & \colorbox{backblue!75}{18.1}  & \colorbox{backblue!75}{65.7} & 73.8 & 69.7 & \colorbox{backred!50}{100.0} & 62.7 \\
      
      SERAC   & \colorbox{backblue!75}{66.1} & \colorbox{backblue!75}{67.7} & \colorbox{backblue!75}{71.8} & 65.3  & 65.1  & 66.5 & \colorbox{backblue!75}{55.6} & \colorbox{backblue!75}{67.5} & \colorbox{backblue!75}{28.7}  & \colorbox{backblue!75}{61.6} \\
      IKE     & 85.7 & 82.4 & \colorbox{backred!50}{99.4} & 47.5  & 44.4  & 75.2 & 59.1 & 91.2 & 99.2  & 76.0 \\
      \midrule

      \rowcolor{gray!10}
      \multicolumn{11}{c}{\fontsize{10}{12}\selectfont \textit{\textbf{Qwen-VL {\small(7B)}}}} \\
      
      FT-LLM  & \colorbox{backred!50}{86.6} & \colorbox{backred!50}{86.6} & 89.9 & \colorbox{backred!50}{100.0}& \colorbox{backred!50}{100.0} & \colorbox{backred!50}{81.8} & \colorbox{backred!50}{87.5} & 89.0 & \colorbox{backred!50}{100.0} & \colorbox{backred!50}{91.3} \\
      FT-VIS  & 81.1 & 79.6 & 80.5 & \colorbox{backblue!75}{69.9} & 74.3 & 75.7 & 74.1 & 80.2 & \colorbox{backred!50}{100.0} & 79.5 \\
      MEND    & 68.1 & 70.5 & \colorbox{backblue!75}{54.9} & 79.7 & 84.8 & 64.1 & 65.7 & \colorbox{backblue!75}{50.2} & \colorbox{backred!50}{100.0} & 70.9 \\
      
      SERAC   & \colorbox{backblue!75}{57.2} & 66.2 & 62.1 & \colorbox{backblue!75}{69.9} & 74.6 & \colorbox{backblue!75}{56.4} & \colorbox{backblue!75}{63.0} & 52.2 & \colorbox{backblue!75}{18.4} & \colorbox{backblue!75}{57.8} \\
      IKE     & 86.5 & 78.1 & \colorbox{backred!50}{91.1} & 72.2 & \colorbox{backblue!75}{60.8} & 74.2 & 68.8 & \colorbox{backred!50}{92.8} & 92.3 & 79.6 \\
      
      \bottomrule
    \end{tabular}%
  }

  \caption{\textbf{Single Editing Performance Comparison ($\%$) on \dataset.} 
  The top and worst performing results are highlighted in 
  \colorbox{backred!50}{red} (1\textsuperscript{st}) 
  and \colorbox{backblue!75}{blue} (bottom) backgrounds, respectively.}
  \label{tab:single_editing}

\end{table}

\noindent  \textbf{Single Editing Shows Strong Effectiveness:} By observing Table~\ref{tab:single_editing}, we make the following observations: \ding{182} FT-LLM demonstrates strong performance as a knowledge updating method, achieving superior results across all evaluated tasks. \ding{183} In contrast, both the SERAC and MEND exhibit comparatively weaker performance, demonstrating limited effectiveness in knowledge updating tasks. \ding{184} Exception of SERAC, all methods achieve excellent performance on A.T.E task, demonstrating the strong robustness of current knowledge editing approaches. \ding{185} Knowledge updating significantly enhances the model's performance on complex I.T.C and C.A tasks.

\noindent  \textbf{Lifelong Editing Still Needs Improvement:} By observing Table~\ref{tab:lifelong}, we make the following observations: \ding{182} Except for P.U.D, F.U.D and A.T.E tasks, knowledge updating performance of FT-LLM, FT-VIS and SERAC has experienced varying degrees of loss. \ding{183} SERAC maintains excellent performance in lifelong editing scenario, with only 10.4\% loss. Its memory-based architecture mitigates catastrophic forgetting through explicit caching, maintaining robust performance in lifelong editing. \ding{184} Performance of SERAC in A.T.E has been improved by 12.6\%, which may be due to lifelong editing making SERAC better suited for robustness tasks.

\begin{table}[t]
  \centering
  \renewcommand{\arraystretch}{1.2} 
  \setlength{\tabcolsep}{1.5pt}     
  
  \resizebox{\linewidth}{!}{
    \begin{tabular}{l|c c c|c c|c|c c|c|c}
      \toprule
      
      \multirow{2.5}{*}{\textbf{Method}} 
        & \multicolumn{3}{c|}{\textbf{Cog.}} 
        & \multicolumn{2}{c|}{\textbf{Tru.}} 
        & \multicolumn{1}{c|}{\textbf{Und.}} 
        & \multicolumn{2}{c|}{\textbf{Rea.}} 
        & \multicolumn{1}{c|}{\textbf{Rob.}} 
        & \multirow{2.5}{*}{\rotatebox{50}{\textbf{Avg}}} \\
        
      \cmidrule(lr){2-4} \cmidrule(lr){5-6} \cmidrule(lr){7-7} \cmidrule(lr){8-9} \cmidrule(lr){10-10}

        & \rotatebox{50}{\textbf{T.A}} 
        & \rotatebox{50}{\textbf{T.I.A}} 
        & \rotatebox{50}{\textbf{T.S.A}}
        & \rotatebox{50}{\textbf{P.U.D}} 
        & \rotatebox{50}{\textbf{F.U.D}} 
        & \rotatebox{50}{\textbf{I.T.C}} 
        & \rotatebox{50}{\textbf{R.K}} 
        & \rotatebox{50}{\textbf{C.A}} 
        & \rotatebox{50}{\textbf{A.T.E}} 
        & \\ 
      \midrule
      
      \multirow{2}{*}{FT-LLM}
      & 31.0 & 32.3 & 25.9 & \textbf{100.0} & \textbf{99.0} & \textbf{9.3} & \textbf{60.4} & 27.6 & \textbf{100.0} & \textbf{54.0} \\
      & \rebox{(-67.0)} & \rebox{(-61.3)} & \rebox{(-67.0)} & \grbox{(+0.0)} & \rebox{(-1.0)} & \rebox{(-86.8)} & \rebox{(-35.6)} & \rebox{(-70.2)} & \grbox{(+0.0)} & \rebox{(-43.2)} \\
      \midrule 
      
      \multirow{2}{*}{FT-VIS}
      & 12.6 & 12.5 & 2.2  & 73.6  & 78.6 & 6.5 & 16.0 & 11.0 & \textbf{100.0} & 34.8 \\
      & \rebox{(-73.1)} & \rebox{(-70.4)} & \rebox{(-92.7)} & \rebox{(-5.6)} & \grbox{(+2.1)} & \rebox{(-71.9)} & \rebox{(-77.3)} & \rebox{(-77.6)} & \grbox{(+0.4)} & \rebox{(-51.8)} \\
      \midrule 

      \multirow{2}{*}{SERAC}
      & \textbf{53.7} & \textbf{53.3} & \textbf{70.1} & 66.0  & 66.4 & 5.9 & 42.7 & \textbf{61.8} & 41.2  & 51.2 \\
      & \rebox{(-12.4)} & \rebox{(-14.4)} & \rebox{(-1.7)} & \grbox{(+0.7)} & \grbox{(+1.3)} & \rebox{(-60.7)} & \rebox{(-12.9)} & \rebox{(-5.7)} & \grbox{(+12.6)} & \rebox{(-10.4)} \\
      \bottomrule
    \end{tabular}%
  }
  
  \caption{\textbf{Lifelong Editing Performance on \dataset.} All results are based on LLaVA-v1.5 {\small(7B)}. \rebox{Red} and \grbox{green} values mean negative and positive effects relative to data in Table~\ref{tab:single_editing}, respectively.}
  \label{tab:lifelong}

\end{table}

\section{Conclusion}

We propose \dataset, a comprehensive benchmark to evaluate LMMs on their time-sensitive knowledge capability. Our evaluation shows that while Gemini-2.5-Pro performs strongly, models still struggle with temporal awarenes , a limitation we explored by using knowledge editing to effectively update missing knowledge in single-edit scenarios. Our observations provide crucial directions for future research: \ding{182} Poor performance in the Awareness dimension suggests future methods must focus on improving the model's ability to distinguish the temporal consistency of internal knowledge and external context. \ding{183} Low scores in the Understanding dimension emphasize the urgent need to enhance the model's semantic comprehension and transformation capability for implicit temporal concepts. \ding{184} Poor performance in the Robustness dimension necessitates the development of more powerful self-correction and adversarial robustness mechanisms. These experimental results establish key technical hurdles and a clear roadmap for advancing LMMs toward dynamic knowledge systems.

\section*{Limitations}

While \dataset includes both original and generalization images, it is limited to static visual data and does not account for complex temporal dynamics, such as video. Additionally, our benchmark focuses on six representative domains, leaving highly specialized and time-critical fields like law and medicine for future exploration.

\section*{Ethical Considerations}

We recognize the ethical implications of deploying LMMs, where ensuring the integrity of time-sensitive multimodal knowledge is vital to prevent the spread of misinformation. Our research identifies critical limitations in current LMMs and demonstrates that knowledge editing can effectively mitigate these issues by updating outdated information, thereby enhancing model reliability.

\section*{Acknowledgement}

This research is supported by Anhui Provincial Natural Science Foundation (Grant No.2408085QF214), the National Key R\&D Program of China (Grant No.2024YFB3213400), the Fundamental Research Funds for the Central Universities (Grant No.WK2080000206), the Opening Project of the State Key Laboratory of General Artificial Intelligence (Project No.SKLAGI2025OP06, No.SKLAGI2024OP10, No.SKLAGI2024OP11) and project ZR2025QC1570 supported by Shandong Provincial Natural Science Foundation.

\bibliography{custom}

\newpage
\newpage
\appendix

\section{The Use of Large Language Models in \dataset}

In this section, we elaborate on the precise role of large language models within \dataset.

\begin{itemize}[leftmargin=*]

    \item \textbf{Usage 1: \dataset's construction.} In the dimension of Awareness of Temporal Misalignment (in Section~\ref{sec:Awareness}), GPT-4o is employed to generate contextual content related to temporal misalignment. This approach is consistent with current academic research norms.

    \item \textbf{Usage 2: \dataset's evaluation.} In Section~\ref{sec:main_result}, we evaluate performance on \dataset using Kimi-Latest, Gemini-2.5-Pro, Doubao-1.5-Vision-Pro, Seed-1.6-Vision and GPT-4.1, following standard benchmarking practices.

    \item \textbf{Usage 3: Paper grammar polishing.} The paper is initially drafted by human authors and subsequently polished for grammar using a large language model. It is not generated entirely by AI. This practice aligns with current academic norms.

\end{itemize}

\section{More details of \dataset}\label{appendix:benchmark_details}

\subsection{Data Construction}

\subsubsection{Original data construction pipeline}\label{appendix:original_construction_pipeline}

\begin{figure*}[t!]
  \centering
\includegraphics[width=1\linewidth]{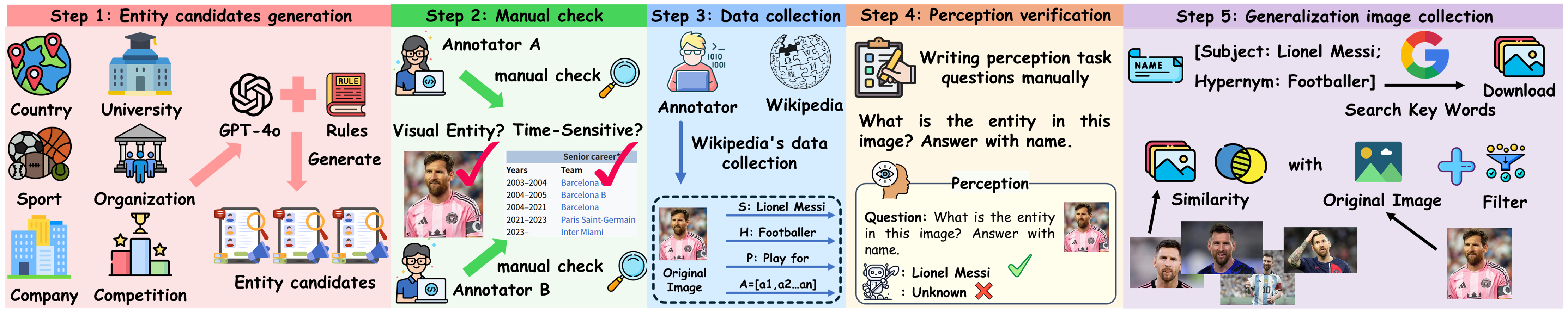}
  \caption{{Original data construction pipeline of \dataset.}}
  \label{fig:Original_data_construction}
\end{figure*}

We detail the original data construction pipeline for \dataset in Figure~\ref{fig:Original_data_construction} and outline the specific steps as follows:

\begin{itemize}[leftmargin=*]
\item \textbf{Step 1:} We define country, sport, company, university, organization and competition as the target domains and subsequently prompt GPT-4o to generate lists of suitable entity candidates for each.
\item \textbf{Step 2:} Two annotators manually search for information on every entity candidate via Wikipedia. Data are retained only if they meet two criteria: the entity must be visual and accurately representable by an image (\eg Lionel Messi), and it must be time-sensitive, meaning its attributes update over time (\eg which team Lionel Messi currently plays for).
\item \textbf{Step 3:} After discarding data where the two annotators disagree, we manually collect the following from Wikipedia for each remaining entry: the subject ($S$) (\eg a person or visual entity name like Lionel Messi), the hypernym ($H$) (\eg Lionel Messi’s hypernym is ``footballer''), the property ($P$) (\eg the property between Lionel Messi and club is ``play for''), a list of attribute values ($A = [a_1, a_2, \cdots, a_n]$, like $a_1$=``Paris Saint Germain F.C. | S:+2021-08-00 | E:+2023-06-30'') for that property which change over time, and the original image (the entity image provided by Wikipedia). Each entity ultimately possesses a quadruple $(S, H, P, A)$ and an original image.

\item \textbf{Step 4:} To evaluate the temporal awareness ability of LMMs, a prerequisite is that the models possess perceptual capability, meaning they must identify the evaluated entity from the image information. We address this by constructing 5 manually written perception task question templates, such as \en{What is the entity in this image?} Answer with name, and randomly assign them to each entity data point, thereby creating a perception capability QA pair <perception task question, subject> for every piece of data. We test the perception QA for each data point using 15 LMMs (\eg LLaVA-v1.5-7B, Qwen-VL, and GPT-4.1). We consider LMMs to lack adequate perception ability for an entity if 10 of these models fail to identify the entity in the image. To avoid interference with the subsequent temporal perception evaluation, we directly discard these failed entities.

\item \textbf{Step 5:} We use the subject plus hypernym as search keywords to download entity images from Google. We then use CLIP to extract features from both the downloaded and original images and calculate their cosine similarity. After excluding samples with a similarity score of 1, we select the top-1 resulting image as the generalization image. Each final data point comprises a quadruple $(S, H, P, A)$, an original image, and a generalization image.

\end{itemize}

\subsubsection{Task data construction pipeline}\label{appendix:task_construction_pipeline}

Next, we will provide a detailed introduction to the task data collection pipeline.

\begin{itemize}[leftmargin=*]
  \item \textbf{Cognition.} \textbf{Time-Agnostic \small(T.A):} We first write task question templates for the 6 knowledge domains (country, sport, company, university, organization and competition), where the Sport templates, for instance, include ``Which club does the {hypernym} in the image currently {property}?'' and ``The {hypernym} in the image currently {property}.'' Subsequently, we fill the hypernym and property from the original data into the corresponding templates. \textbf{Temporal Interval-Aware \small(T.I.A):} We similarly write task question templates for each knowledge domain; for example, the country templates are \en{Who was {property} the {hypernym} in the image from $T_{\text{start}}$ to $T_{\text{end}}$?} and \en{From $T_{\text{start}}$ to $T_{\text{end}}$, {property} the {hypernym} in the image was.} \textbf{Timestamp-Aware \small(T.S.A):} We write task question templates, such as the Company templates: \en{Who was {property} the {hypernym} in the image in $T_{\text{stamp}}$?} and \en{In $T_{\text{stamp}}$, {property} the {hypernym} in the image was.} Here, $T_{\text{stamp}}$ is a timestamp randomly selected from $T_{\text{start}}$ to $T_{\text{end}}$.
  \item \textbf{Awareness.} \textbf{Future Misaligned Context \small(F.M.C):} The construction of the question and answer aligns with the Timestamp-Aware task, utilizing the past timestamp $T_{\text{past}}$. Besides, we input (S, P, $a_{\text{current}}$) to prompt GPT-4o, which generates a relevant text description that serves as the Future Misaligned Context. The final task data (Future Misaligned Context, Question, and Answer) is processed as a single input unit. \textbf{Future Misaligned Context \small(P.M.C):} Similarly to the Future Misaligned Context, we construct the QA using the current timestamp $T_{\text{current}}$ and generate the ``Past Misaligned Context'' using (S, P, $a_{\text{past}}$).
  \item \textbf{Trustworthiness.} \textbf{Past Unanswerable Date \small(P.U.D):} Similarly to the Timestamp-Aware task, we randomly generate a Past Unanswerable Date for the attribute, which serves as $T_{\text{Past Unanswerable Date}}$. \textbf{Future Unanswerable Date \small(F.U.D):} Similarly to the Timestamp-Aware task, we randomly generate a Future Unanswerable Date for the attribute, which serves as $T_{\text{Future Unanswerable Date}}$.
  \item \textbf{Understanding.} \textbf{Implicit Temporal Concept \small(I.T.C):} We use historical events to replace explicit time periods, such as the phrase ``when Jeff Bezos served as CEO of Amazon'', which corresponds to the period ``from July 5, 1994, to July 5, 2021'' (in Figure~\ref{fig:overview}). These historical events, which replace explicit time periods, are uniquely matched from the original data's attribute. For instance, the time period when Jeff Bezos serves as CEO of Amazon, during which Lionel Messi plays exclusively for FC Barcelona, demonstrates temporal uniqueness.
  \item \textbf{Reasoning.} \textbf{Ranking \small(R.K):} We randomly select $a_1$ and $a_2$ from the original data's attribute list and write task question templates. For example, one template is: ``{attribute-1} and {attribute-2} all were {property} the {hypernym} in the image, respectively. Can you identify which one the former {property} was?'' \textbf{Calculation \small(C.A):} We first randomly select $a_1$ and $a_2$ from the original data's attribute list. We then select two timestamps, $t_{\text{1}}$ and $t_{\text{2}}$, from $a_1$'s and $a_2$'s $T_{\text{start}}$ to $T_{\text{end}}$ ranges, respectively, and calculate the time difference $T_{\triangle}$. Finally, we write task question templates, such as: {attribute} served as {property} the {hypernym} in the image in $t_{\text{1}}$. Can you identify who occupied this position after $T_{\triangle}$ years?.
  \item \textbf{Robustness.} \textbf{Adversarial Temporal Error \small(A.T.E):} We extract the QA pairs where all models fail the Cognition task. We then construct task question templates, such as: Your answer to the original question is wrong. ``Was {attribute} {property} the {hypernym} in the image from $T_{\text{start}}$ to $T_{\text{end}}$?'', which require the model to output either Yes or No.
\end{itemize}

\subsubsection{Inter Annotator Agreement Numbers}\label{appendix:Inter_Annotator_Agreement_Numebers}

We calculate Cohen's Kappa ($k = 0.8273$) using the formula:
\begin{equation}
k = 1 - \frac{1 - p_o}{1 - p_e}
\end{equation}
where $p_o$ denotes the observed agreement ratio between experts and $p_e$ denotes the hypothetical probability of chance agreement. Samples meeting the aforementioned screening criteria are classified as positive samples; others are negative samples. 

\vspace{5pt}

\begin{table}[h]
  \centering
  \renewcommand{\arraystretch}{1.2} 
  \setlength{\tabcolsep}{8pt}      
  
  \resizebox{0.8\linewidth}{!}{%
    \begin{tabular}{l|c c|c}
      \toprule
      \multirow{2.5}{*}{\textbf{Annotator B}} & \multicolumn{2}{c|}{\textbf{Annotator A}} & \multirow{2.5}{*}{\textbf{Total}} \\
      \cmidrule(lr){2-3} 
      & \textbf{Positive} & \textbf{Negative} & \\
      \midrule
      
      \textbf{Positive} & 455 & 19 & 474 \\
      
      \textbf{Negative} & 18 & 120 & 138 \\
      
      \midrule
      \rowcolor{gray!10}
      \textbf{Total} & 473 & 139 & 612 \\
      \bottomrule
    \end{tabular}%
  }
  
  \caption{Specific screening results of Inter Annotator Agreement Numbers.}
  \label{tab:kappa_agreement}
\end{table}

\subsubsection{\dataset’s Quality}\label{appendix:data_quality}

\textbf{For original data}, based on Figure~\ref{fig:Original_data_construction} and Section~\ref{appendix:original_construction_pipeline}, we observe that only the image data collected from Google excludes manual screening. To ensure high data quality, we employ CLIP to extract visual features from these images and retain only those with the highest cosine similarity to the original Wikipedia images. Since all other pipeline stages involve human verification, the overall quality of the dataset is rigorously maintained.

\textbf{For task data}, in Dimension 2 Awareness, we use GPT-4o to generate the ``Future and Past Misaligned Contexts''. Since this process may lead to semantic distortion and the introduction of bias, we address these concerns before data synthesis. Specifically, we ensure semantic fidelity and avoid bias through mandatory task instructions and diverse task examples, respectively. For example, one instruction is: ``You must generate authentic and relevant descriptions based on the provided information''.

At the same time, we also conduct human studies to verify data quality in Table~\ref{tab:human_study}. We randomly sample 20 data points from both the F.M.C. and P.M.C. tasks, requiring two annotators to manually write misaligned contexts for each. The annotators then compare the GPT-4o generated contexts against the human-written contexts to check for semantic distortion and bias, assigning a score between 0 and 10. A higher score indicates better quality for the GPT-4o context.

\begin{table}[h]
  \centering
  \renewcommand{\arraystretch}{1.2} 
  \setlength{\tabcolsep}{4pt}      
  
  \resizebox{\linewidth}{!}{
    \begin{tabular}{l|c c}
      \toprule
       & \textbf{F.M.C} & \textbf{P.M.C} \\
      \midrule
      
      \rowcolor{gray!10}
      \multicolumn{3}{c}{\fontsize{10}{12}\selectfont \textit{\textbf{Annotator A}}} \\
      GPT-4o generated contexts vs Manual writing 1 & 9.75 & 9.70 \\
      GPT-4o generated contexts vs Manual writing 2 & 9.57 & 9.65 \\
      \textbf{Mean-variance} & 9.66$\pm$0.13 & 9.68$\pm$0.04 \\
      
      \midrule
      
      \rowcolor{gray!10}
      \multicolumn{3}{c}{\fontsize{10}{12}\selectfont \textit{\textbf{Annotator B}}} \\
      GPT-4o generated contexts vs Manual writing 1 & 9.70 & 9.68 \\
      GPT-4o generated contexts vs Manual writing 2 & 9.82 & 9.70 \\
      \textbf{Mean-variance} & 9.76$\pm$0.08 & 9.69$\pm$0.01 \\
      
      \bottomrule
    \end{tabular}%
  }
  
  \vspace{-5pt} 
  \caption{Human studies for synthetic data and manual writing data.}
  \label{tab:human_study}
\end{table}

\subsubsection{\dataset’s Evolvability}\label{appendix:benchmark_Evolvability}

Owing to the time-sensitive nature of \dataset, we will perform quarterly updates to endow the benchmark with evolvability. Unlike conventional benchmarks that merely replace outdated data, \dataset offers a fundamentally distinct form of evolution. It not only evaluates model performance on time-sensitive knowledge but also probes models’ internal knowledge boundaries (in Section~\ref{sec:Exploratory}). To this end, we design an efficient pipeline to update the attribute list of each knowledge entry every quarter. This pipeline enables continuous renewal of knowledge, persistent evaluation of model knowledge boundaries, and provides the community with a dynamic and evolving evaluation resource. We outline \dataset's update pipeline:

\begin{itemize}[leftmargin=*]
  \item (1) Leveraging existing \dataset subject $S$ data, we retrieve corresponding Wikipedia text data offline (e.g., searching ``Lionel Messi'').

  \item (2) For club affiliation information, we extract information from Wikipedia's career sections using GPT-4o with strict parsing rules (the career field contains Lionel Messi's club affiliation information).

  \item (3) Newly extracted club data is compared against \dataset's current records, triggering updates when discrepancies occur. This efficient pipeline ensures automated, continuous \dataset updates, providing the community with an evolving evaluation resource.
\end{itemize}

Combined with this automated update pipeline, our proposed \dataset benchmark can not only evaluate current state-of-the-art LMMs, \textbf{but also be used to evaluate newly emerging and more powerful LMMs in the future}.

\subsubsection{\dataset’s Quantity}

\vspace{5pt}

\begin{table}[h!]
  \centering
  \renewcommand{\arraystretch}{1.2} 
  \setlength{\tabcolsep}{2.5pt} 
  
  \resizebox{\linewidth}{!}{
    \begin{tabular}{c c c|c c|c c|c|c c|c|c}
      \toprule
      \multicolumn{3}{c|}{\textbf{Cog.}} 
      & \multicolumn{2}{c|}{\textbf{Awa.}} 
      & \multicolumn{2}{c|}{\textbf{Tru.}} 
      & \multicolumn{1}{c|}{\textbf{Und.}} 
      & \multicolumn{2}{c|}{\textbf{Rea.}} 
      & \multicolumn{1}{c|}{\textbf{Rob.}} 
      & \multirow{2.5}{*}{\rotatebox{50}{\textbf{Sum}}} \\ 
      \cmidrule(lr){1-11} 
      
      \rotatebox{50}{\textbf{T.A}} & \rotatebox{50}{\textbf{T.I.A}} & \rotatebox{50}{\textbf{T.S.A}}
      & \rotatebox{50}{\textbf{F.M.C}} & \rotatebox{50}{\textbf{P.M.C}} 
      & \rotatebox{50}{\textbf{P.U.D}} & \rotatebox{50}{\textbf{F.U.D}} 
      & \rotatebox{50}{\textbf{I.T.C}}   
      & \rotatebox{50}{\textbf{R.K}} & \rotatebox{50}{\textbf{C.A}} 
      & \rotatebox{50}{\textbf{A.T.E}} & \\
      \midrule
      
      255 & 172 & 237 & 236 & 181 & 207 & 207 & 255 & 81 & 81 & 192 & 2104 \\
      \bottomrule
    \end{tabular}%
  }
  
  \vspace{-5pt} 
  \caption{The detailed quantity of time-sensitive knowledge for each task.}
  \label{tab:my_label}
\end{table}

\section{Experiment Resources about \dataset}\label{appendix:Experiment_Resources}

\begin{table*}[h!]
  \centering
    

  \renewcommand{\arraystretch}{1.1}  
  \resizebox{\textwidth}{!}{%
    \begin{tabular}{l|c c c|c c|c c|c|c c|c|c}
      \toprule
      \multirow{2.5}{*}{\textbf{(Release Time) Models}} 
        & \multicolumn{3}{c|}{\textbf{Cog.}} 
        & \multicolumn{2}{c|}{\textbf{Awa.}} 
        & \multicolumn{2}{c|}{\textbf{Tru.}} 
        & \multicolumn{1}{c|}{\textbf{Und.}} 
        & \multicolumn{2}{c|}{\textbf{Rea.}} 
        & \multicolumn{1}{c|}{\textbf{Rob.}} 
        & \multirow{2.5}{*}{\textbf{Avg.}} \\
      \cmidrule{2-12}
        & \textbf{T.A}  & \textbf{T.I.A} & \textbf{T.S.A}
        & \textbf{F.M.C} & \textbf{P.M.C} 
        & \textbf{P.U.D} & \textbf{F.U.D} 
        & \textbf{I.T.C}   
        & \textbf{R.K} & \textbf{C.A} 
        & \textbf{A.T.E} \\

      \midrule
      \multicolumn{13}{l}{\fontsize{10}{12}\selectfont \textit{\textbf{Open-source LMMs}}} \\	
      \rowcolor{gray!10}
      \multicolumn{13}{c}{\fontsize{10}{12}\selectfont \textit{\textbf{Model size under 10B}}} \\
    
      {\small(2023.04)} LLaVA-v1.5 {\small(7B)} & 7.89 & 11.44 & 16.88 & 10.60 & 9.49 & 53.99 & 50.00 & 1.95 & 15.33 & 6.38 & 0.39 & \colorbox{backblue!75}{16.76} \\
    {\small(2023.08)} Qwen-VL {\small(7B)} & 14.56 & 20.30 & 47.09 & 7.66 & 8.81 & 80.00 & 69.40 & 4.94 & 23.13 & 18.96 & \colorbox{backblue!75}{0.00} & 26.80 \\
    {\small(2023.11)} mPLUG-Owl2 {\small(7B)} & 13.40 & 17.05 & 50.94 & 48.26 & 44.21 & \colorbox{backblue!75}{11.19} & \colorbox{backblue!75}{44.20} & 3.34 & 43.40 & 16.59 & 6.12 & 27.15 \\
    {\small(2024.01)} LLaVA-Next$_{L.}$ {\small(8B)} & 9.39 & 16.68 & 46.39 & 47.51 & 38.20 & \colorbox{backred!50}{99.64} & 99.88 & 3.47 & 36.08 & 10.85 & 0.13 & 37.11 \\
    {\small(2024.01)} LLaVA-Next$_{M.}$ {\small(7B)} & 13.37 & 18.74 & 46.59 & 37.34 & 32.05 & 96.74 & 90.22 & 4.43 & 38.85 & 24.23 & \colorbox{backblue!75}{0.00} & 36.60 \\
    {\small(2024.01)} LLaVA-Next$_{V.}$ {\small(7B)} & 13.89 & 18.34 & 39.15 & 27.60 & 22.54 & 81.16 & 87.92 & 3.99 & 32.23 & 15.25 & \colorbox{backyellow_soft!40}{31.25} & 33.94 \\
    {\small(2024.08)} LLaVA-OV {\small(7B)} & 14.22 & 15.24 & 31.91 & 35.12 & 34.84 & 39.61 & 76.21 & 4.86 & 52.56 & 14.73 & 2.21 & 29.23 \\
    {\small(2024.08)} mPlug-Owl3 {\small(8B)} & 9.94 & 14.07 & 33.09 & 21.87 & 20.86 & 97.60 & 99.76 & 3.27 & 41.53 & 7.62 & 3.65 & 32.11 \\
    {\small(2024.08)} MiniCPM-V2.6 {\small(8B)} & 24.11 & 25.91 & 58.78 & 41.37 & 34.63 & 81.52 & 97.83 & 5.81 & \colorbox{backyellow_soft!40}{53.67} & 27.74 & 14.45 & 42.35 \\
    {\small(2024.09)} Qwen2-VL$_{I.}$ {\small(7B)} & 19.20 & 21.34 & 37.49 & 21.92 & 14.71 & \colorbox{backyellow_soft!40}{99.52} & 99.76 & 6.09 & 50.27 & 18.40 & 9.90 & 36.24 \\
    {\small(2024.12)} InternVL2.5 {\small(1B)} & \colorbox{backblue!75} {4.53} & \colorbox{backblue!75}{2.65} & \colorbox{backblue!75}{4.86} & \colorbox{backblue!75}{3.48} & \colorbox{backblue!75}{3.06} & 97.95 & 98.43 & \colorbox{backblue!75}{1.19} & 42.35 & \colorbox{backblue!75}{3.85} & \colorbox{backblue!75}{0.00} & 23.85 \\
    {\small(2024.12)} InternVL2.5 {\small(2B)} & 6.67 & 7.29 & 10.21 & 5.96 & 4.98 & 96.74 & 95.89 & 2.04 & 13.77 & 5.27 & 0.78 & 22.69 \\
    {\small(2024.12)} InternVL2.5 {\small(4B)} & 21.02 & 17.35 & 35.32 & 34.06 & 31.36 & 98.43 & 99.28 & 4.26 & 47.74 & 22.07 & 1.56 & 37.50 \\
    {\small(2024.12)} InternVL2.5 {\small(8B)} & 21.71 & 23.29 & 49.14 & 47.38 & 42.64 & 98.31 & \colorbox{backyellow_soft!40}{99.88} & 6.00 & \colorbox{backred!50}{62.11} & 24.52 & \colorbox{backblue!75}{0.00} & 43.18 \\
    {\small(2025.02)} Qwen2.5-VL$_{I.}$ {\small(3B)} & 19.55 & 16.39 & 25.16 & 15.20 & 14.61 & 40.10 & 57.25 & 5.28 & 50.58 & 16.46 & 9.38 & 24.54 \\
    {\small(2025.02)} Qwen2.5-VL$_{I.}$ {\small(7B)} & 21.59 & 22.29 & 47.47 & 45.77 & 38.83 & \colorbox{backred!50}{99.64} & 99.76 & 5.74 & 39.22 & 28.35 & 22.29 & 42.81 \\

      \rowcolor{gray!10}
      \multicolumn{13}{c}{\fontsize{10}{12}\selectfont \textit{\textbf{Model size under 65B}}} \\
    {\small(2024.12)} InternVL2.5 {\small(26B)} & 23.85 & 26.20 & 62.74 & 54.07 & 52.18 & 97.22 & 99.52 & 6.52 & 27.71 & 25.33 & 8.33 & 43.97 \\
    {\small(2024.12)} InternVL2.5 {\small(38B)} & 29.71 & 32.50 & 73.72 & 68.91 & 62.41 & 92.63 & 99.15 & 5.48 & 32.83 & 32.82 & 11.33 & 49.23 \\

      \rowcolor{gray!10}
      \multicolumn{13}{c}{\fontsize{10}{12}\selectfont \textit{\textbf{Model size under 100B}}} \\     
    {\small(2024.12)} InternVL2.5 {\small(78B)} & 30.44 & 35.91 & 75.35 & 74.59 & 73.79 & 81.16 & 97.58 & 7.75 & 12.80 & 43.09 & 8.33 & 49.16 \\
    {\small(2025.02)} Qwen2.5-VL$_{I.}$ {\small(72B)} & 32.42 & 36.97 & 76.21 & 75.32 & 73.56 & 91.67 & 97.95 & 7.78 & \colorbox{backblue!75}{11.91} & 38.07 & 5.73 & 49.78 \\

    \midrule
      \multicolumn{13}{l}{\fontsize{10}{12}\selectfont \textit{\textbf{Closed-source LMMs}}} \\
    {\small(2025.02)} Kimi-Latest & 28.55 & 31.63 & 76.34 & 73.19 & 71.16 & 72.10 & 85.27 & 8.45 & 46.48 & 47.12 & 6.38 & 49.70 \\
    {\small(2025.03)} Doubao-1.5-Vision-Pro & 36.87 & 34.33 & 76.52 & 78.39 & 74.61 & 93.12 & \colorbox{backred!50}{100.00} & 6.21 & 19.71 & 38.63 & 12.24 & 51.88 \\
    {\small(2025.03)} Gemini-2.5-Pro & 35.21 & \colorbox{backred!50}{58.86} & \colorbox{backred!50}{87.06} & \colorbox{backred!50}{86.37} & \colorbox{backred!50}{86.67} & 75.50 & 93.77 & \colorbox{backred!50}{17.39} & 39.72 & \colorbox{backred!50}{81.21} & \colorbox{backred!50}{31.94} & \colorbox{backred!50}{63.07} \\
    {\small(2025.04)} GPT-4.1 & \colorbox{backyellow_soft!40}{37.26} & 43.42 & \colorbox{backyellow_soft!40}{84.93} & \colorbox{backyellow_soft!40}{82.47} & 82.02 & 64.44 & 91.30 & \colorbox{backyellow_soft!40}{10.11} & 16.77 & 62.03 & 17.58 & 53.85 \\
    {\small(2025.08)} Seed-1.6-Vision  & \colorbox{backred!50}{38.50} & \colorbox{backyellow_soft!40}{48.55} & 82.83 & 79.85 & \colorbox{backyellow_soft!40}{83.59} & 74.15 & 96.86 & 9.22 & 22.00 & \colorbox{backyellow_soft!40}{62.55} & 31.05 & \colorbox{backyellow_soft!40}{57.20} \\

      \bottomrule
    \end{tabular}%
  }
  \vspace{-5pt}
  \caption{\textbf{Complete F1-Score Performance Comparison ($\%$) on \dataset.} 
The top two and worst results are highlighted in 
\colorbox{backred!50}{red} (1\textsuperscript{st}), 
 \colorbox{backyellow_soft!40}{yellow} (2\textsuperscript{nd}) and \colorbox{backblue!75}{blue} (bottom) backgrounds, respectively. Subscripts  ${L}$, ${M}$, ${V}$ and ${I}$ stand for LLaMA3-8B, Mistral-7B, Vicuna-7B and Instruct, respectively.}

  \label{table:all_f1}

\end{table*}

\textbf{Probing Time-Sensitive Knowledge:}  Regarding the validation experiments of LMMs on \dataset, for models with parameter sizes of 38B or less, we conduct experiments on 4 NVIDIA A100 PCIEs machines (40 GiB each); For models with parameter sizes greater than 38B, we conduct experiments on 4 NVIDIA H100 (96 GiB each).

\textbf{Editing Time-Sensitive Knowledge:} We conduct knowledge editing experiment on one H100 (96 GiB each) regarding LMMs.

\section{More experimental results about \dataset}\label{appendix:more_results}

\begin{table*}[h!]
  \centering
  

  \renewcommand{\arraystretch}{1.1}  
  \resizebox{\textwidth}{!}{%
    \begin{tabular}{l|c c c|c c|c c|c|c c|c|c}
      \toprule
      \multirow{2.5}{*}{\textbf{(Release Time) Models}} 
        & \multicolumn{3}{c|}{\textbf{Cog.}} 
        & \multicolumn{2}{c|}{\textbf{Awa.}} 
        & \multicolumn{2}{c|}{\textbf{Tru.}} 
        & \multicolumn{1}{c|}{\textbf{Und.}} 
        & \multicolumn{2}{c|}{\textbf{Rea.}} 
        & \multicolumn{1}{c|}{\textbf{Rob.}} 
        & \multirow{2.5}{*}{\textbf{Avg.}} \\
      \cmidrule{2-12}
        & \textbf{T.A}  & \textbf{T.I.A} & \textbf{T.S.A}
        & \textbf{F.M.C} & \textbf{P.M.C} 
        & \textbf{P.U.D} & \textbf{F.U.D} 
        & \textbf{I.T.C}   
        & \textbf{R.K} & \textbf{C.A} 
        & \textbf{A.T.E} \\

      \midrule
      \multicolumn{13}{l}{\fontsize{10}{12}\selectfont \textit{\textbf{Open-source LMMs}}} \\
      \rowcolor{gray!10}
      \multicolumn{13}{c}{\fontsize{10}{12}\selectfont \textit{\textbf{Model size under 10B}}} \\
    {\small(2023.04)} LLaVA-v1.5 {\small(7B)} & 6.96 & 9.25 & 16.88 & 7.66 & 6.40 & 53.99 & 50.00 & 1.57 & 15.12 & 6.17 & 0.39 & \colorbox{backblue!75}{15.85} \\
    {\small(2023.08)} Qwen-VL {\small(7B)} & 12.45 & 17.30 & 42.09 & 6.04 & 6.91 & 81.28 & 70.17 & 3.53 & 25.00 & 17.59 & \colorbox{backblue!75}{0.00} & 25.67 \\
    {\small(2023.11)} mPLUG-Owl2 {\small(7B)} & 10.59 & 14.53 & 44.62 & 42.69 & 38.67 & \colorbox{backblue!75}{11.47} & \colorbox{backblue!75}{44.20} & 2.16 & 42.90 & 14.20 & 6.12 & 24.74 \\
    {\small(2024.01)} LLaVA-Next$_{L.}$ {\small(8B)} & 8.24 & 12.21 & 39.03 & 41.10 & 31.63 & \colorbox{backred!50}{99.64} & 99.88 & 2.35 & 35.19 & 8.33 & 0.13 & 34.34 \\
    {\small(2024.01)} LLaVA-Next$_{M.}$ {\small(7B)} & 10.69 & 14.53 & 41.14 & 33.69 & 28.87 & 96.74 & 90.22 & 3.73 & 38.58 & 20.99 & \colorbox{backblue!75}{0.00} & 34.47 \\
    {\small(2024.01)} LLaVA-Next$_{V.}$ {\small(7B)} & 11.47 & 14.83 & 34.39 & 23.62 & 17.82 & 81.16 & 87.92 & 2.55 & 31.17 & 10.80 & 31.25 & 31.54 \\
    {\small(2024.08)} LLaVA-OV {\small(7B)} & 11.86 & 11.34 & 26.79 & 30.93 & 31.35 & 39.61 & 76.21 & 3.63 & 51.54 & 8.95 & 2.21 & 26.77 \\
    {\small(2024.08)} mPlug-Owl3 {\small(8B)} & 9.80 & 10.03 & 29.01 & 29.77 & 28.31 & 97.95 & 99.76 & 3.14 & 41.98 & 7.10 & 3.65 & 32.77 \\
    {\small(2024.08)} MiniCPM-V2.6 {\small(8B)} & 22.16 & 21.66 & 55.70 & 38.88 & 31.35 & 81.52 & 97.83 & 4.22 & \colorbox{backyellow_soft!40}{52.78} & 24.38 & 14.45 & 40.45 \\
    {\small(2024.09)} Qwen2-VL$_{I.}$ {\small(7B)} & 15.98 & 16.72 & 31.96 & 17.90 & 11.46 & \colorbox{backyellow_soft!40}{99.52} & 99.76 & 4.61 & 49.38 & 14.20 & 9.90 & 33.76 \\
    {\small(2024.12)} InternVL2.5 {\small(1B)} & 6.96 & \colorbox{backblue!75}{3.49} & \colorbox{backblue!75}{7.28} & \colorbox{backblue!75}{3.92} & 3.31 & 97.95 & 98.43 & 2.35 & 45.06 & \colorbox{backblue!75}{3.40} & \colorbox{backblue!75}{0.00} & 24.74 \\
    {\small(2024.12)} InternVL2.5 {\small(2B)} & \colorbox{backblue!75}{5.59} & 5.52 & 9.07 & 4.03 & \colorbox{backblue!75}{3.18} & 96.74 & 95.89 & \colorbox{backblue!75}{0.88} & 13.27 & 4.32 & 0.78 & 21.75 \\
    {\small(2024.12)} InternVL2.5 {\small(4B)} & 18.63 & 13.66 & 32.91 & 31.36 & 28.31 & 98.43 & 99.28 & 3.04 & 47.53 & 20.06 & 1.56 & 35.89 \\
    {\small(2024.12)} InternVL2.5 {\small(8B)} & 20.49 & 18.46 & 44.83 & 42.37 & 38.26 & 98.31 & \colorbox{backyellow_soft!40}{99.88} & 4.22 & \colorbox{backred!50}{61.73} & 19.14 & \colorbox{backblue!75}{0.00} & 40.70 \\
    {\small(2025.02)} Qwen2.5-VL$_{I.}$ {\small(3B)} & 17.65 & 13.66 & 21.41 & 12.08 & 11.88 & 40.10 & 57.25 & 3.73 & 50.31 & 13.58 & 9.38 & 22.82 \\
    {\small(2025.02)} Qwen2.5-VL$_{I.}$ {\small(7B)} & 18.33 & 16.86 & 41.67 & 40.04 & 33.98 & \colorbox{backred!50}{99.64} & 99.76 & 4.02 & 38.89 & 25.00 & 16.86 & 39.55 \\
    
    \rowcolor{gray!10}
    \multicolumn{13}{c}{\textit{\textbf{Model size under 65B}}} \\
 
    {\small(2024.12)} InternVL2.5 {\small(26B)} & 21.96 & 21.37 & 59.39 & 49.79 & 49.72 & 97.22 & 99.52 & 5.00 & 26.85 & 20.99 & 8.33 & 41.83 \\
    {\small(2024.12)} InternVL2.5 {\small(38B)} & 28.43 & 27.47 & 70.15 & 65.78 & 59.81 & 92.63 & 99.15 & 4.31 & 31.79 & 28.70 & 11.33 & 47.23 \\

    \rowcolor{gray!10}
    \multicolumn{13}{c}{\textit{\textbf{Model size under 100B}}} \\

    {\small(2024.12)} InternVL2.5 {\small(78B)} & 29.31 & 28.63 & 70.25 & 69.92 & 70.86 & 81.16 & 97.58 & 5.98 & 11.73 & 38.58 & 8.33 & 46.58\\
    {\small(2025.02)} Qwen2.5-VL$_{I.}$ {\small(72B)} & 29.22 & 31.10 & 71.41 & 70.44 & 69.34 & 91.67 & 97.95 & 6.18 & \colorbox{backblue!75}{11.42} & 34.88 & 5.73 & 47.21 \\
    \midrule
    \multicolumn{13}{l}{\textit{\textbf{Closed-source LMMs}}} \\

    {\small(2025.02)} Kimi-Latest & 26.41 & 26.60 & 72.43 & 68.64 & 67.27 & 72.10 & 85.39 & 7.06 & 45.99 & 42.59 & 6.38 & 47.35 \\
    {\small(2025.02)} Doubao-1.5-Vision-Pro & 35.78 & 27.91 & 69.83 & 74.36 & 70.76 & 93.12 & \colorbox{backred!50}{100.00} & 5.29 & 18.52 & 34.57 & 12.24 & 49.31 \\
    {\small(2025.03)} Gemini-2.5-Pro & 34.25 & \colorbox{backred!50}{56.40} & \colorbox{backred!50}{84.96} & \colorbox{backred!50}{83.09} & \colorbox{backred!50}{84.30} & 80.31 & 97.10 & \colorbox{backred!50}{18.73} & 38.48 & \colorbox{backred!50}{76.54} & \colorbox{backred!50}{39.58} & \colorbox{backred!50}{63.07} \\
    {\small(2025.04)} GPT-4.1 & \colorbox{backred!50}{37.58} & 37.94 & \colorbox{backyellow_soft!40}{80.91} & \colorbox{backyellow_soft!40}{78.07} & 77.49 & 65.22 & 91.30 & \colorbox{backyellow_soft!40}{8.63} & 15.74 & \colorbox{backyellow_soft!40}{59.57} & 17.58 & 51.82 \\
    {\small(2025.08)} Seed-1.6-Vision & \colorbox{backyellow_soft!40}{37.19} & \colorbox{backyellow_soft!40}{41.76} & 78.69 & 75.95 & \colorbox{backyellow_soft!40}{80.71} & 74.15 & 96.86 & 7.55 & 21.60 & \colorbox{backyellow_soft!40}{59.57} & \colorbox{backyellow_soft!40}{32.68} & \colorbox{backyellow_soft!40}{55.16} \\
      
      \bottomrule
    \end{tabular}%
  }
  \vspace{-5pt}
\caption{\textbf{Complete CEM Performance Comparison ($\%$) on \dataset.} 
The top two and worst results are highlighted in 
\colorbox{backred!50}{red} (1\textsuperscript{st}), 
 \colorbox{backyellow_soft!40}{yellow} (2\textsuperscript{nd}) and \colorbox{backblue!75}{blue} (bottom) backgrounds, respectively. Subscripts  ${L}$, ${M}$, ${V}$ and ${I}$ stand for LLaMA3-8B, Mistral-7B, Vicuna-7B and Instruct, respectively.}

 \label{table:all_cem}
\end{table*}

\subsection{Experimental results based on different metrics about \dataset}

\subsubsection{F1-Score}\label{appendix:metric_f1}

In this section, we present the complete experimental results on \dataset. To further validate the reliability of our conclusions, we also employed the F1-Score as an additional evaluation metric.

The F1-Score is a metric for assessing model performance by quantifying the word-level similarity between a model's output and the ground truth answer. It is the harmonic mean of Precision and Recall~\citep{chan2024rqrag}.

To calculate it, we first represent both the ground truth and the prediction as sets of words. Let the ground truth be \( \mathcal{W}(y_q) = \{ y_1, \dots, y_m \} \) and the model's prediction be \( \mathcal{W}(\hat{Y}) = \{ \hat{y}_1, \dots, \hat{y}_n \} \). The number of common words between these sets, known as the overlap \( \mathcal{U}(\hat{Y}, y_q) \), is computed using an indicator function \( \mathbf{1}[\cdot] \):
\begin{equation}
\mathcal{U}(\hat{Y}, y_q) = \sum_{t \in \mathcal{W}(y_q)} \mathbf{1}[t \in \mathcal{W}(\hat{Y})]
\end{equation}
Precision, \( \mathcal{P}(\hat{Y}, Y) \), is the fraction of relevant words among the predicted words. It is formally defined as:
\begin{equation}
\mathcal{P}(\hat{Y}, Y) = \frac{\mathcal{U}(\hat{Y}, y_q)}{|\mathcal{W}(\hat{Y})|}
\end{equation}
Recall, \( \mathcal{R}(\hat{Y}, Y) \), is the fraction of ground truth words that the model successfully identified. It is defined as:
\begin{equation}
\mathcal{R}(\hat{Y}, Y) = \frac{\mathcal{U}(\hat{Y}, y_q)}{|\mathcal{W}(y_q)|}
\end{equation}

According to the results in Table~\ref{table:all_f1}, we found that the conclusion drawn when using F1-Score as the evaluation metric is consistent with the conclusion drawn when using CEM as the evaluation metric, highlighting the reliability of our results and observations.

\subsubsection{Cover Exact Match}\label{appendix:metric_cem}

In Table~\ref{table:all_cem}, we also present the complete experimental results based on CEM for reference.

\subsubsection{LLM as judge}\label{appendix:metric_llm}

\begin{table*}[t!] 
  \centering
  \renewcommand{\arraystretch}{1.1}  
  
  \resizebox{\textwidth}{!}{%
    \begin{tabular}{l|c c c|c c|c c|c|c c|c|c}
      \toprule
      \multirow{2.5}{*}{\textbf{{\small(Release Time)} Models}} 
        & \multicolumn{3}{c|}{\textbf{Cog.}} 
        & \multicolumn{2}{c|}{\textbf{Awa.}} 
        & \multicolumn{2}{c|}{\textbf{Tru.}} 
        & \multicolumn{1}{c|}{\textbf{Und.}} 
        & \multicolumn{2}{c|}{\textbf{Rea.}} 
        & \multicolumn{1}{c|}{\textbf{Rob.}} 
        & \multirow{2.5}{*}{\textbf{Avg.}} \\
      \cmidrule{2-12}
        & \textbf{T.A} \daugshifted & \textbf{T.I.A}\daugshifted  & \textbf{T.S.A} \daugshifted
        & \textbf{F.M.C}  \daugshifted & \textbf{P.M.C} \daugshifted
        & \textbf{P.U.D} \daugshifted & \textbf{F.U.D}  \daugshifted
        & \textbf{I.T.C}  \daugshifted  
        & \textbf{R.K} \daugshifted & \textbf{C.A}  \daugshifted
        & \textbf{A.T.E} \daugshifted \\                
      
      \midrule
      \rowcolor{gray!10}
      \multicolumn{13}{c}{\fontsize{10}{12}\selectfont \textit{\textbf{Open-source LMMs}}} \\    

      {\small(2023.04)} LLaVA-v1.5 {\small(7B)}   
      &\colorbox{backblue!75}{10.46} &\colorbox{backblue!75}{13.01} &\colorbox{backblue!75}{20.93} &\colorbox{backblue!75}{16.91} &\colorbox{backblue!75}{16.92} &53.99 &50.01 &\colorbox{backblue!75}{2.89} &\colorbox{backblue!75}{24.44} &\colorbox{backblue!75}{7.80} &0.39&\colorbox{backblue!75}{19.80} \\
      {\small(2023.08)} Qwen-VL {\small(7B)}      
      &20.20 &25.29 &55.46 &18.64 &19.05 &81.27 &70.17 &9.10 &39.52 &27.22 &\colorbox{backblue!75}{0.00}&33.27 \\
      {\small(2023.11)} mPLUG-Owl2 {\small(7B)} 
      &16.50&20.06&56.93&52.92&49.24&\colorbox{backblue!75}{12.00}&\colorbox{backblue!75}{44.42}&5.38&52.10&23.79&6.12&30.86 \\
      {\small(2024.01)} LLaVA-Next$_{M.}$ {\small(7B)} 
      &18.55&21.74&52.03&44.50&40.70&96.75&90.23&7.00&46.17&29.59&\colorbox{backblue!75}{0.00}&40.66 \\
      {\small(2024.08)} LLaVA-OV {\small(7B)}     
      &19.08&19.80&36.79&40.67&40.65&39.92&76.62&8.26&57.16&19.89&2.21&32.82 \\
      {\small(2024.08)} mPlug-Owl3 {\small(8B)}   
      &16.51&18.30&41.89&40.63&38.72&98.07&99.76&6.31&46.33&13.30&3.66&38.50 \\
      {\small(2024.08)} MiniCPM-V2.6 {\small(8B)} 
      &28.41&29.36&62.90&47.49&41.82&81.52&97.83&9.16&\colorbox{backyellow_soft!40}{60.40}&34.14&14.45&46.13 \\
      {\small(2024.09)} Qwen2-VL$_{I.}$ {\small(7B)}   
      &26.37&27.62&44.76&30.00&24.44&\colorbox{backyellow_soft!40}{99.52}&99.76&10.60&56.62&27.26&9.90&41.53 \\
      {\small(2024.12)} InternVL2.5 {\small(8B)} 
      &24.57&26.48&55.14&54.32&49.50&98.31&\colorbox{backyellow_soft!40}{99.88}&9.58&\colorbox{backred!50}{65.78}&31.16&\colorbox{backblue!75}{0.00}&46.79 \\
      {\small(2025.02)} Qwen2.5-VL$_{I.}$ {\small(7B)} 
      &26.48&27.78&53.21&51.75&45.83&\colorbox{backred!50}{99.64}&99.76&9.83&48.07&34.64&17.78&46.80 \\

      \midrule
      \rowcolor{gray!10}
      \multicolumn{13}{c}{\fontsize{10}{12}\selectfont \textit{\textbf{Closed-source LMMs}}} \\    

      {\small(2025.02)} Kimi-Latest &33.69&34.56&78.89&76.91&74.44&72.12&86.59&12.33&54.11&52.93&6.38&53.00 \\
      {\small(2025.02)} Doubao-1.5-Vision-Pro &40.25&37.80&80.59&81.41&78.06&93.12&\colorbox{backred!50}{100.00}&10.07&40.07&44.26&12.24&56.17 \\
      {\small(2025.03)} Gemini-2.5-Pro &\colorbox{backred!50}{62.04}&\colorbox{backred!50}{62.04}&\colorbox{backred!50}{90.40}&\colorbox{backred!50}{88.94}&\colorbox{backred!50}{89.62}&79.22&96.28&\colorbox{backred!50}{20.84}&47.47&\colorbox{backred!50}{84.78}&\colorbox{backred!50}{39.50}& \colorbox{backred!50}{69.20}\\
      {\small(2025.04)} GPT-4.1 &41.16&47.41&\colorbox{backyellow_soft!40}{87.47}&\colorbox{backyellow_soft!40}{84.99}&85.27&65.36&91.41&\colorbox{backyellow_soft!40}{13.63}&37.41&66.81&17.58&58.05 \\
      {\small(2025.08)} Seed-1.6-Vision &\colorbox{backyellow_soft!40}{42.61}&\colorbox{backyellow_soft!40}{51.36}&86.59&83.89&\colorbox{backyellow_soft!40}{86.93}&74.15&96.62&13.37&42.22&\colorbox{backyellow_soft!40}{68.88}&\colorbox{backyellow_soft!40}{32.47}&\colorbox{backyellow_soft!40}{61.74} \\
      
      \bottomrule
    \end{tabular}%
  }

  \vspace{-5pt} 
  \caption{\textbf{Overall Performance Comparison ($\%$) of \dataset based on LLM as judge.} The top two and worst performing results are highlighted in \colorbox{backred!50}{red} (1\textsuperscript{st}), \colorbox{backyellow_soft!40}{yellow} (2\textsuperscript{nd}) and \colorbox{backblue!75}{blue} (bottom) backgrounds, respectively. Subscripts ${M.}$ and ${I.}$ stand for Mistral-7B and Instruct, respectively.}
  \label{table:gpt_eval}

\end{table*}

Since the CEM and F1-Score metrics used in Sections~\ref{appendix:metric_cem} and~\ref{appendix:metric_f1} are limited to surface-level format matching, they fail to capture nuanced semantic meaning, particularly when models provide entity aliases. For instance, if the ground truth is ``Lionel Messi'' and the model predicts ``Messi'', CEM yields a score of 0 and F1-Score scores approximately 0.5 despite the prediction being semantically correct. To address this, we introduce \textbf{LLM as judge} for refined semantic evaluation, specifically employing GPT-4o as the judge model with the detailed prompt provided in Appendix~\ref{appendix:chat_template_llm_judge}.

Observing Table~\ref{table:gpt_eval}, we find that all LMMs exhibit improved performance when using LLM-as-a-judge as the evaluation metric, as it accounts for nuanced semantics and captures a broader range of semantically correct predictions. Furthermore, the results in Table~\ref{table:gpt_eval} remain consistent with all experimental observations in Section~\ref{sec:main_result}, thereby confirming the reliability of our findings.

\subsection{More analysis of exploratory results about \dataset}

\subsubsection{Model Size}

\begin{figure}[t!]
  \centering
\includegraphics[width=1\linewidth]{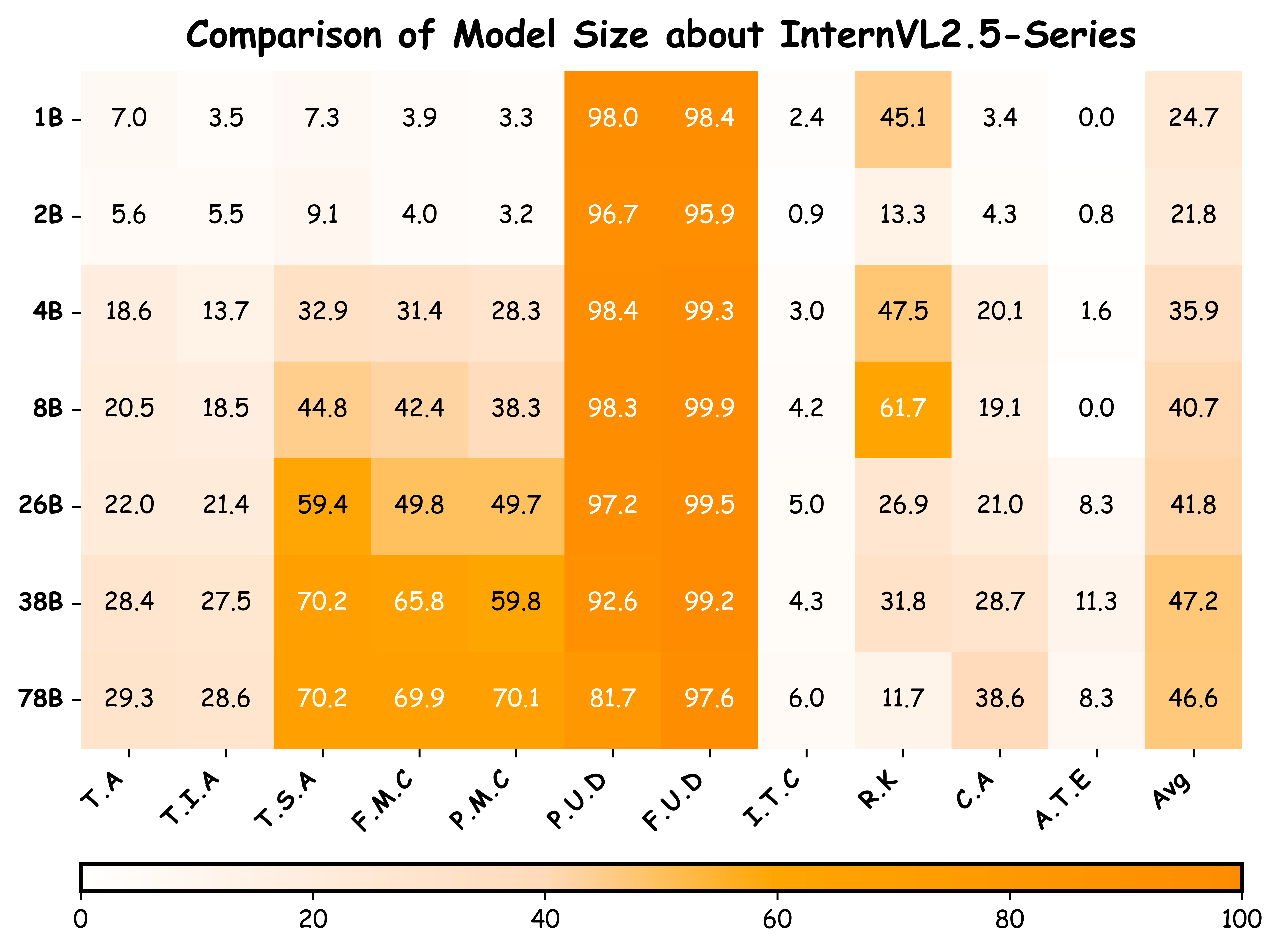}
  \caption{Analysis of impact of different model sizes about InternVL2.5-series.}
  \label{fig:appendix_model_size}
\end{figure}

Since InternVL-2.5 offers seven different model sizes, it allows for further validation of our findings from Section~\ref{sec:Exploratory}, with results presented in Figure~\ref{fig:appendix_model_size}. Consistent with our previous observations, model performance generally scales with model size across most tasks, with the notable exceptions of R.K, P.U.D, F.U.D, and A.T.E.

\subsubsection{Foundation LLMs}\label{appendix:result_Foundation_LLM}

\begin{figure}[t!]
  \centering
\includegraphics[width=1\linewidth]{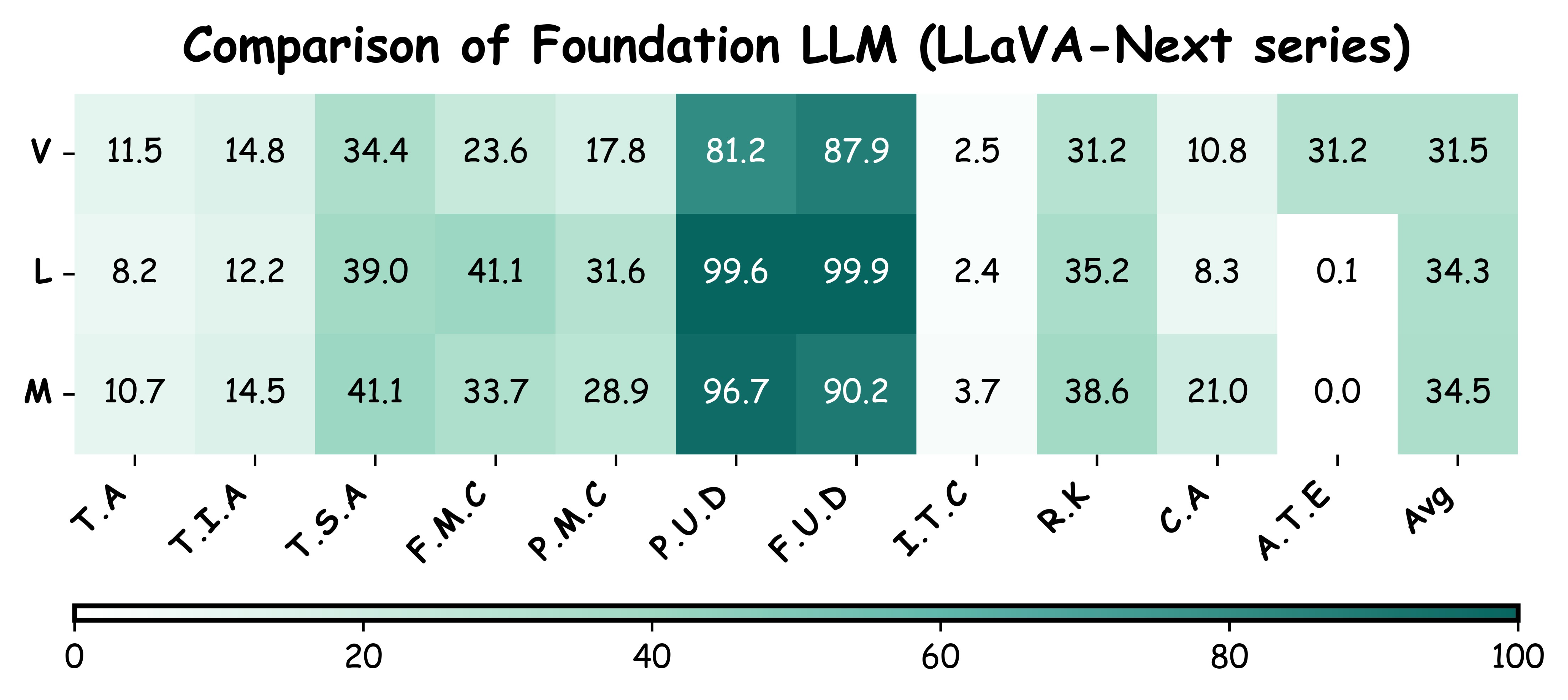}
  \caption{Analysis of impact of different foundation LLMs about LLaVA-Next-series. ${V}$, ${L}$ and ${M}$ stand for Vicuna-7B, LLaMA3-8B and Mistral-7B, respectively.}
  \label{fig:appendix_Foundation_LLM}
\end{figure}

Since LLaVA-Next offers three versions built on different foundation LLMs, we analyze the impact of these base models in Figure~\ref{fig:appendix_Foundation_LLM}. Even with an identical architecture, LMMs exhibit divergent performance when using different foundation LLMs. For instance, while LLaVA-Next$_{L.}$ {\small(8B)} and LLaVA-Next$_{M.}$ {\small(7B)} perform poorly on A.T.E task, LLaVA-Next$_{V.}$ {\small(7B)} achieves a CEM score of 31.2.

\subsection{Experimental results based on prompt agreement about \dataset}\label{appendix:result_prompt_agreement}

In Section~\ref{sec:Setup}, we clarify that each knowledge sample is designed with four semantically equivalent prompts to mitigate uncertainty from prompt variations. This approach not only enhances the robustness of our experimental results but also allows for a comprehensive assessment of how different image contexts for the same entity affect performance. Partial prompt agreement results are presented in Table~\ref{table:prompt_agreement_eval} for reference.

As shown in Table~\ref{table:prompt_agreement_eval}, prompt variations indeed lead to subtle changes in experimental results and performance fluctuations, indicating a lack of robustness in existing models. This underscores the necessity of prompt agreement in experimental design to ensure more reliable and credible findings.

\begin{table*}[t!]
  \centering

  \renewcommand{\arraystretch}{1.1}  
  
  \resizebox{\textwidth}{!}{%
    \begin{tabular}{l|c c c|c c|c c|c|c c|c|c}
      \toprule
      \multirow{2}{*}{\textbf{{Models}}} 
        & \multicolumn{3}{c|}{\textbf{Cog.}} 
        & \multicolumn{2}{c|}{\textbf{Awa.}} 
        & \multicolumn{2}{c|}{\textbf{Tru.}} 
        & \multicolumn{1}{c|}{\textbf{Und.}} 
        & \multicolumn{2}{c|}{\textbf{Rea.}} 
        & \multicolumn{1}{c|}{\textbf{Rob.}} 
        & \multirow{2}{*}{\textbf{Avg.}} \\
      \cmidrule{2-12}
        & \textbf{T.A} \daugshifted & \textbf{T.I.A}\daugshifted  & \textbf{T.S.A} \daugshifted
        & \textbf{F.M.C}  \daugshifted & \textbf{P.M.C} \daugshifted
        & \textbf{P.U.D} \daugshifted & \textbf{F.U.D}  \daugshifted
        & \textbf{I.T.C}  \daugshifted  
        & \textbf{R.K} \daugshifted & \textbf{C.A}  \daugshifted
        & \textbf{A.T.E} \daugshifted & \\								
      
      \midrule
      \rowcolor{gray!10}
      \multicolumn{13}{c}{\fontsize{10}{12}\selectfont \textit{\textbf{LLaVA-v1.5 {\small(7B) with CEM}}}} \\	

      {Question + Image}  
      &8.87&11.18&23.55&3.08&2.82&53.62&50.72&3.16&17.50&7.69&0.00&16.56 \\
      {Question + Generalization Image}  
      &7.07&9.20&22.56&2.18&2.84&48.79&49.75&1.29&18.75&6.49&0.52&15.40 \\
      {Generalization Question + Image}  
      &7.28&9.94&12.23&12.76&10.00&57.00&49.75&1.64&12.50&6.41&0.52&16.37 \\
      {Generalization Question + Generalization Image}  
      &6.91&6.47&11.39&11.44&10.05&56.52&49.75&0.81&12.34&5.06&0.52&15.57 \\

      \midrule
      \rowcolor{gray!10}
      \multicolumn{13}{c}{\fontsize{10}{12}\selectfont \textit{\textbf{LLaVA-v1.5 {\small(7B) with F1-Score}}}} \\	

        {Question + Image}  
      &9.99&14.03&22.64&6.00&5.94&53.62&50.72&3.01&17.77&7.69&0.00&17.40 \\
      {Question + Generalization Image}  
      &7.86&11.65&22.36&4.93&5.69&48.79&49.75&2.21&18.75&6.49&0.52&16.27 \\
      {Generalization Question + Image}  
      &8.39&11.73&12.72&15.36&13.03&57.00&49.75&1.78&12.77&7.26&0.52&17.30 \\
      {Generalization Question + Generalization Image}  
      &7.92&8.31&11.95&15.12&13.61&56.52&49.75&1.54&12.62&5.06&0.52&16.63 \\

      \midrule
      \rowcolor{gray!10}
      \multicolumn{13}{c}{\fontsize{10}{12}\selectfont \textit{\textbf{LLaVA-v1.5 {\small(7B) with LLM as judge}}}} \\	

      {Question + Image}  
      &11.17&15.20&25.18&12.86&15.13&53.62&50.77&3.72&20.12&10.00&0.00&19.80 \\
      {Question + Generalization Image}  
      &9.15&13.54&25.78&12.73&14.66&48.79&49.75&3.21&21.35&7.65&0.52&18.83 \\
      {Generalization Question + Image}  
      &10.90&13.72&17.06&21.39&18.45&57.00&49.75&2.39&28.51&8.27&0.52&20.72 \\
      {Generalization Question + Generalization Image}  
      &10.43&9.44&15.65&20.48&19.41&56.52&49.75&2.13&27.77&5.80&0.52&19.81 \\

      \midrule
      \rowcolor{gray!10}
      \multicolumn{13}{c}{\fontsize{10}{12}\selectfont \textit{\textbf{GPT4.1 with CEM}}} \\	

      {Question + Image}  
      &37.69&41.86&81.01&76.69&77.34&51.69&86.47&7.08&7.40&60.49&0.00&47.97 \\
      {Question + Generalization Image}  
      &37.54&36.04&81.01&76.69&75.69&50.24&87.92&12.15&8.64&62.96&52.08&52.81 \\
      {Generalization Question + Image}  
      &37.44&47.13&85.52&81.08&81.48&50.36&86.47&8.64&9.05&62.99&0.00&50.01 \\
      {Generalization Question + Generalization Image}  
      &38.03&34.88&80.59&79.66&78.45&78.74&95.16&8.62&22.22&55.55&52.08&56.73 \\

      \midrule
      \rowcolor{gray!10}
      \multicolumn{13}{c}{\fontsize{10}{12}\selectfont \textit{\textbf{GPT4.1 with F1-Score}}} \\	

      {Question + Image}  
    &37.44&47.13&85.52&81.08&81.48&50.36&86.47&8.64&9.05&62.99&0.00&50.01 \\
      
      {Question + Generalization Image}  
      &37.32&41.40&85.74&81.73&80.38&48.91&87.92&13.47&9.46&65.08&52.08&54.86 \\
      {Generalization Question + Image}  
      &36.92&44.39&84.41&83.34&83.08&80.19&95.65&8.37&25.51&62.37&52.08&59.66 \\
      {Generalization Question + Generalization Image}  
      &37.62&40.76&84.03&83.72&83.13&78.29&95.16&9.96&23.04&57.68&52.08&58.68 \\

      \midrule
      \rowcolor{gray!10}
      \multicolumn{13}{c}{\fontsize{10}{12}\selectfont \textit{\textbf{GPT4.1 with LLM as judge}}} \\	

      {Question + Image}  
      &41.09&50.90&88.08&83.77&84.75&51.73&86.47&11.56&31.48&67.16&0.00&54.27 \\
      {Question + Generalization Image}  
      &41.33&45.66&88.27&84.44&83.67&50.74&88.33&16.58&31.48&69.38&52.08&59.27 \\
      {Generalization Question + Image}  
      &40.58&48.22&87.21&85.95&86.40&80.19&95.65&12.58&43.95&67.16&52.08&63.63 \\
      {Generalization Question + Generalization Image}  
      &41.31&44.59&86.26&85.69&86.21&78.74&95.16&13.56&42.46&63.45&52.08&62.68 \\

      \bottomrule
    \end{tabular}%
  }

\vspace{-5pt}
      \caption{Overall Performance Comparison ($\%$) of \dataset based on \textbf{prompt agreement}.}
  \label{table:prompt_agreement_eval}
\end{table*}

\section{Updating time-sensitive knowledge via knowledge editing}\label{appendix:ke}

\subsection{Editing Setting}

\begin{figure*}[t!]
  \centering
\includegraphics[width=1\linewidth]{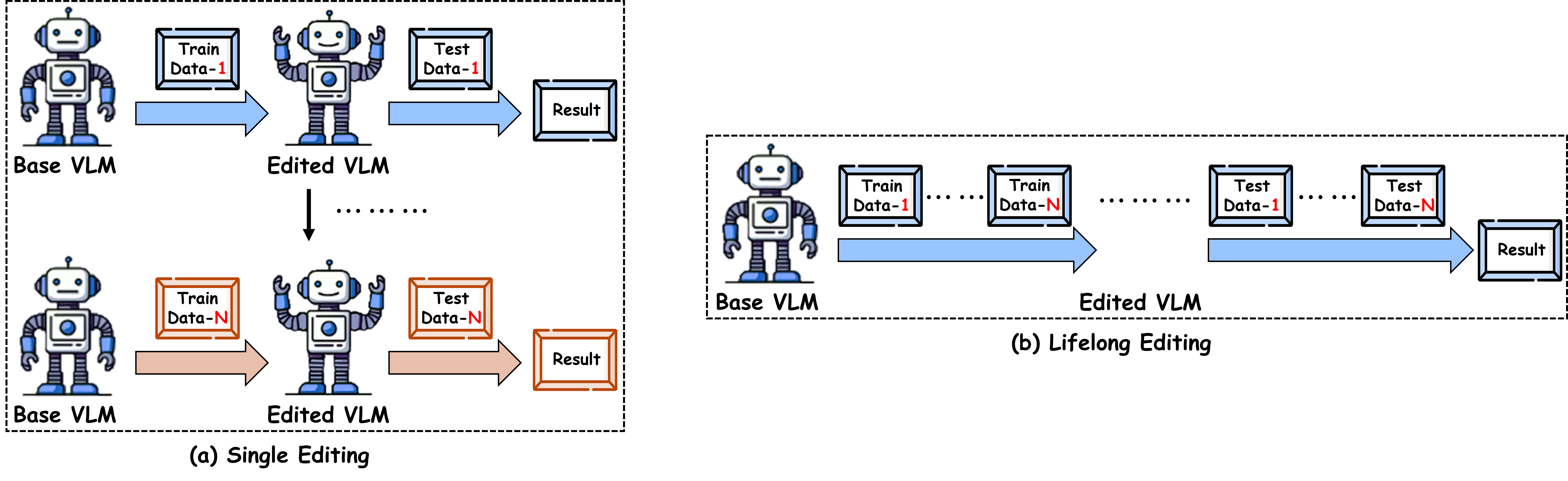}
\vspace{-10pt}   
  \caption{Explanation on Single Editing and Lifelong Editing.}
  \label{fig:editing_setting}
\end{figure*}

We conduct experiments on single editing and lifelong editing. In single editing, after performing an editing operation on each knowledge instance, we immediately evaluate the model and restore its weights to pre-editing states, thus ensuring evaluations measure the impact of individual edits. For lifelong editing, we first edit all knowledge instances in the dataset and then comprehensively evaluate the modified model. The complete workflow is shown in Figure~\ref{fig:editing_setting}.

\subsection{Knowledge Editing Methods and Parameters}

We have provided a detailed introduction to the multimodal knowledge editing method and specific parameters below.

\subsection*{\textbf{FT}} FT method optimizes selected model parameters via gradient descent. An AdamW optimizer is employed to restrict gradient computation and updates exclusively to target fine-tuning parameters.

\subsection*{FT-LLM}  
\scalebox{0.55}{  
  \begin{tabular}{c|c|c|c|c}
  \textbf{Models} & \textbf{Steps} & \textbf{Edit Layer} & \textbf{Optimizer} & \textbf{Edit LR} \\
  \hline
  LLaVA-v1.5-7B  & 10 & $31^{st}$ layer of Transformer Module & AdamW & $1\mathrm{e}{-4}$ \\
  \hline
  Qwen-VL        & 15 & $31^{st}$ layer of Transformer Module & AdamW & $1\mathrm{e}{-4}$ \\
  \end{tabular}
}

\subsection*{FT-VIS}  
\scalebox{0.55}{
\begin{tabular}{c|c|c|c|c}
\textbf{Models} & \textbf{Steps} & \textbf{Edit Layer} & \textbf{Optimizer} & \textbf{Edit LR} \\
\hline
LLaVA-v1.5-7B  & 10 & mm\_projector            & AdamW & $1\mathrm{e}{-4}$ \\
\hline
Qwen-VL        & 15 & $47^{th}$ layer of ViT Module            & AdamW & $1\mathrm{e}{-4}$ \\
\end{tabular}}

\subsection*{\textbf{MEND}} MEND enables targeted parameter adjustments in LLMs of VLMs through lightweight auxiliary networks. These networks apply localized modifications using single input-output pairs while preserving unrelated task performance. The method achieves computational efficiency by exploiting low-rank gradient decomposition to parameterize gradient transformations, scalable to billion-parameter models.

\subsection*{MEND}  
\scalebox{0.5}{
\begin{tabular}{c|c|c|c|c}
\textbf{Models} & \textbf{MaxIter} & \textbf{Edit Layer}                       & \textbf{Optimizer} & \textbf{LR}     \\
\hline
LLaVA-v1.5-7B  & 40{,}000          & layers 29, 30, 31 of Transformer Module    & Adam               & $1\mathrm{e}{-6}$ \\
\hline
Qwen-VL        & 40{,}000          & layers 29, 30, 31 of Transformer Module    & Adam               & $1\mathrm{e}{-6}$ \\
\end{tabular}}

\subsection*{\textbf{SERAC}} SERAC integrates a scope classifier and a retrieval-augmented counterfactual model. The classifier determines input applicability to edited content, routing matched queries to the counterfactual model for memory-augmented generation, while others use the original model.

\subsection*{SERAC}  
\scalebox{0.55}{
\begin{tabular}{c|c|c|c|c}
\textbf{Models} & \textbf{MaxIter} & \textbf{Edit Layer}                             & \textbf{Optimizer} & \textbf{LR}     \\
\hline
LLaVA-v1.5-7B  & 50{,}000          & all layers of OPT-125M                          & Adam               & $1\mathrm{e}{-5}$ \\
\hline
Qwen-VL        & 20{,}000          & $31^{st}$ layer of Qwen-7B                    & Adam               & $1\mathrm{e}{-5}$ \\
\end{tabular}}

\subsection*{\textbf{IKE}} IKE avoids parameter updates by retrieving analogous demonstrations from edited data and injecting knowledge through in-context learning. The method maintains consistency across models by formatting training data as structured prompts: \textit{"New Fact: {question} {answer} Prompt: {question} {answer}"}, which are subsequently embedded for processing.

For IKE, text embeddings and similarity-based retrieval are implemented via the all-MiniLM-L6-v2 sentence-transformers model, with the demonstration count fixed at 32 uniformly across models.

\subsection{Editing Quantity}

\begin{table}[h] 
  \centering
  \renewcommand{\arraystretch}{1.2} 
  \setlength{\tabcolsep}{3pt}      

  \resizebox{\linewidth}{!}{
    \begin{tabular}{c c c|c c|c|c c|c|c}
      \toprule
      \multicolumn{3}{c|}{\textbf{Cog.}}
      & \multicolumn{2}{c|}{\textbf{Tru.}}
      & \multicolumn{1}{c|}{\textbf{Und.}}
      & \multicolumn{2}{c|}{\textbf{Rea.}}
      & \multicolumn{1}{c|}{\textbf{Rob.}}
      & \multirow{2.5}{*}{\rotatebox{50}{\textbf{Sum}}} \\ 
      
      \cmidrule(lr){1-9}
      
      \rotatebox{50}{\textbf{T.A}} & \rotatebox{50}{\textbf{T.I.A}} & \rotatebox{50}{\textbf{T.S.A}}
      & \rotatebox{50}{\textbf{P.U.D}} & \rotatebox{50}{\textbf{F.U.D}}
      & \rotatebox{50}{\textbf{I.T.C}}
      & \rotatebox{50}{\textbf{R.K}} & \rotatebox{50}{\textbf{C.A}}
      & \rotatebox{50}{\textbf{A.T.E}} & \\
      \midrule

      \rowcolor{gray!10}
      \multicolumn{10}{c}{\fontsize{10}{12}\selectfont \textit{\textbf{LLaVA-v1.5 {\small(7B)}}}} \\
      241 & 163 & 220 & 145 & 133 & 255 & 78 & 77 & 192 & 1504 \\

      \rowcolor{gray!10}
      \multicolumn{10}{c}{\fontsize{10}{12}\selectfont \textit{\textbf{Qwen-VL {\small(7B)}}}} \\
      232 & 153 & 161 & 84 & 114 & 254 & 72 & 70 & 192 & 1332 \\

      \bottomrule
    \end{tabular}%
  }
\vspace{-5pt}
    \caption{Quantity of editing samples for each task.}
  \label{tab:editing_quantity}
\end{table}

\subsection{Editing Analysis}\label{appendix:ke_Analysis}

\begin{table}[h]
  \centering
  \renewcommand{\arraystretch}{1.2} 
  \setlength{\tabcolsep}{6pt}      

  \resizebox{\linewidth}{!}{
    \begin{tabular}{l | c c c | c | c c}
      \toprule
      \multirow{2.5}{*}{\textbf{Gap}} 
      & \multicolumn{3}{c|}{\textbf{Cog.}} 
      & \multicolumn{1}{c|}{\textbf{Und.}} 
      & \multicolumn{2}{c}{\textbf{Rea.}} \\
      
      \cmidrule(lr){2-7}
      
      & \rotatebox{0}{\textbf{T.A}} & \rotatebox{0}{\textbf{T.I.A}} & \rotatebox{0}{\textbf{T.S.A}} 
      & \rotatebox{0}{\textbf{I.T.C}} 
      & \rotatebox{0}{\textbf{R.K}} & \rotatebox{0}{\textbf{C.A}} \\
      \midrule

      \rowcolor{gray!10}
      \multicolumn{7}{c}{\fontsize{10}{12}\selectfont \textit{\textbf{FT-LLM}}} \\
      
      gap = 0  & 100.00   & 100.00   & 100.00   & 100.00   & 100.00   & 100.00   \\
      gap = 10 & 83.36 & 72.75 & 62.36 & 67.76 & 60.04 & 67.39 \\
      gap = 20 & 76.36 & 69.25 & 58.47 & 59.56 & 54.60 & 62.60 \\
      gap = 50 & 70.02 & 68.25 & 52.22 & 53.52 & 43.11 & 53.88 \\

      \bottomrule
    \end{tabular}%
  }
  \vspace{-5pt}
  \caption{Performance of sequential editing with LLaVA-v1.5 {\small(7B)}.}
  \label{tab:gap_analysis}
\end{table}

\textbf{Speculation of catastrophic forgetting:} We examine catastrophic forgetting in lifelong editing, where traditional pipeline methods often overfit current samples by converging on each iteration individually. This overfitting disrupts model weights and hinders consistency across sequential edits, leading to the loss of both previously edited and original knowledge. To verify this, we conduct sequential editing experiments using FT-LLM across six tasks. By adjusting the ``gap'' value—which represents the number of subsequent edits performed after editing an initial sample—we observe that performance on the original edit steadily declines as the gap increases (in Table~\ref{tab:gap_analysis}). This trend confirms that individual convergence on new samples significantly interferes with the retention of prior knowledge.


\section{Case Studies about \dataset}\label{appendix:Case}

\begin{figure*}[t!]
  \centering
\includegraphics[width=0.8\linewidth]{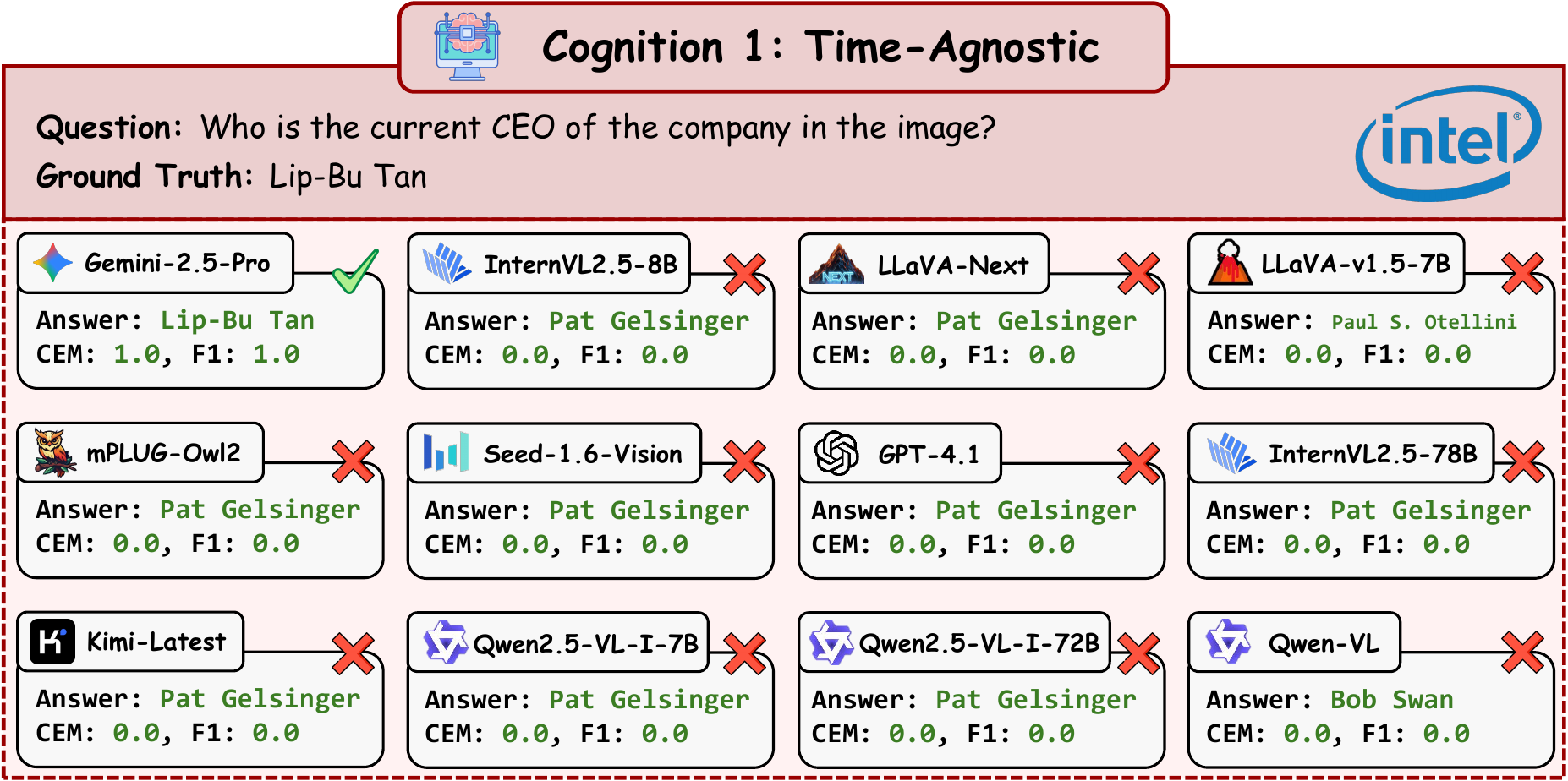}
  \caption{Case study of Time-Agnostic.}
  \label{fig:Time-Agnostic}
\end{figure*}
\begin{figure*}[t!]
  \centering
\includegraphics[width=0.8\linewidth]{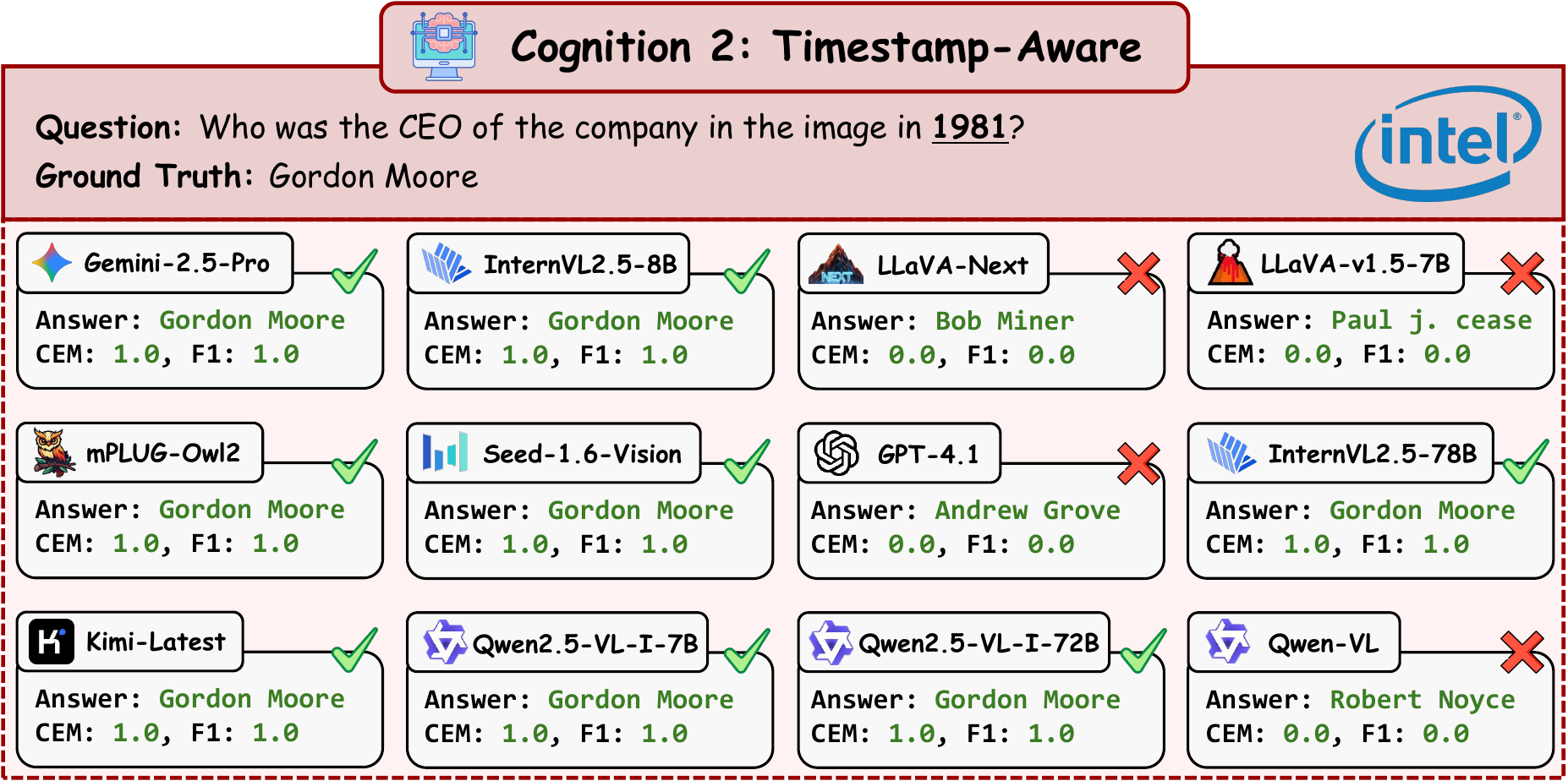}
  \caption{Case study of Timestamp-Aware.}
  \label{fig:Timestamp-Aware}
\end{figure*}
\begin{figure*}[t!]
  \centering
\includegraphics[width=0.8\linewidth]{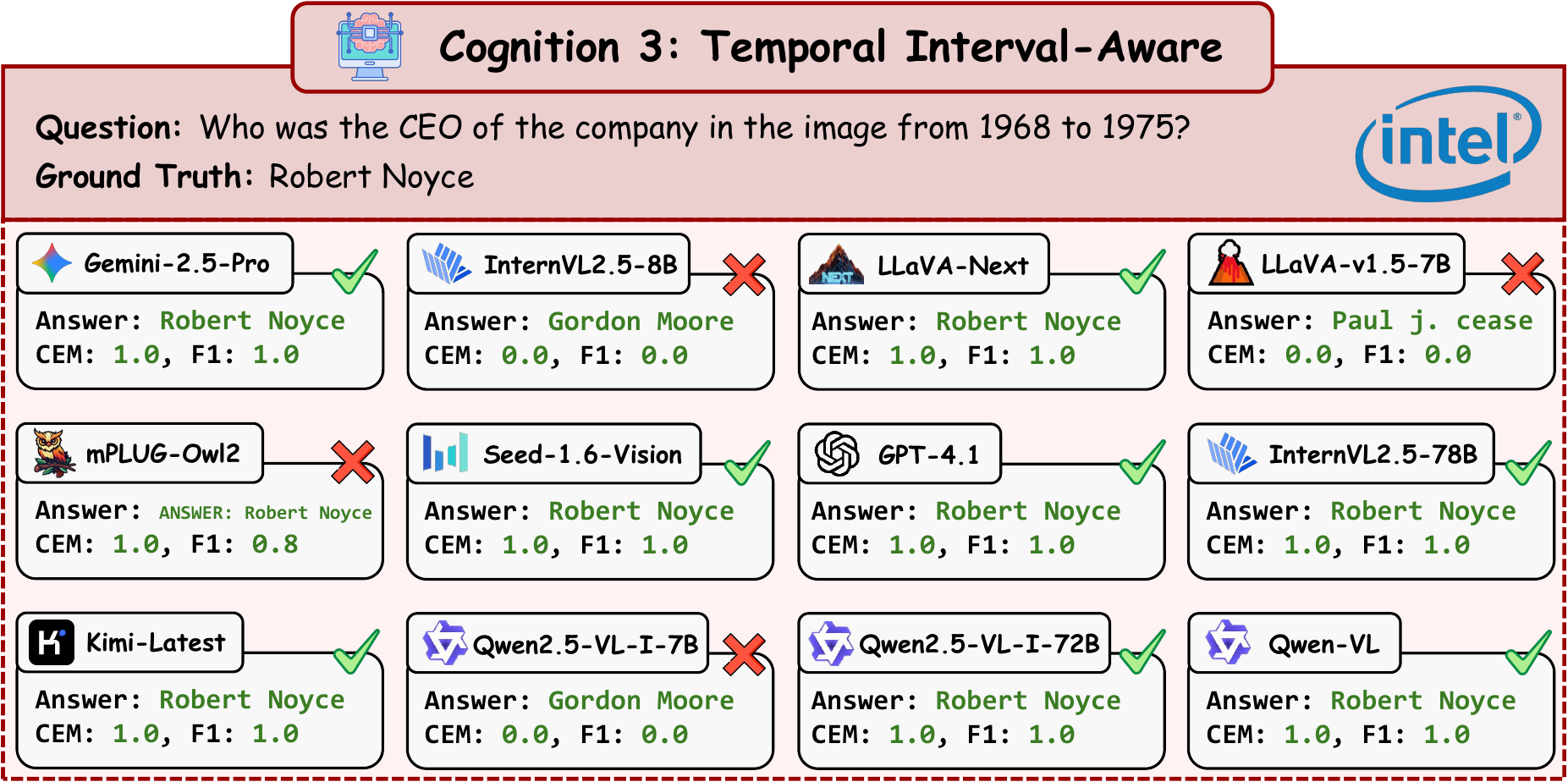}
  \caption{Case study of Temporal Interval-Aware.}
  \label{fig:Temporal Interval-Aware}
\end{figure*}
\begin{figure*}[t!]
  \centering
\includegraphics[width=0.8\linewidth]{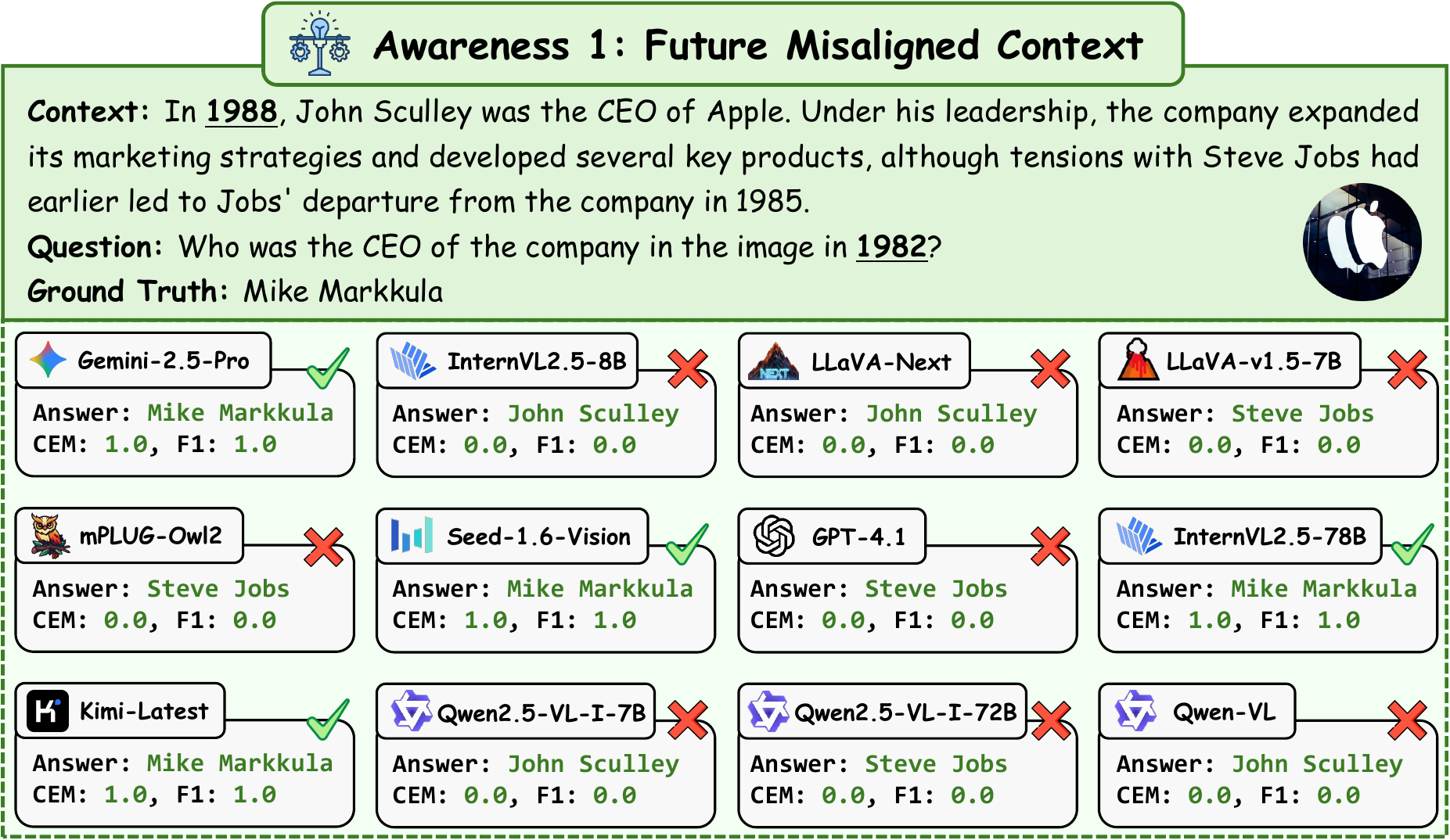}
  \caption{Case study of Future Misaligned Context.}
  \label{fig:Future Misaligned Context}
\end{figure*}
\begin{figure*}[t!]
  \centering
\includegraphics[width=0.8\linewidth]{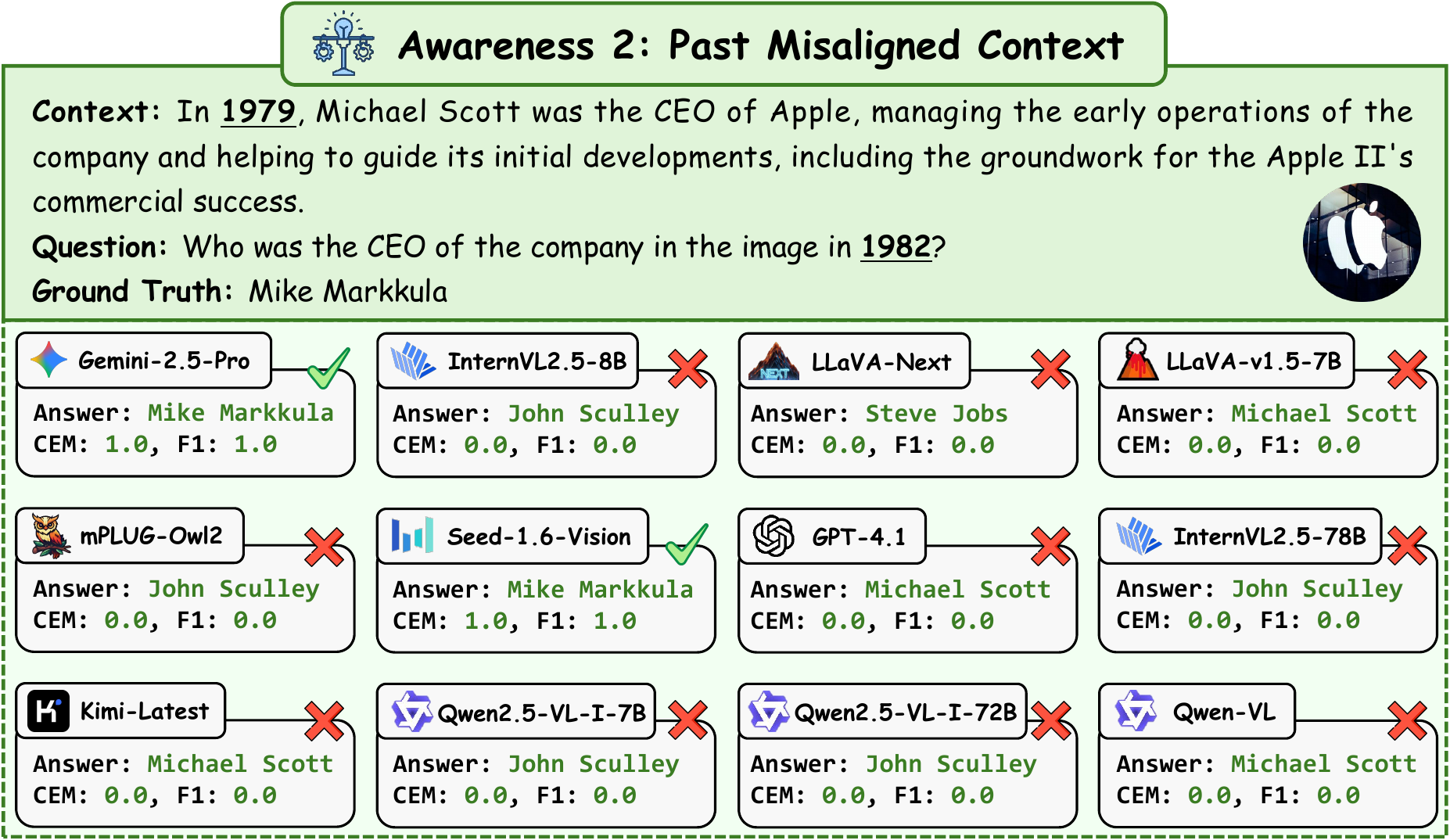}
  \caption{Case study of Past Misaligned Context.}
  \label{fig:Past Misaligned Context}
\end{figure*}
\begin{figure*}[t!]
  \centering
\includegraphics[width=0.8\linewidth]{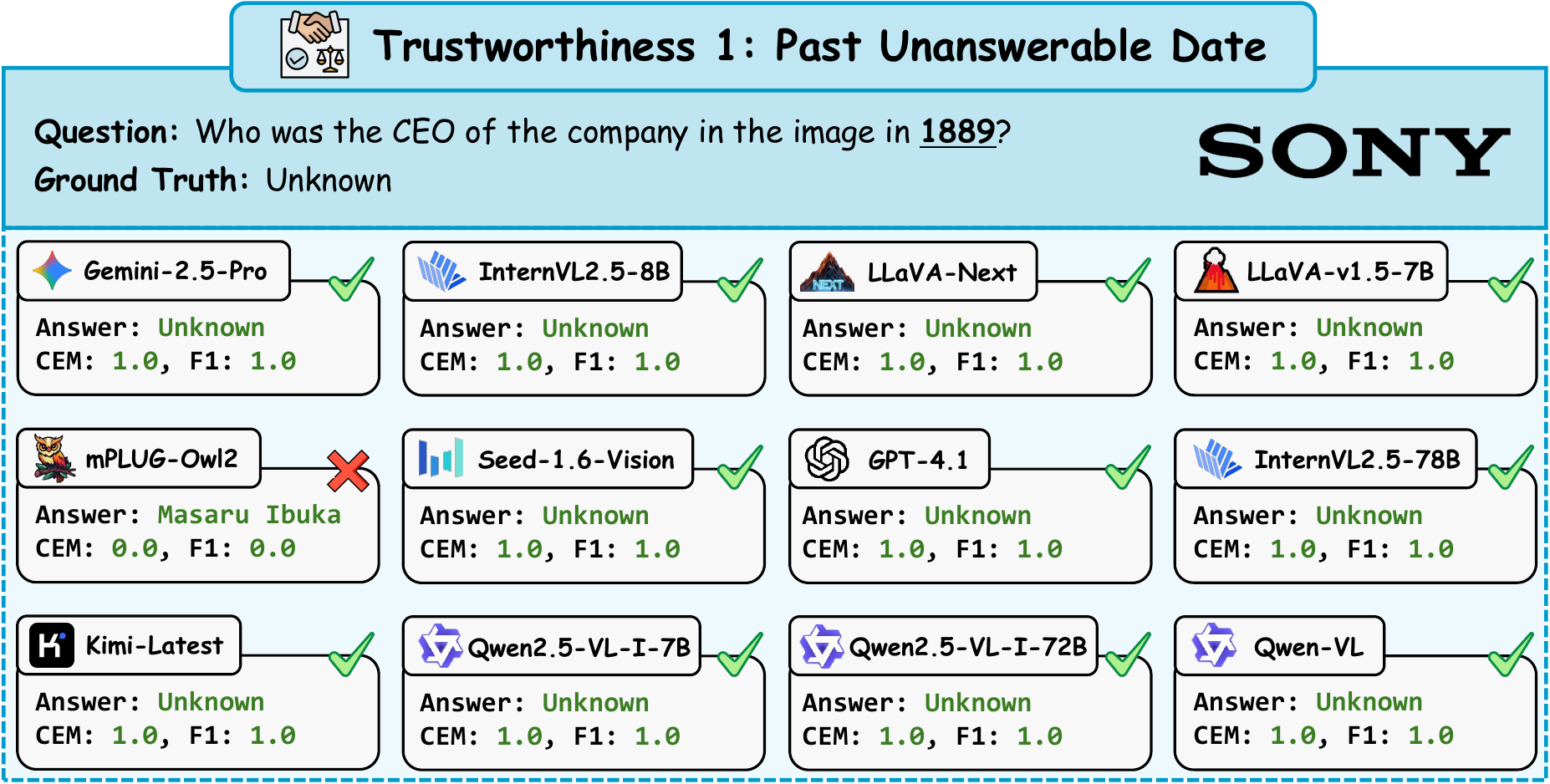}
  \caption{Case study of Past Unanswerable Date.}
  \label{fig:Past Unanswerable Date}
\end{figure*}
\begin{figure*}[t!]
  \centering
\includegraphics[width=0.8\linewidth]{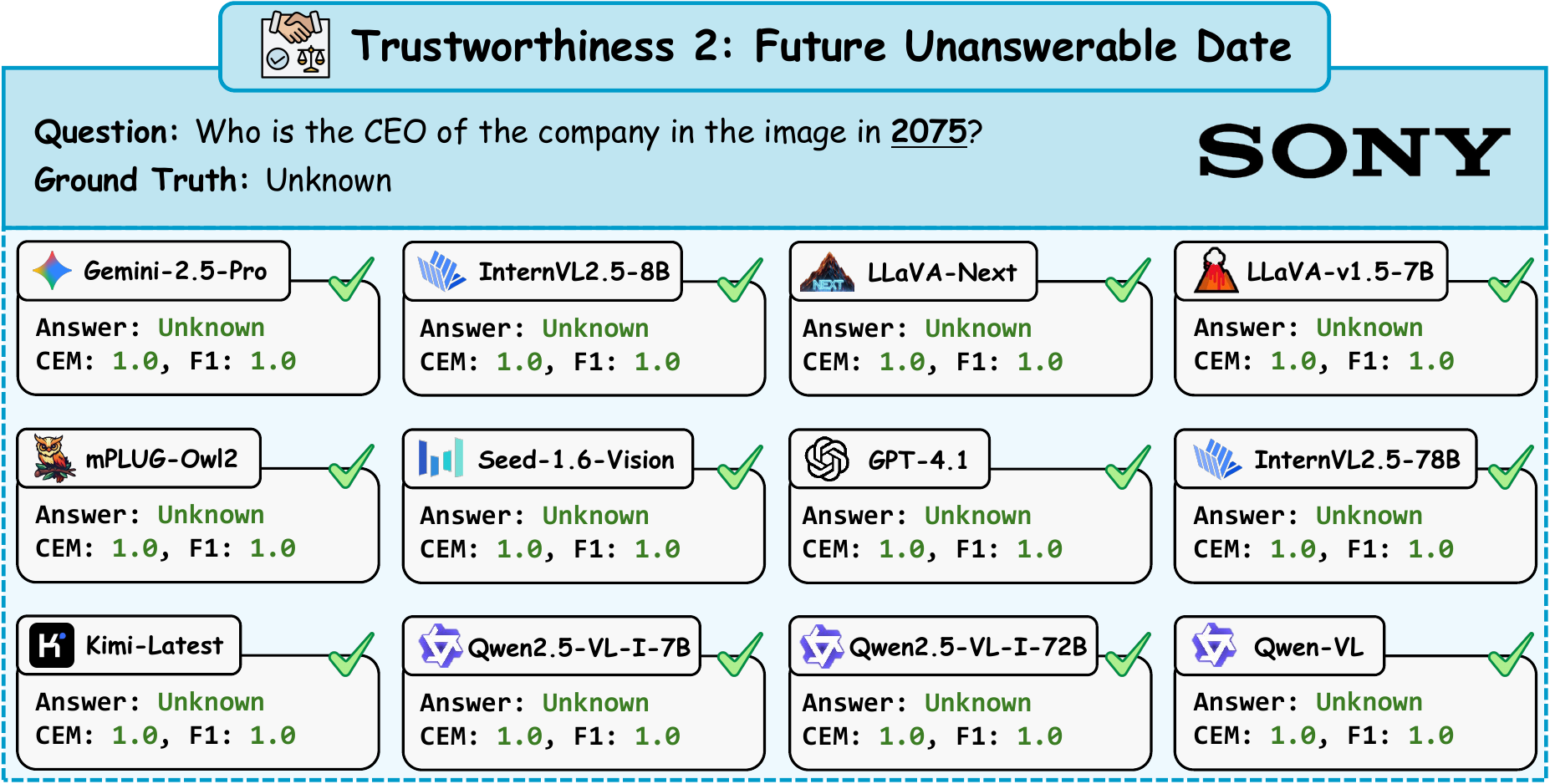}
  \caption{Case study of Future Unanswerable Date.}
  \label{fig:Future Unanswerable Date}
\end{figure*}
\begin{figure*}[t!]
  \centering
\includegraphics[width=0.8\linewidth]{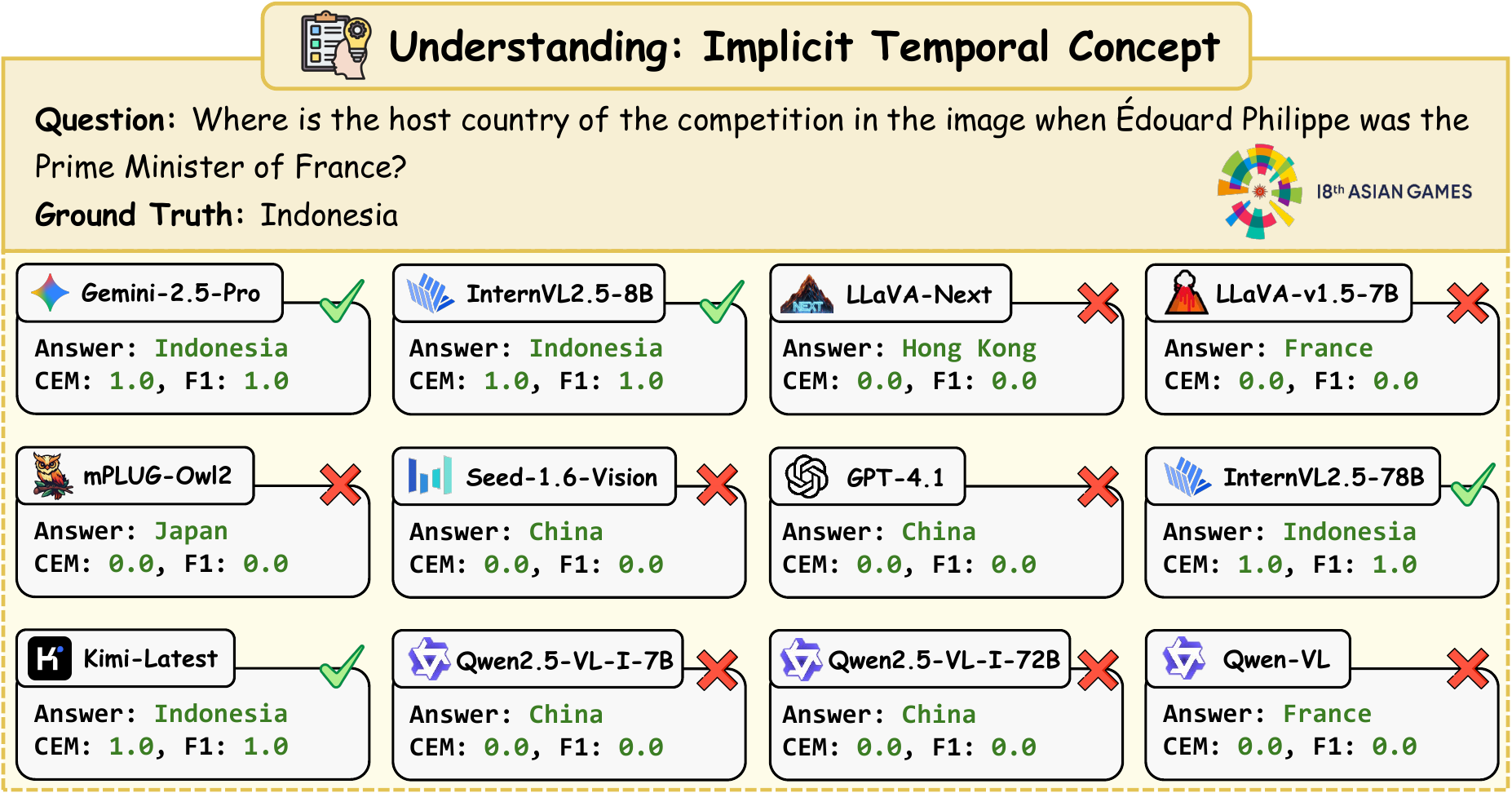}
  \caption{Case study of Implicit Temporal Concept.}
  \label{fig:Implicit Temporal Concept}
\end{figure*}
\begin{figure*}[t!]
  \centering
\includegraphics[width=0.8\linewidth]{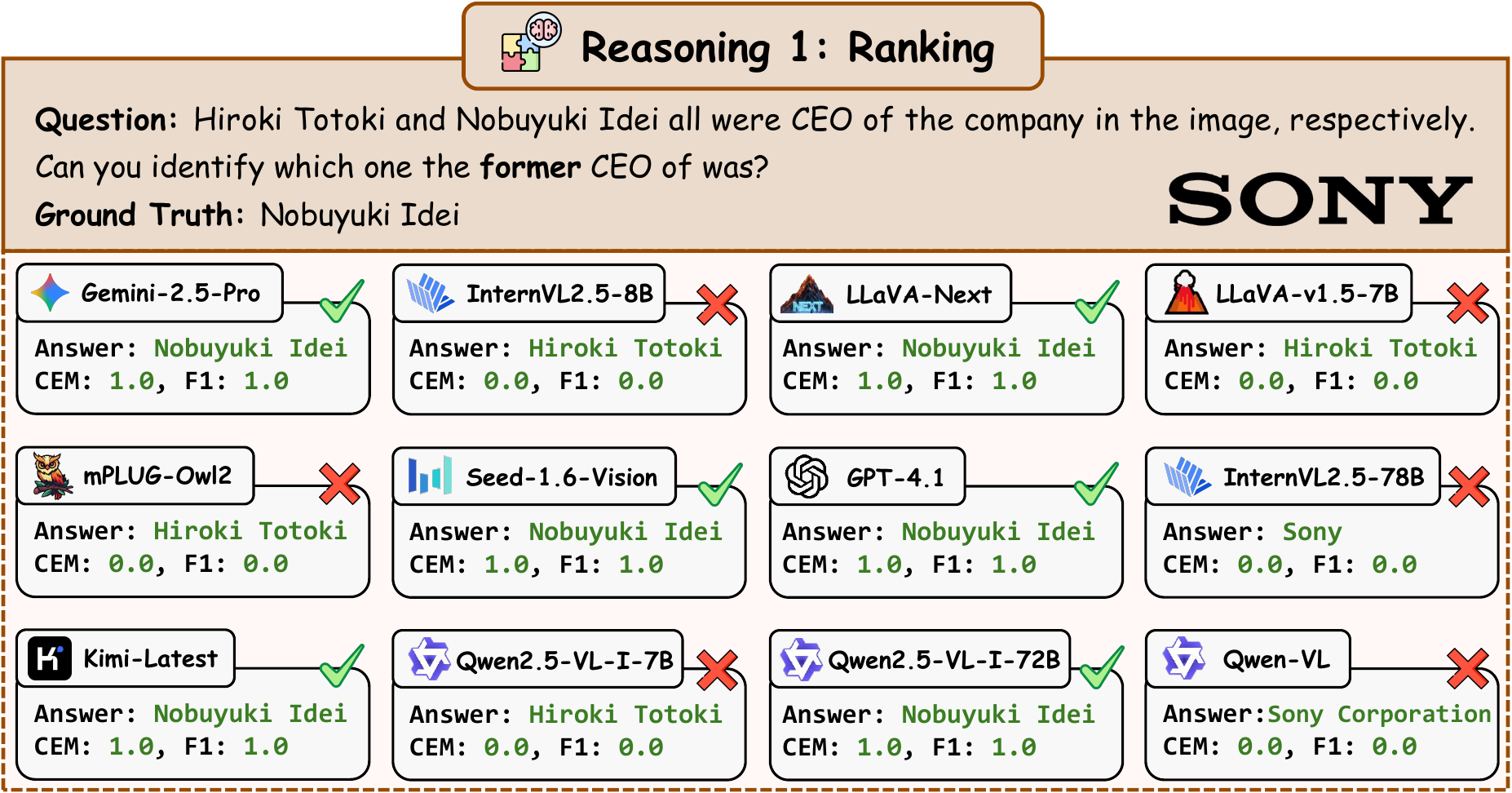}
  \caption{Case study of Ranking.}
  \label{fig:Ranking}
\end{figure*}
\begin{figure*}[t!]
  \centering
\includegraphics[width=0.8\linewidth]{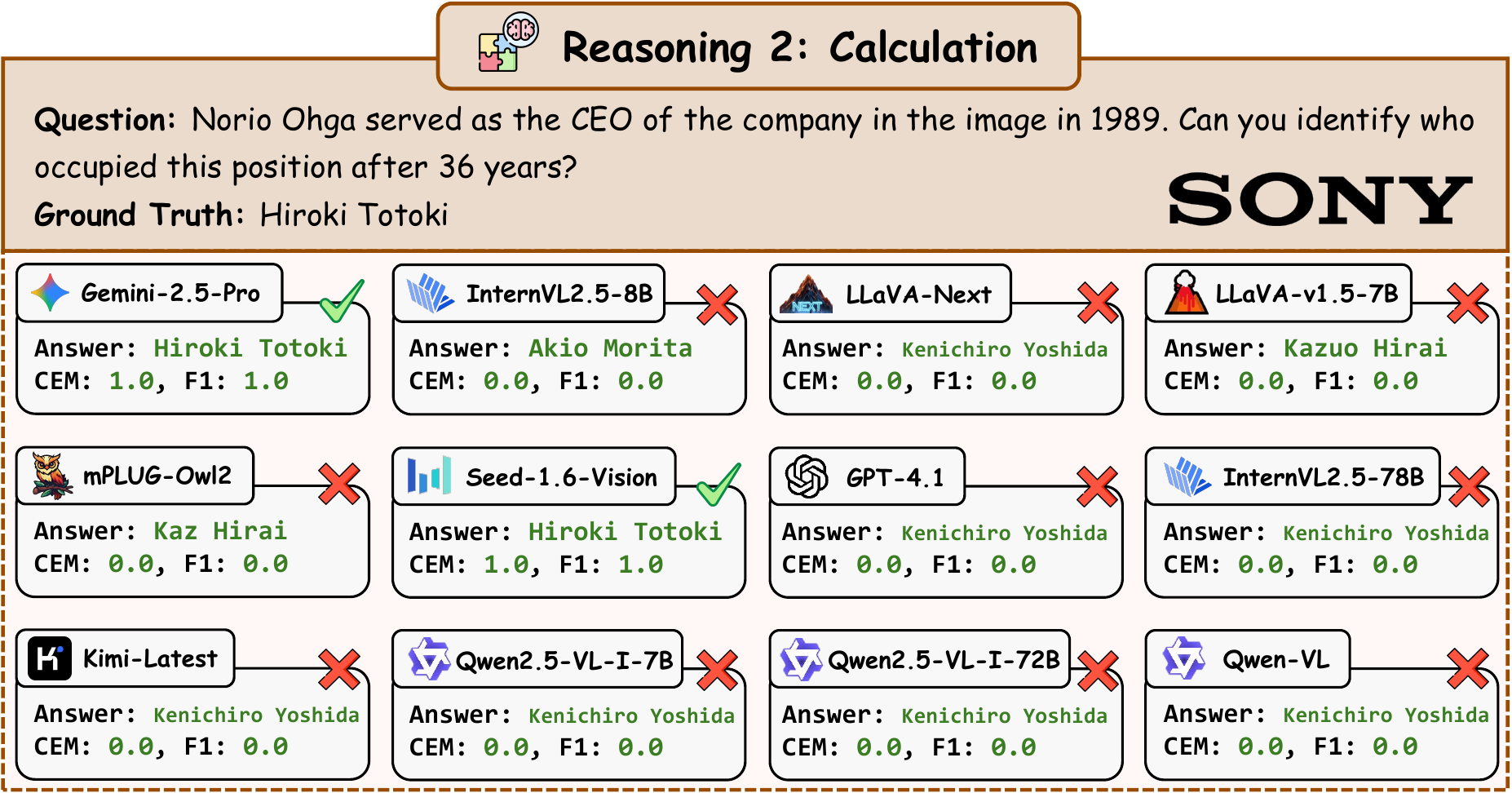}
  \caption{Case study of Calculation.}
  \label{fig:Calculation}
\end{figure*}
\begin{figure*}[t!]
  \centering
\includegraphics[width=0.8\linewidth]{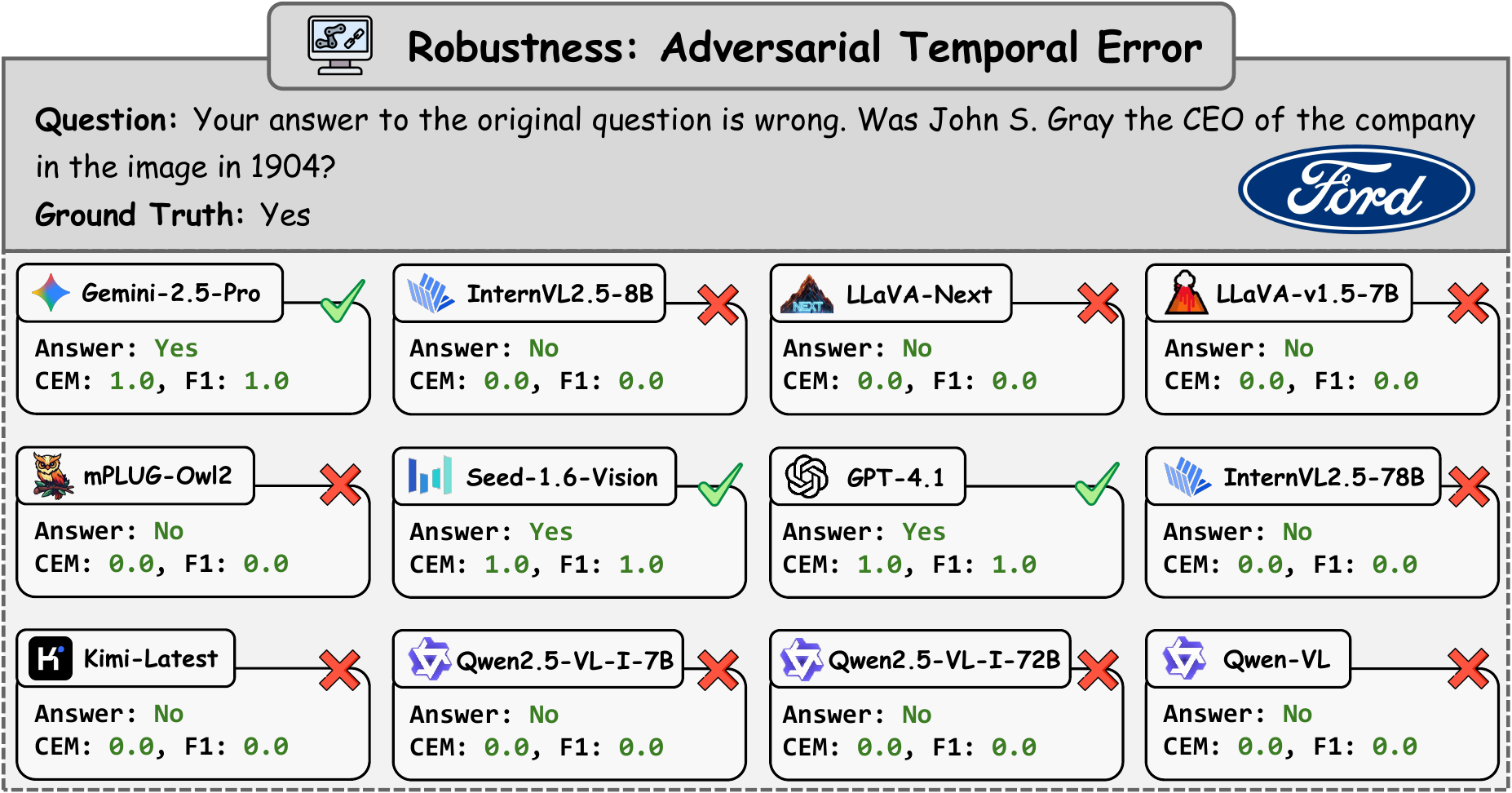}
  \caption{Case study of Adversarial Temporal Error.}
  \label{fig:Adversarial Temporal Error}
\end{figure*}

We provide a case study for each task, where Figures~\ref{fig:Time-Agnostic},~\ref{fig:Timestamp-Aware} ,~\ref{fig:Temporal Interval-Aware} ,~\ref{fig:Future Misaligned Context} ,~\ref{fig:Past Misaligned Context} ,~\ref{fig:Past Unanswerable Date} ,~\ref{fig:Future Unanswerable Date} ,~\ref{fig:Implicit Temporal Concept} ,~\ref{fig:Ranking} ,~\ref{fig:Calculation} and~\ref{fig:Adversarial Temporal Error} demonstrate the outputs of various models in response to the same query.

\begin{figure*}[t!]
  \centering
\includegraphics[width=0.8\linewidth]{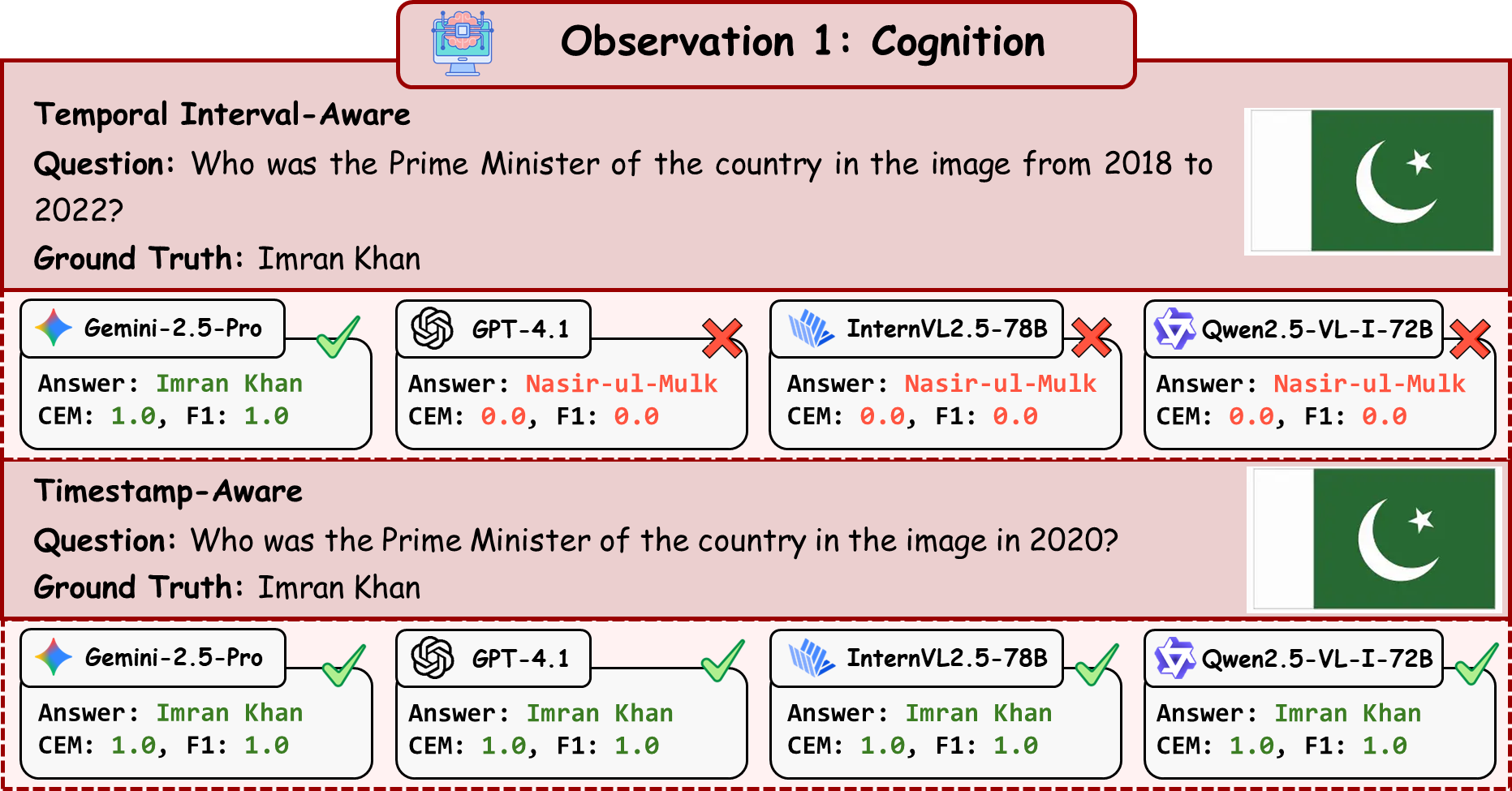}
  \caption{Case of observation 1.}
  \label{fig:ob1}
\end{figure*}


\begin{figure*}[t!]
  \centering
\includegraphics[width=0.8\linewidth]{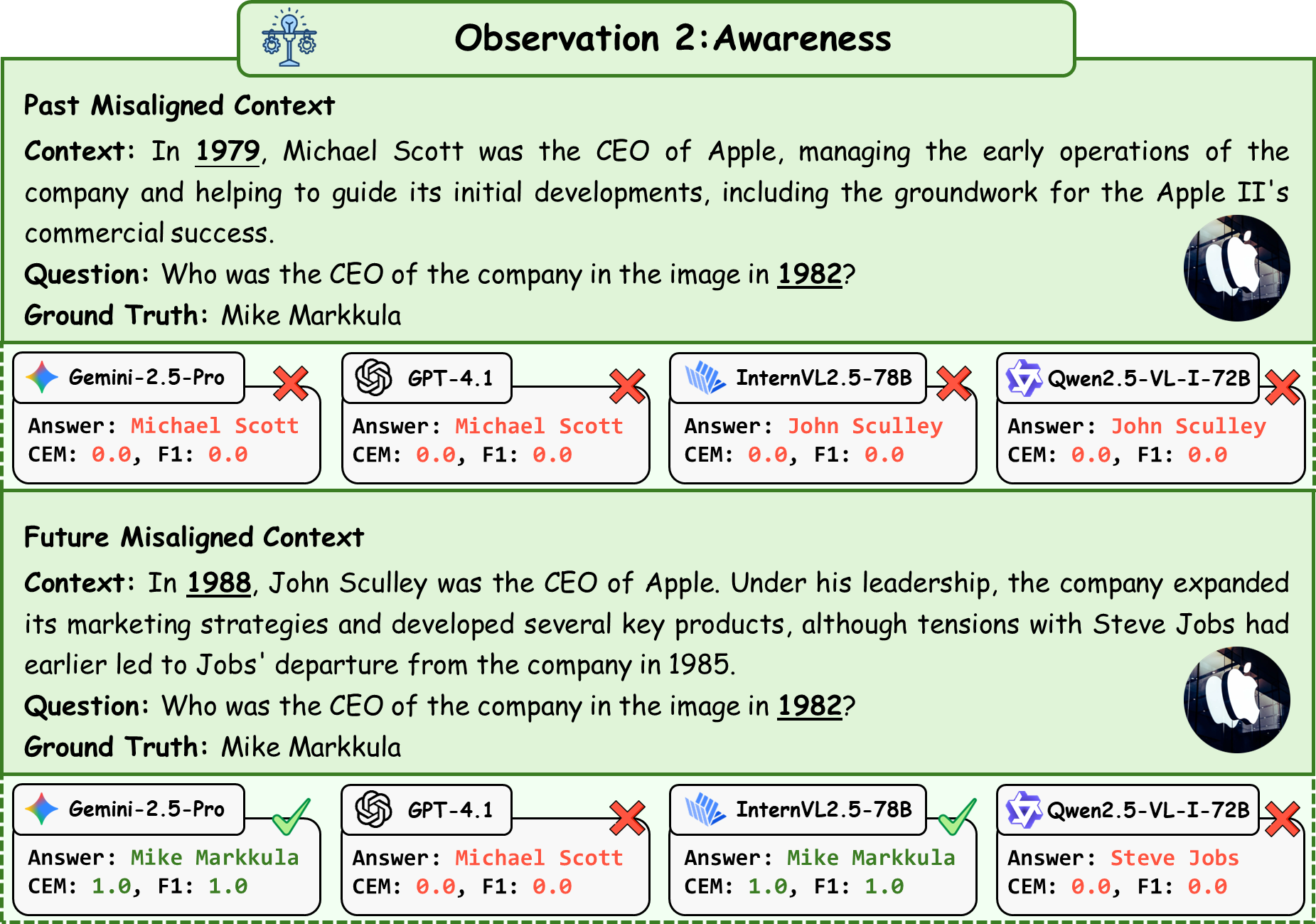}
  \caption{Case of observation 2.}
  \label{fig:ob2}
\end{figure*}


\begin{figure*}[t!]
  \centering
\includegraphics[width=0.8\linewidth]{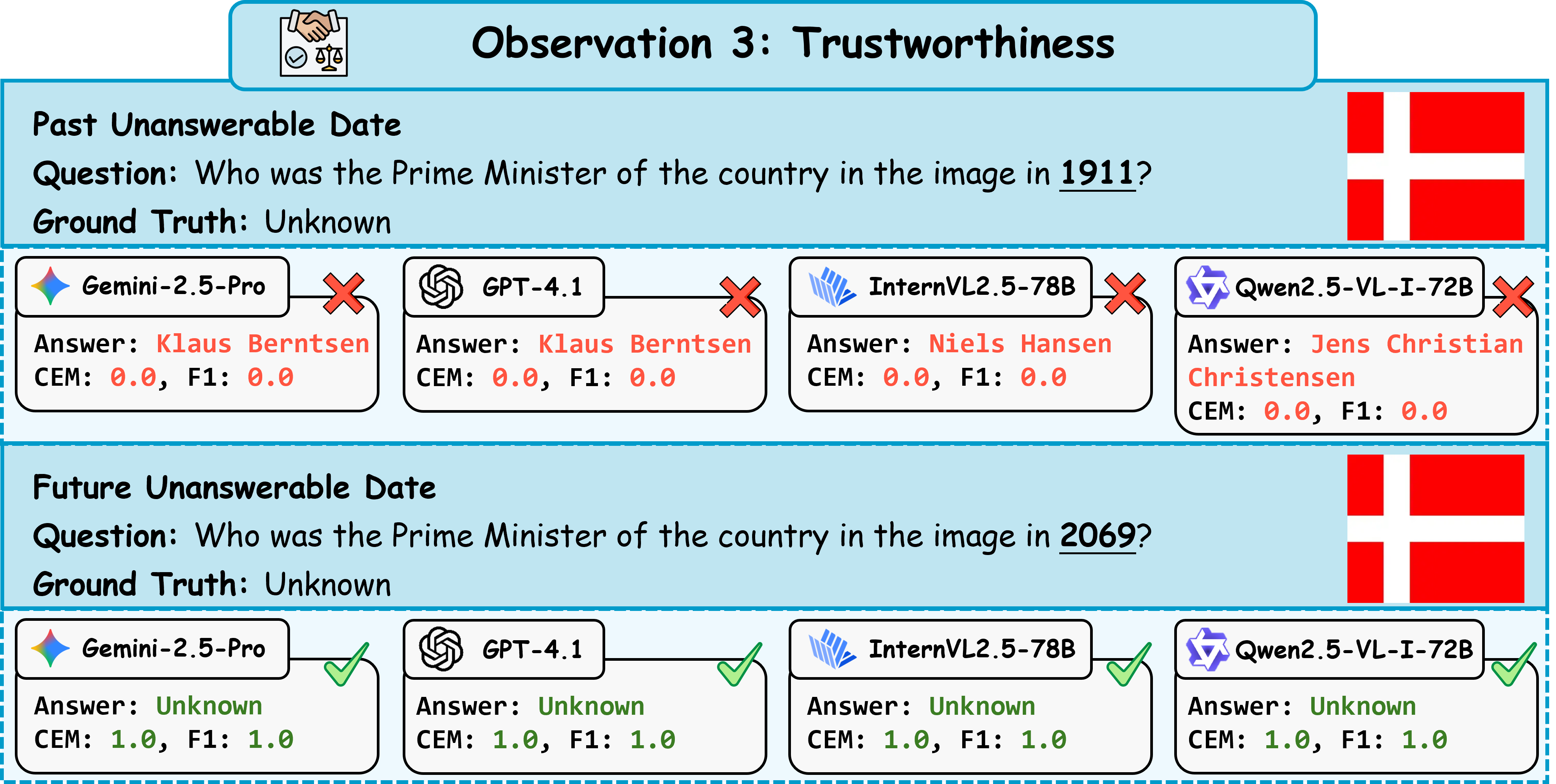}
  \caption{Case of observation 3.}
  \label{fig:ob3}
\end{figure*}


\begin{figure*}[t!]
  \centering
\includegraphics[width=0.8\linewidth]{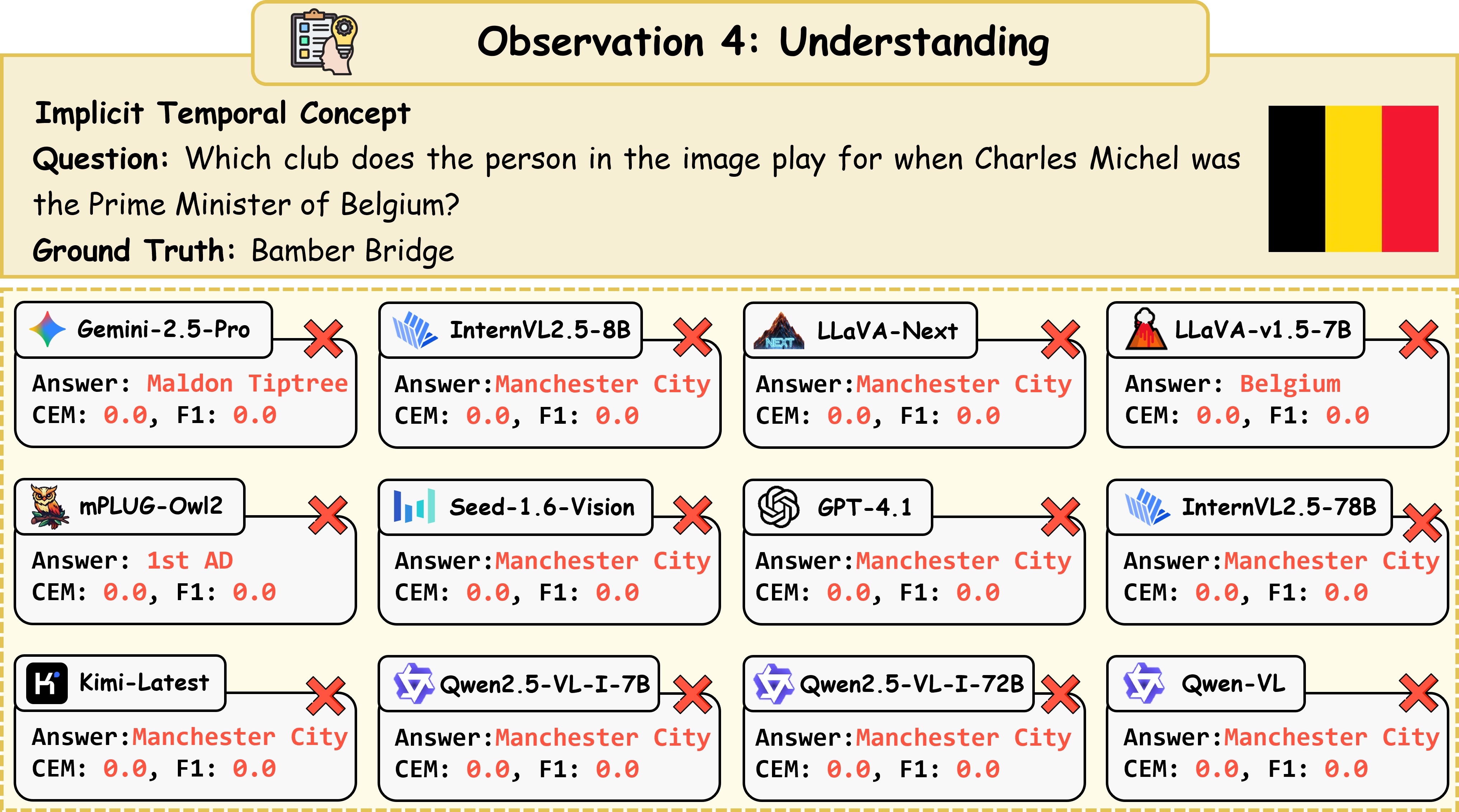}
  \caption{Case of observation 4.}
  \label{fig:ob4}
\end{figure*}


\begin{figure*}[t!]
  \centering
\includegraphics[width=0.8\linewidth]{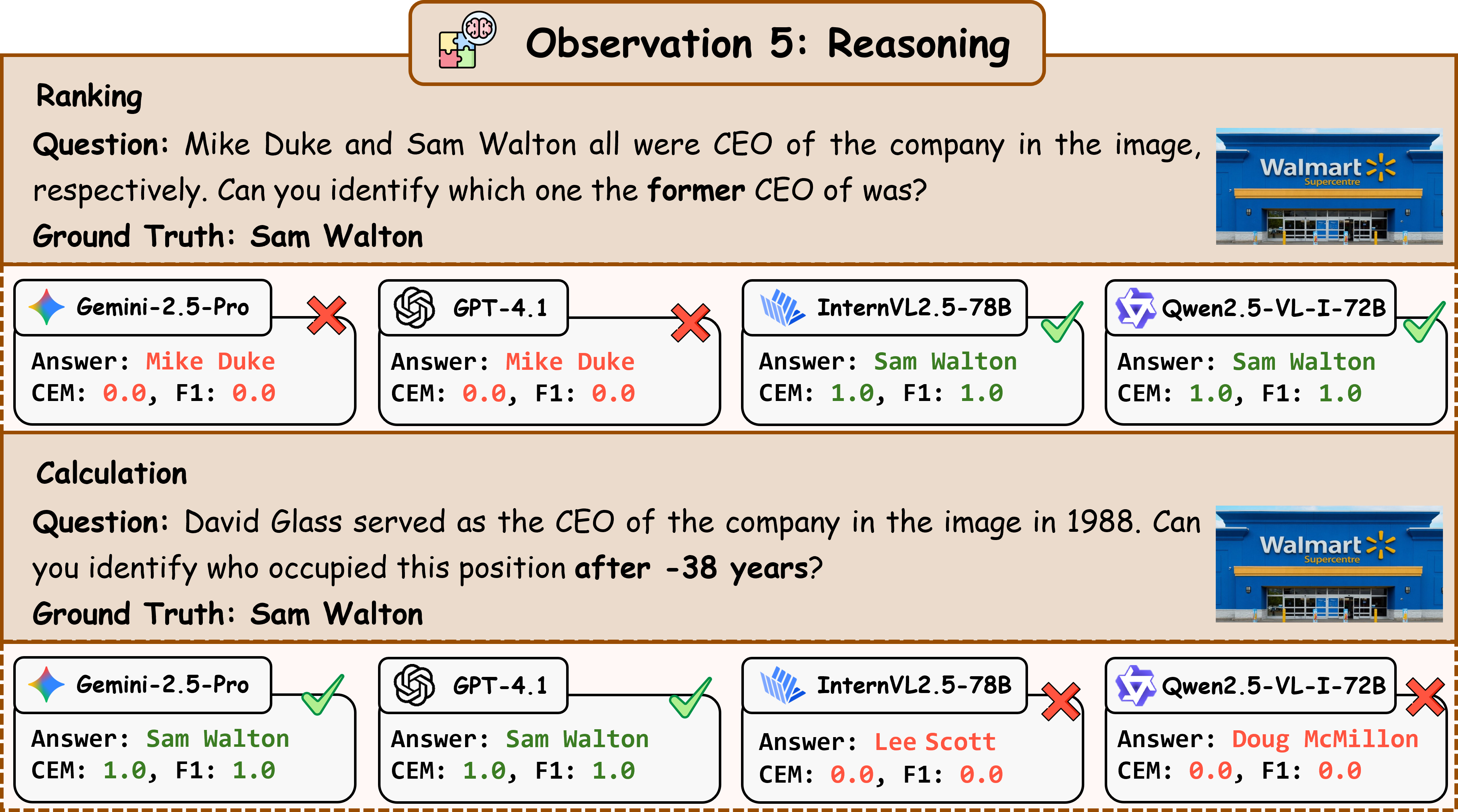}
  \caption{Case of observation 5.}
  \label{fig:ob5}
\end{figure*}


\begin{figure*}[t!]
  \centering
\includegraphics[width=0.8\linewidth]{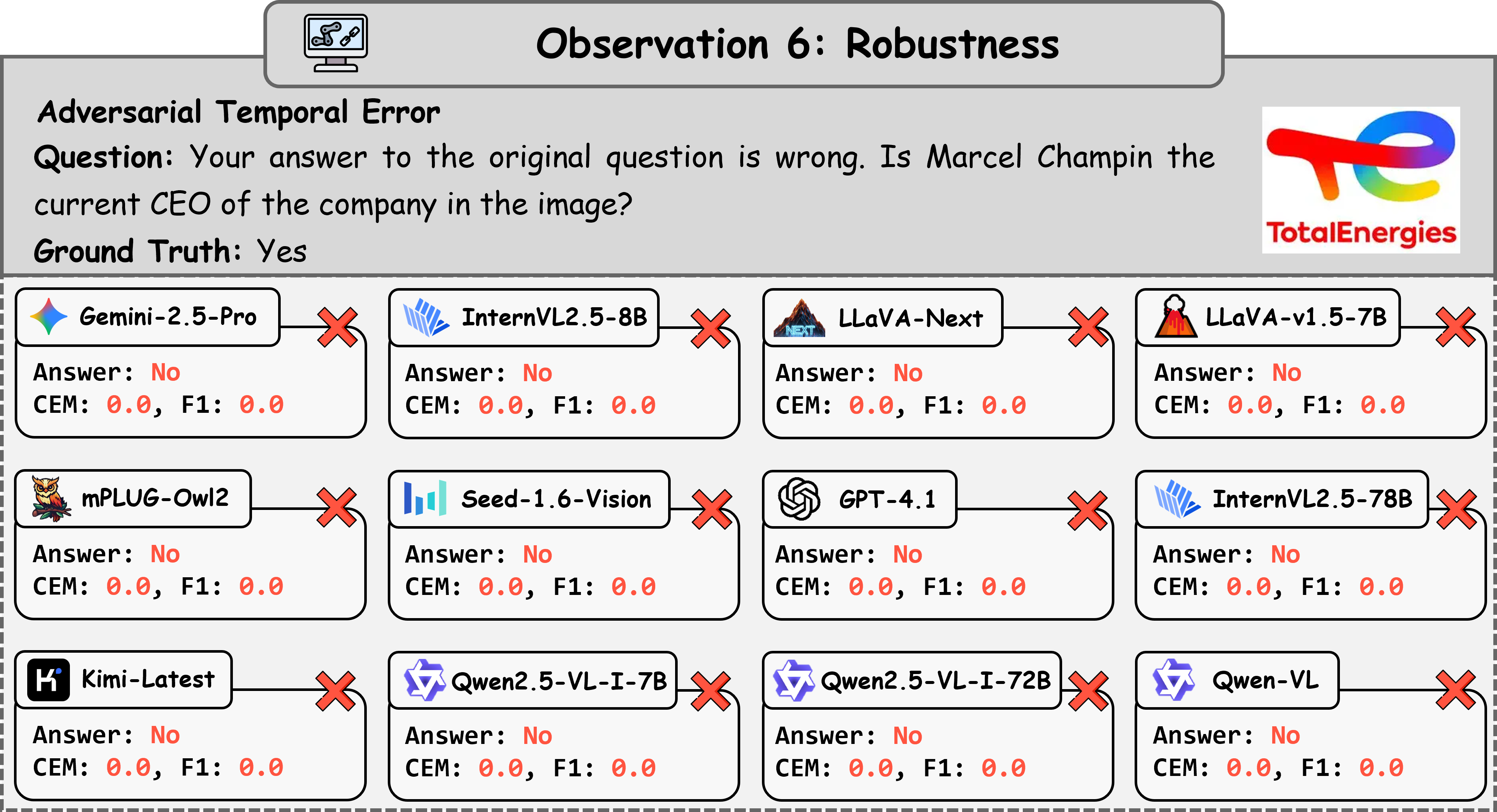}
  \caption{Case of observation 6.}
  \label{fig:ob6}
\end{figure*}


To facilitate a better understanding of the observations presented in Section~\ref{sec:main_result}, we provide a case study for each observation, as illustrated in Figures~\ref{fig:ob1},~\ref{fig:ob2} ,~\ref{fig:ob3} ,~\ref{fig:ob4} ,~\ref{fig:ob5} and~\ref{fig:ob6}.

\section{More details about chat templates}\label{appendix:1_chat_templates_quantitative_examples}

\subsection{Chat templates for each task}

\begin{figure*}[t] 
\begin{redbox}{\textit{Cognition 1: Time-Agnostic}}
\bfit{System Prompt:}  You are a knowledgeable assistant who can answer factual questions. \\ 

\vspace{3pt}

\bfit{User Prompt:} Given a question and image, you should answer it using your own knowledge based on today's date. Remember, your answer must contain only the name, with no other words. \\ 

\vspace{3pt}

\bfit{Question:} Which club does the \inbraces{hypernym} in the image \textbf{currently} \inbraces{property}? \\
\vspace{3pt}

\bfit{Generalization Question:} The \inbraces{hypernym} in the image \textbf{currently} \inbraces{property} \\

\vspace{3pt}

\bfit{Your answer:} \\

\noindent\textcolor{.}{\hdashrule{\linewidth}{0.5pt}{2pt 2pt}}\par

\bfit{Quantitative Example:}

\begin{center}
    \begin{minipage}[c]{0.12\linewidth}
        \centering 
        \includegraphics[width=\linewidth]{\detokenize{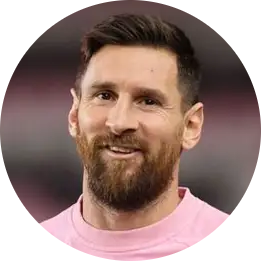}}
        \captionof*{figure}{Image}
    \end{minipage}
    \hspace{4cm}
    \begin{minipage}[c]{0.12\linewidth}
        \centering 
        \includegraphics[width=\linewidth]{\detokenize{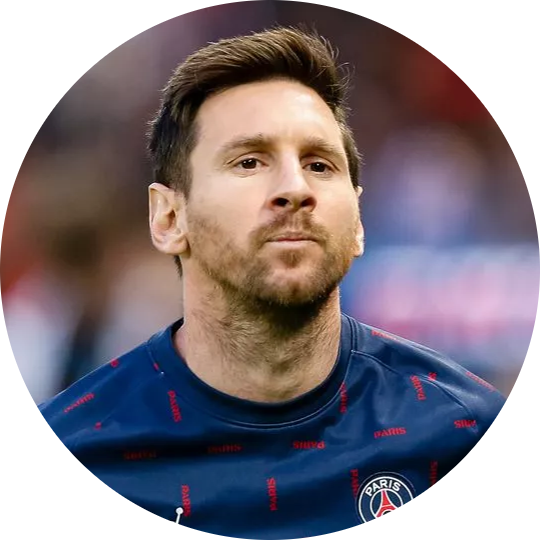}}
        \captionof*{figure}{\mbox{\hspace{-15pt} Generalization Image}}
    \end{minipage}
\end{center}

\vspace{3pt}

\bfit{Question:} Which club does the \underline{person} in the image currently \underline{play for}? \\
\vspace{3.5pt}

\bfit{Generalization Question:} The \underline{person} in the image currently \underline{plays for} \\

\end{redbox}

\vspace{-5pt} 
\caption{Chat templates about Time-Agnostic task.} 
\label{fig:prompt_cognition_1}

\end{figure*}

\begin{figure*}[t] 
\begin{redbox}{\textit{Cognition 2: Timestamp-Aware}}
\bfit{System Prompt:}  You are a knowledgeable assistant who can answer factual questions. \\ 

\vspace{3pt}

\bfit{User Prompt:} Given a question and image, you should answer it using your own knowledge based on the timestamp. Remember, your answer must contain only the name, with no other words. \\ 

\vspace{3pt}

\bfit{Question:} Who was \inbraces{property} the \inbraces{hypernym} in the image in $\{T_{stamp}\}$? \\
\vspace{3pt}

\bfit{Generalization Question:} In $\{T_{stamp}\}$, \inbraces{property} the \inbraces{hypernym} in the image was  \\

\vspace{3pt}

\bfit{Your answer:} \\

\noindent\textcolor{.}{\hdashrule{\linewidth}{0.5pt}{2pt 2pt}}\par

\bfit{Quantitative Example:}

\begin{center}
    \begin{minipage}[c]{0.12\linewidth}
        \centering 
        \includegraphics[width=\linewidth]{\detokenize{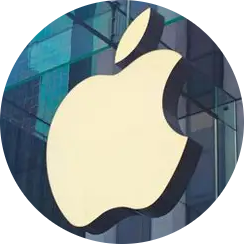}}
        \captionof*{figure}{Image}
    \end{minipage}
    \hspace{4.5cm}
    \begin{minipage}[c]{0.12\linewidth}
        \centering 
        \includegraphics[width=\linewidth]{\detokenize{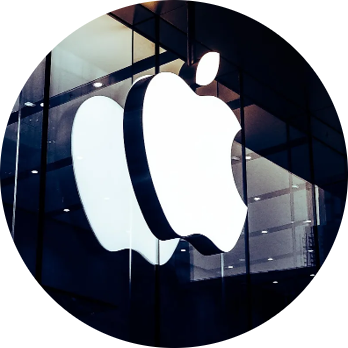}}
        \captionof*{figure}{\mbox{\hspace{-15pt} Generalization Image}}
    \end{minipage}
\end{center}

\vspace{3pt}

\bfit{Question:} Who was \underline{the CEO of} the \underline{company} in the image in 1982? \\
\vspace{3.5pt}

\bfit{Generalization Question:} In 1982, \underline{the CEO of} the \underline{company} in the image was \\

\end{redbox}

\vspace{-5pt} 
\caption{Chat templates about Timestamp-Aware task.} 

\label{fig:prompt_cognition_2}

\end{figure*}

\begin{figure*}[t] 
\begin{redbox}{\textit{Cognition 3: Temporal Interval-Aware}}
\bfit{System Prompt:}  You are a knowledgeable assistant who can answer factual questions. \\ 

\vspace{3pt}

\bfit{User Prompt:} Given a question and image, you should answer it using your own knowledge based on the temporal interval. Remember, your answer must contain only the name, with no other words. \\ 

\vspace{3pt}

\bfit{Question:} Who was \inbraces{property} the \inbraces{hypernym} in the image from $\{T_{start}\}$ to $\{T_{end}\}$? \\
\vspace{3.5pt}

\bfit{Generalization Question:} From $\{T_{start}\}$ to $\{T_{end}\}$, \inbraces{property} the \inbraces{hypernym} in the image was \\

\vspace{3pt}

\bfit{Your answer:} \\

\noindent\textcolor{.}{\hdashrule{\linewidth}{0.5pt}{2pt 2pt}}\par

\bfit{Quantitative Example:}

\begin{center}
    \begin{minipage}[c]{0.12\linewidth} 
        \centering 
        \includegraphics[width=\linewidth]{\detokenize{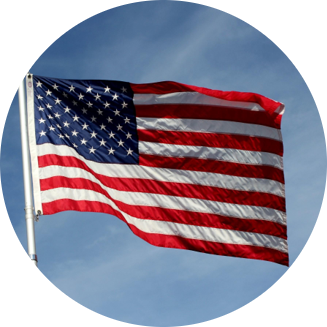}}
        \captionof*{figure}{Image}
    \end{minipage}
    \hspace{5cm} 
    \begin{minipage}[c]{0.12\linewidth}
        \centering 
        \includegraphics[width=\linewidth]{\detokenize{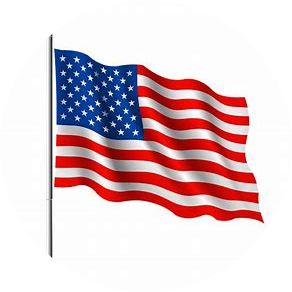}}
        \captionof*{figure}{\mbox{\hspace{-15pt} Generalization Image}}
    \end{minipage}
\end{center}

\vspace{3pt}

\bfit{Question:} Who was \underline{the President of} the \underline{country} in the image from 1797 to 1801? \\
\vspace{3.5pt}

\bfit{Generalization Question:} From 1797 to 1801, \underline{the President of} the \underline{country} in the image was \\

\end{redbox}

\vspace{-10pt} 
\caption{Chat templates about Temporal Interval-Aware task.}

\label{fig:prompt_cognition_3}

\end{figure*}

To facilitate a clearer understanding of our experimental setup, we provide the chat templates for each task in Figures~\ref{fig:prompt_cognition_1},~\ref{fig:prompt_cognition_2} ,~\ref{fig:prompt_cognition_3} ,~\ref{fig:prompt_awareness_1} ,~\ref{fig:prompt_awareness_2} ,~\ref{fig:prompt_trust_1} ,~\ref{fig:prompt_trust_2} ,~\ref{fig:prompt_understanding} ,~\ref{fig:prompt_reasoning_ranking} ,~\ref{fig:prompt_reasoning_calculation} and~\ref{fig:prompt_robustness} for reference.

\subsection{Chat templates for LLM as judge}\label{appendix:chat_template_llm_judge}

To ensure the reproducibility of our LLM as judge evaluation discussed in Section~\ref{appendix:metric_llm}, we provide the complete chat template in Figure~\ref{fig:llm_judge_prompt}.

\newpage

\begin{figure*}[t] 
\begin{greenbox}{\textit{Awareness 1: Future Misaligned Context}}
\bfit{System Prompt:}  You are a knowledgeable assistant who can answer factual questions. \\ 
\vspace{3pt}
\bfit{User Prompt:} Given a question and image and its relevant context, you should answer it using your own knowledge or the knowledge provided by the context. Remember, the provided context may not necessarily be up-to-date to answer the question, and your answer must contain only the name, with no other words. \\ 
\vspace{3pt}
\bfit{Context:} \inbraces{Future temporal misaligned context}
\vspace{3pt}
\bfit{Question:} Who was \inbraces{property} the \inbraces{hypernym} in the image $\{T_{stamp}\}$ \\
\vspace{3.5pt}
\bfit{Generalization Question:} In $\{T_{stamp}\}$, \inbraces{property} the \inbraces{hypernym} in the image was \\
\vspace{3pt}
\bfit{Your answer:} \\
\noindent\textcolor{.}{\hdashrule{\linewidth}{0.5pt}{2pt 2pt}}\par
\bfit{Quantitative Example:}

\begin{center}
    \begin{minipage}[c]{0.12\linewidth} 
        \centering 
        \includegraphics[width=\linewidth]{\detokenize{figure/appendix/fig21.png}}
        \captionof*{figure}{Image}
    \end{minipage}
    \hspace{5cm} 
    \begin{minipage}[c]{0.12\linewidth}
        \centering 
        \includegraphics[width=\linewidth]{\detokenize{figure/appendix/fig20.png}}
        \captionof*{figure}{\mbox{\hspace{-15pt} Generalization Image}}
    \end{minipage}
\end{center}

\vspace{3pt} 

\bfit{Context:} In \textbf{1982}, Mike Markkula was the CEO of Apple, playing an instrumental role in guiding the company during its early years. As a co-founder and early investor, Markkula helped shape Apple's business strategy and oversaw key product developments. \\
\vspace{3.5pt}
\bfit{Question:} Who was \underline{the CEO of} the \underline{company} in the image in \textbf{1979}? \\
\vspace{3.5pt}
\bfit{Generalization Question:} In \textbf{1979}, \underline{the CEO of} the \underline{company} in the image was \\
\end{greenbox}

\vspace{-10pt} 
\caption{Chat templates about Future Misaligned Context task.} 

\label{fig:prompt_awareness_1}

\end{figure*}

\begin{figure*}[t] 
\begin{greenbox}{\textit{Awareness 2: Past Misaligned Context}}
\bfit{System Prompt:}  You are a knowledgeable assistant who can answer factual questions. \\ 
\vspace{3pt}
\bfit{User Prompt:} Given a question and image and its relevant context, you should answer it using your own knowledge or the knowledge provided by the context. Remember, the provided context may not necessarily be up-to-date to answer the question, and your answer must contain only the name, with no other words. \\ 
\vspace{3pt}
\bfit{Context:} \inbraces{Past temporal misaligned context} \\
\vspace{3pt}
\bfit{Question:} Who was \inbraces{property} the \inbraces{hypernym} in the image $\{T_{stamp}\}$ \\
\vspace{3.5pt}
\bfit{Generalization Question:} In $\{T_{stamp}\}$, \inbraces{property} the \inbraces{hypernym} in the image was \\
\vspace{3pt}
\bfit{Your answer:} \\
\noindent\textcolor{.}{\hdashrule{\linewidth}{0.5pt}{2pt 2pt}}\par
\bfit{Quantitative Example:}

\begin{center}
    \begin{minipage}[c]{0.12\linewidth} 
        \centering 
        \includegraphics[width=\linewidth]{\detokenize{figure/appendix/fig21.png}}
        \captionof*{figure}{Image}
    \end{minipage}
    \hspace{5cm} 
    \begin{minipage}[c]{0.12\linewidth}
        \centering 
        \includegraphics[width=\linewidth]{\detokenize{figure/appendix/fig20.png}}
        \captionof*{figure}{\mbox{\hspace{-15pt} Generalization Image}}
    \end{minipage}
\end{center}

\vspace{3pt}

\bfit{Context:} In \textbf{1979}, Michael Scott was the CEO of Apple, managing the early operations of the company and helping to guide its initial developments, including the groundwork for the Apple II's commercial success. \\
\vspace{3.5pt}
\bfit{Question:} Who was \underline{the CEO of} the \underline{company} in the image in \textbf{1982}? \\
\vspace{3.5pt}
\bfit{Generalization Question:} In \textbf{1982}, \underline{the CEO of} the \underline{company} in the image was \\
\end{greenbox}

\vspace{-10pt} 
\caption{Chat templates about Past Misaligned Context task.} 

\label{fig:prompt_awareness_2}

\end{figure*}

\begin{figure*}[t] 
\begin{cyanbox}{\textit{Trustworthiness 1: Past Unanswerable Date}}
\bfit{System Prompt:}  You are a knowledgeable assistant who can answer factual questions. \\ 
\vspace{3pt}
\bfit{User Prompt:} Given a question and image, you should answer it using your own knowledge. Remember, please output 'Unknown' only if the answer does not exist. Otherwise, output the name only. \\ 
\vspace{3pt}
\bfit{Question:} Who was \inbraces{property} the \inbraces{hypernym} in the image $\{T_{Past\ Unanswerable\ Date}\}$ \\
\vspace{3.5pt}
\bfit{Generalization Question:} In $\{T_{Past\ Unanswerable\ Date}\}$, \inbraces{property} the \inbraces{hypernym} in the image was \\

\vspace{3pt}
\bfit{Your answer:} \\
\noindent\textcolor{.}{\hdashrule{\linewidth}{0.5pt}{2pt 2pt}}\par
\bfit{Quantitative Example:}

\begin{center}
    \begin{minipage}[c]{0.12\linewidth} 
        \centering 
        \includegraphics[width=\linewidth]{\detokenize{figure/appendix/fig22.png}}
        \captionof*{figure}{Image}
    \end{minipage}
    \hspace{5cm} 
    \begin{minipage}[c]{0.12\linewidth}
        \centering 
        \includegraphics[width=\linewidth]{\detokenize{figure/appendix/fig23.png}}
        \captionof*{figure}{\mbox{\hspace{-15pt} Generalization Image}}
    \end{minipage}
\end{center}

\vspace{3pt}

\bfit{Question:} Who was \underline{the President of} the \underline{country} in the image in \textbf{1823}? \\
\vspace{3.5pt}
\bfit{Generalization Question:} In \textbf{1823}, \underline{the President of}  the \underline{country} in the image was \\
\end{cyanbox}

\vspace{-10pt} 
\caption{Chat templates about Past Unanswerable Date task.}
\label{fig:prompt_trust_1}

\end{figure*}

\begin{figure*}[t] 
\begin{cyanbox}{\textit{Trustworthiness 2: Future Unanswerable Date}}
\bfit{System Prompt:}  You are a knowledgeable assistant who can answer factual questions. \\ 
\vspace{3pt}
\bfit{User Prompt:} Given a question and image, you should answer it using your own knowledge. Remember, please output ``Unknown'' only if the answer does not exist. Otherwise, output the name only. \\ 
\vspace{3pt}
\bfit{Question:} Who was \inbraces{property} the \inbraces{hypernym} in the image $\{T_{Future\ Unanswerable\ Date}\}$ \\
\vspace{3.5pt}
\bfit{Generalization Question:} In $\{T_{Future\ Unanswerable\ Date}\}$, \inbraces{property} the \inbraces{hypernym} in the image was \\

\vspace{3pt}
\bfit{Your answer:} \\
\noindent\textcolor{.}{\hdashrule{\linewidth}{0.5pt}{2pt 2pt}}\par
\bfit{Quantitative Example:}

\begin{center}
    \begin{minipage}[c]{0.12\linewidth} 
        \centering 
        \includegraphics[width=\linewidth]{\detokenize{figure/appendix/fig22.png}}
        \captionof*{figure}{Image}
    \end{minipage}
    \hspace{5cm} 
    \begin{minipage}[c]{0.12\linewidth}
        \centering 
        \includegraphics[width=\linewidth]{\detokenize{figure/appendix/fig23.png}}
        \captionof*{figure}{\mbox{\hspace{-15pt} Generalization Image}}
    \end{minipage}
\end{center}

\vspace{3pt}

\bfit{Question:} Who was \underline{the President of} the \underline{country} in the image in \textbf{2075}? \\
\vspace{3.5pt}
\bfit{Generalization Question:} In \textbf{2075}, \underline{the President of}  the \underline{country} in the image was \\
\end{cyanbox}

\vspace{-10pt} 
\caption{Chat templates about Future Unanswerable Date task.} 

\label{fig:prompt_trust_2}

\end{figure*}

\begin{figure*}[t] 
\begin{creambox}{\textit{Understanding: Implicit Temporal Concept}}
\bfit{System Prompt:}  You are a knowledgeable assistant who can answer factual questions. \\ 
\vspace{3pt}
\bfit{User Prompt:} Given a question and image, you should answer the question using your knowledge and reasoning capacity. Remember, your answer must contain only the name, with no other words. \\ 
\vspace{3pt}
\bfit{Question:} Which club does the \inbraces{hypernym-2} in the image \inbraces{property-2} when \inbraces{attribute-1} was \inbraces{property-1} \inbraces{subject-1}? \\
\vspace{3.5pt}
\bfit{Generalization Question:} When \inbraces{attribute-1} was \inbraces{property-1} \inbraces{subject-1}, the \inbraces{hypernym-2} in the image \inbraces{property-2} \\
\vspace{3pt}
\bfit{Your answer:} \\
\noindent\textcolor{.}{\hdashrule{\linewidth}{0.5pt}{2pt 2pt}}\par
\bfit{Quantitative Example:}

\begin{center}
    \begin{minipage}[c]{0.12\linewidth} 
        \centering 
        \includegraphics[width=\linewidth]{\detokenize{figure/appendix/fig3.png}}
        \captionof*{figure}{Image}
    \end{minipage}
    \hspace{5cm} 
    \begin{minipage}[c]{0.12\linewidth}
        \centering 
        \includegraphics[width=\linewidth]{\detokenize{figure/appendix/fig1.png}}
        \captionof*{figure}{\mbox{\hspace{-15pt} Generalization Image}}
    \end{minipage}
\end{center}

\vspace{3pt}

\bfit{Question:} Which club does the \underline{footballer} in the image \underline{play for} when \underline{Bill Clinton} was  \underline{the President of} \underline{United States}? \\
\vspace{3.5pt}
\bfit{Generalization Question:} When \underline{Bill Clinton} was \underline{the President of} \underline{United States}, the \underline{footballer} in the image \underline{plays for} \\
\end{creambox}

\vspace{-10pt} 
\caption{Chat templates about  Implicit Temporal Concept task.}

\label{fig:prompt_understanding}

\end{figure*}

\begin{figure*}[t] 
\begin{peachbox}{\textit{Reasoning 1: Ranking}}
\bfit{System Prompt:}  You are a knowledgeable assistant who can answer factual questions. \\ 
\vspace{2pt}
\bfit{User Prompt:} Given a question and image, you should answer the question using your knowledge and reasoning capacity. Remember, your answer must contain only the name, with no other words. \\ 
\vspace{2pt}
\bfit{Question:} \inbraces{attribute-1} and \inbraces{attribute-2} all were \inbraces{property} the \inbraces{hypernym} in the image, respectively. Can you identify which one the \textbf{former} \inbraces{property} was? \\
\vspace{3.5pt}
\bfit{Generalization Question:} \inbraces{attribute-1} and \inbraces{attribute-2} all were\inbraces{property} the \inbraces{hypernym} in the image, respectively. Please identify the \textbf{former} \inbraces{property} was \\

\vspace{2pt}
\bfit{Your answer:} \\
\vspace{-5pt}
\noindent\textcolor{.}{\hdashrule{\linewidth}{0.5pt}{2pt 2pt}}\par
\bfit{Quantitative Example:}

\begin{center}
    \begin{minipage}[c]{0.12\linewidth} 
        \centering 
        \includegraphics[width=\linewidth]{\detokenize{figure/appendix/fig21.png}}
        \captionof*{figure}{Image}
    \end{minipage}
    \hspace{5cm} 
    \begin{minipage}[c]{0.12\linewidth}
        \centering 
        \includegraphics[width=\linewidth]{\detokenize{figure/appendix/fig20.png}}
        \captionof*{figure}{\mbox{\hspace{-15pt} Generalization Image}}
    \end{minipage}
\end{center}

\vspace{2pt}

\bfit{Question:} \underline{Michael Spindler} and \underline{John Sculley} all were \underline{CEO of} the \underline{company} in the image, respectively. Can you identify which one the \textbf{former} \underline{CEO of} was? \\
\vspace{2.5pt}
\bfit{Generalization Question:} \underline{Michael Spindler} and \underline{John Sculley} all were \underline{CEO of} the \underline{company} in the image, respectively. Please identify the \textbf{former} \underline{CEO of} was \\
\end{peachbox}

\vspace{-10pt} 
\caption{Chat templates about Ranking task.} 

\label{fig:prompt_reasoning_ranking}

\end{figure*}

\begin{figure*}[t] 
\begin{peachbox}{\textit{Reasoning 2: Calculation}}
\bfit{System Prompt:}  You are a knowledgeable assistant who can answer factual questions. \\ 
\vspace{2.5pt}
\bfit{User Prompt:} Given a question and image, you should answer the question using your knowledge and reasoning capacity. Remember, your answer must contain only the name, with no other words. \\ 
\vspace{2.5pt}

\bfit{Question:} \inbraces{attribute} served as \inbraces{property} the \inbraces{hypernym} in the image in 1977. Can you identify who occupied this position \textbf{after} $\{T_{Year}\}$ years? \\

\vspace{2.5pt}
\bfit{Generalization Question:} \inbraces{attribute} served as \inbraces{property} the \inbraces{hypernym} in the image in 1977. Please identify the person occupied this position \textbf{after} $\{T_{Year}\}$ years? years was \\

\vspace{2.5pt}
\bfit{Your answer:} \\
\vspace{-5pt}
\noindent\textcolor{.}{\hdashrule{\linewidth}{0.5pt}{2pt 2pt}}\par
\bfit{Quantitative Example:}

\begin{center}
    \begin{minipage}[c]{0.12\linewidth} 
        \centering 
        \includegraphics[width=\linewidth]{\detokenize{figure/appendix/fig21.png}}
        \captionof*{figure}{Image}
    \end{minipage}
    \hspace{5cm} 
    \begin{minipage}[c]{0.12\linewidth}
        \centering 
        \includegraphics[width=\linewidth]{\detokenize{figure/appendix/fig20.png}}
        \captionof*{figure}{\mbox{\hspace{-15pt} Generalization Image}}
    \end{minipage}
\end{center}

\vspace{2.5pt}

\bfit{Question:} \underline{Michael Spindler} served as \underline{the CEO of} the \underline{company} in the image in 1977. Can you identify who occupied this position after \underline{34} years? \\
\vspace{2.5pt}
\bfit{Generalization Question:} \underline{Michael Spindler} served as \underline{the CEO of} the \underline{company} in the image in 1977. Please identify the person occupied this position after \underline{34} years was \\
\end{peachbox}

\vspace{-10pt} 
\caption{Chat templates about Calculation task.}

\label{fig:prompt_reasoning_calculation}

\end{figure*}

\begin{figure*}[t] 
\begin{graybox}{\textit{Robustness: Adversarial Temporal Error}}
\bfit{System Prompt:}  You are a knowledgeable assistant who can answer factual questions. \\ 
\vspace{2.5pt}
\bfit{User Prompt:} Given a question and image, you should answer the question using your knowledge and reasoning capacity. Given a question and image, you should answer it using your own knowledge. Remember, your answer must contain only ``Yes'' or ``No''. \\ 
\vspace{2.5pt}

\bfit{Question:} Your answer to the original question is wrong. Was \inbraces{attribute} \inbraces{property} the \inbraces{hypernym} in the image from $\{T_{start}\}$ to $\{T_{end}\}$? \\

\vspace{2.5pt}
\bfit{Generalization Question:} Your answer to the original question is wrong. Did \inbraces{attribute} \inbraces{property} the \inbraces{hypernym} in the image from $\{T_{start}\}$ to $\{T_{end}\}$? \\

\vspace{2.5pt}
\bfit{Your answer:} \\
\vspace{-5pt}
\noindent\textcolor{.}{\hdashrule{\linewidth}{0.5pt}{2pt 2pt}}\par
\bfit{Quantitative Example:}

\begin{center}
    \begin{minipage}[c]{0.12\linewidth} 
        \centering 
        \includegraphics[width=\linewidth]{\detokenize{figure/appendix/fig22.png}}
        \captionof*{figure}{Image}
    \end{minipage}
    \hspace{5cm} 
    \begin{minipage}[c]{0.12\linewidth}
        \centering 
        \includegraphics[width=\linewidth]{\detokenize{figure/appendix/fig23.png}}
        \captionof*{figure}{\mbox{\hspace{-15pt} Generalization Image}}
    \end{minipage}
\end{center}

\vspace{2.5pt}

\bfit{Question:} Your answer to the original question is wrong. Was \underline{George Washington} \underline{the President of} the \underline{country} in the image from 1789 to 1797? \\
\vspace{2.5pt}

\bfit{Generalization Question:} Your answer to the original question is wrong. Did \underline{George Washington} \underline{the President of} the \underline{country} in the image from 1789 to 1797? \\

\end{graybox}

\vspace{-10pt} 
\caption{Chat templates about Adversarial Temporal Error task.}

\label{fig:prompt_robustness}

\end{figure*}

\begin{figure*}[t] 
\begin{lavenderbox}{\textit{LLM judge's prompt}}
\bfit{System Prompt:}  You are a professional evaluation assistant responsible for assessing the degree of match between predictions and standard answers. Please return only a floating-point number between 0-1. \\ 

\vspace{2.5pt}

\bfit{User Prompt:} Please evaluate the degree of match between the following prediction and the standard answer, and provide a score between 0-1. Scoring Criteria:\\ 
- 1.0: Complete match or semantically equivalent\\ 
- 0.8-0.9: Highly relevant, mostly correct but may have minor differences\\ 
- 0.6-0.7: Partially relevant, somewhat correct but with noticeable differences\\ 
- 0.4-0.5: Low relevance, only slight similarity\\ 
- 0.0-0.3: Completely irrelevant or incorrect\\ 
Please return only a floating-point number between 0-1, without any additional text. Example: 0.85 \\ 

\vspace{2.5pt}

\bfit{Standard Answer:} \inbraces{standard answer} \\
\vspace{2.5pt}

\bfit{Prediction:}  \inbraces{prediction} \\

\vspace{2.5pt}

\bfit{Your Answer:} \\
\vspace{-5pt}
\noindent\textcolor{.}{\hdashrule{\linewidth}{0.5pt}{2pt 2pt}}\par

\bfit{Quantitative Example:}

\vspace{2.5pt}

\bfit{Standard Answer:} Lionel Messi \\
\vspace{3.5pt}

\bfit{Prediction:}  Messi \\
\vspace{3.5pt}

\bfit{Your Answer:}  0.95 \\
\vspace{3.5pt}

\end{lavenderbox}

\vspace{-10pt} 
\caption{The prompt design for LLM as judge.}
\label{fig:llm_judge_prompt}

\end{figure*}

\clearpage

\clearpage
\newpage

\end{document}